%% file: main.tex
\documentclass{article}

\usepackage[preprint]{main}

\usepackage[utf8]{inputenc} % allow utf-8 input
\usepackage[T1]{fontenc}    % use 8-bit T1 fonts
\usepackage{hyperref}       % hyperlinks
\usepackage{url}            % simple URL typesetting
\usepackage{booktabs}       % professional-quality tables
\usepackage{amsfonts}       % blackboard math symbols
\usepackage{nicefrac}       % compact symbols for 1/2, etc.
\usepackage{microtype}      % microtypography
\usepackage{xcolor}         % colors

\usepackage{multirow}
\usepackage{graphicx}
\usepackage{amsmath}
\usepackage{amssymb}
\usepackage{bbding}
\usepackage{pifont}
\usepackage{soul}
\usepackage{threeparttable}
\usepackage{wrapfig}
\usepackage{array}
\usepackage{caption}

\definecolor{my_green}{RGB}{51,102,0}
\definecolor{my_red}{RGB}{204, 0, 0}

\definecolor{paired-light-blue}{RGB}{198, 219, 239}
\definecolor{paired-dark-blue}{RGB}{49, 130, 188}
\definecolor{paired-light-orange}{RGB}{251, 208, 162}
\definecolor{paired-dark-orange}{RGB}{230, 85, 12}
\definecolor{paired-light-green}{RGB}{199, 233, 193}
\definecolor{paired-dark-green}{RGB}{49, 163, 83}
\definecolor{paired-light-purple}{RGB}{218, 218, 235}
\definecolor{paired-dark-purple}{RGB}{117, 107, 176}
\definecolor{paired-light-gray}{RGB}{217, 217, 217}
\definecolor{paired-dark-gray}{RGB}{99, 99, 99}
\definecolor{paired-light-pink}{RGB}{222, 158, 214}
\definecolor{paired-dark-pink}{RGB}{123, 65, 115}
\definecolor{paired-light-red}{RGB}{231, 150, 156}
\definecolor{paired-dark-red}{RGB}{131, 60, 56}
\definecolor{paired-light-yellow}{RGB}{231, 204, 149}
\definecolor{paired-dark-yellow}{RGB}{141, 109, 49}  
\definecolor{myblue}{RGB}{218,232,252}
\definecolor{mygray}{RGB}{220,220,220}
\definecolor{mypink}{RGB}{251,49,153}

\title{Open-Sora Plan: Open-Source Large Video Generation Model}

\author{Open-Sora Plan Team\thanks{See Contributions section for full author list.}
}

\begin{document}

\maketitle

\input{sec/0_abstract}    
\input{sec/1_intro}

\input{sec/2_overview}
\input{sec/3_model}
\input{sec/5_strategy}
\input{sec/4_data}

\input{sec/6_result}
\input{sec/7_limitation}
\input{sec/8_conclusion}
\input{sec/9_author}
{
    \small
    \bibliographystyle{ieeenat_fullname}
    \bibliography{main}
}

\input{sec/X_suppl}
\end{document}

%% file: sec/0_abstract.tex
\begin{abstract}
We introduce \textbf{Open-Sora Plan}, an open-source project that aims to contribute a large generation model for generating desired high-resolution videos with long durations based on various user inputs.
Our project comprises multiple components for the entire video generation process, including a Wavelet-Flow Variational Autoencoder, a Joint Image-Video Skiparse Denoiser, and various condition controllers.
Moreover, many assistant strategies for efficient training and inference are designed, and a multi-dimensional data curation pipeline is proposed for obtaining desired high-quality data.
Benefiting from efficient thoughts, our Open-Sora Plan achieves impressive video generation results in both qualitative and quantitative evaluations. 
We hope our careful design and practical experience can inspire the video generation research community.
All our codes and model weights are publicly available at \textcolor{blue}{\url{https://github.com/PKU-YuanGroup/Open-Sora-Plan}}.
\end{abstract}

%% file: sec/1_intro.tex
\section{Introduction}
\label{sec:intro}
Driven by the recent progress of the diffusion model~\citep{ho2020denoising,song2020denoising} and transformer~\citep{vaswani2017attention,peebles2023scalable} architecture, visual content generation demonstrates impressive creation capacity conditioned on given prompts, which attracts broad interests and emerging attempts.
Since the image generation methods~\citep{stable_diffusion,li2024hunyuan} achieve outstanding performance and are applied extensively, the video generation model is expected to make significant advancements to empower a variety of creative industries including entertainment, advertising, film, \textit{etc.}
Many early attempts~\citep{guo2023animatediff,dynamicrafter} successfully generate video with low resolution and short frames, but few efforts challenge the high-quality and long-duration video generation due to the unimaginable computation and data cost.

However, the technique report of Sora~\citep{videoworldsimulators2024}, the video generation model created by OpenAI, with impressive showcases is released suddenly, shocking the entire video generation community while pointing out a promising way to create remarkable videos.
As one of the first open-source projects aiming to re-implement a powerful Sora-like video generation model, our Open-Sora Plan attracts wide attention and contributes many first attempts to the video generation community, which inspires many subsequent works.

In this work, we summarize our practical experiences in recent months and present the technical details of our Open-Sora Plan, which generates high-quality and long-duration videos queried by various categories of conditions including text prompts, multiple images, and structure control signals (canny, depth, sketch, \textit{etc.}).
As illustrated in Fig.~\ref{fig:overview}, we divide the video generation model into three key components and propose improvements for each part:

\begin{figure*}[th]
\centering
\includegraphics[width=1.0\textwidth]{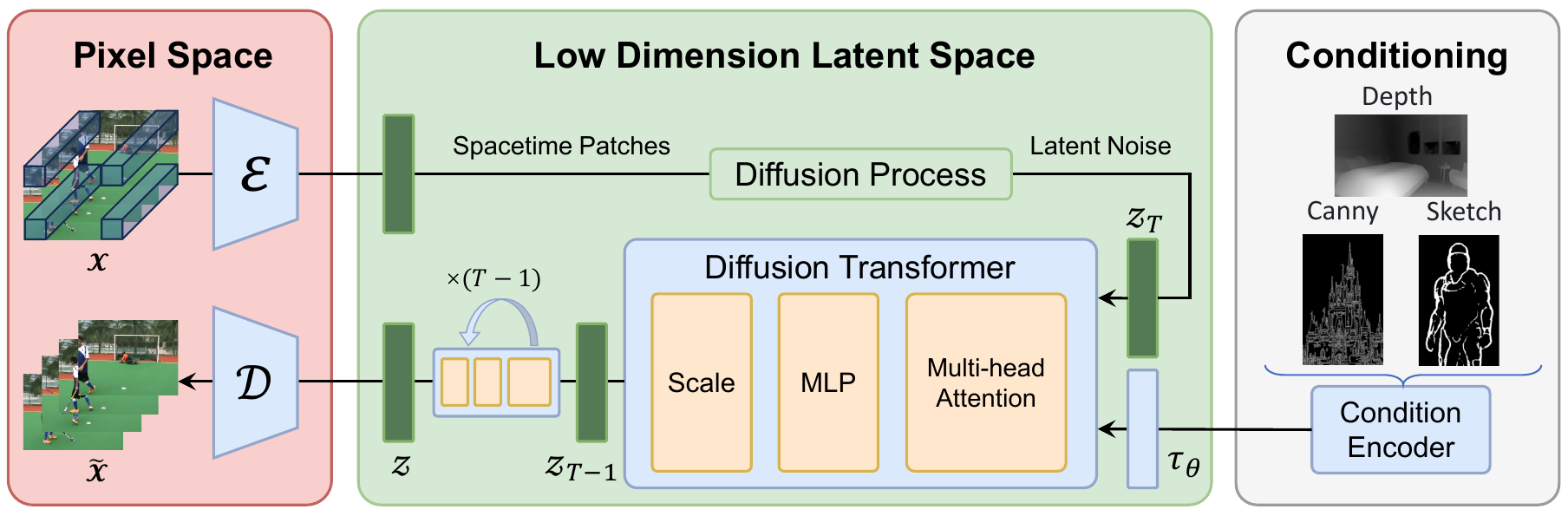}
\caption{The model architecture of the Open-Sora Plan consists of a VAE, a Diffusion Transformer, and conditional encoders. The conditional injection encoders enable precise manipulation of individual frames (whether it's the first frame, a subset of frames, or all frames) using designated structural signals, such as images, canny edges, depth maps, and sketches.
}
\vspace{-10pt}
\label{fig:overview}
\end{figure*}

\begin{itemize}
\item \textbf{Wavelet-Flow Variational Autoencoder.}
To reduce memory usage and enhance training speed, we propose WF-VAE, a model that obtains multi-scale features in the frequency domain through multi-level wavelet transform. 
These features are then injected into a convolutional backbone using a pyramid structure. 
% This approach creates a highway from video flow to latent representation for the primary low-frequency information of the video, simplifying the model structure. While significantly reducing computational costs, WF-VAE achieves comparable reconstruction performance to existing VAEs. 
We also introduced the \textbf{Causal Cache} method to address the issue of latent space disruption caused by tiling inference. 
 % Further details are presented in Section ~\ref{sec:vae}.

\item \textbf{Joint Image-Video Skiparse Denoiser.} 
% The Open-Sora Plan v1.3 is a 2.7B model capable of generating high-quality images or videos based on specified input prompts. 
% The model can produce videos of arbitrary duration within 5 seconds, supporting multiple aspect ratios. 
We \textbf{first} change the 2+1D Sora-like video generation denoiser to a 3D full attention structure, significantly enhancing the model's ability to understand the world, including object motion, camera movement, physics, and human actions.
Our denoiser is capable of creating both high-quality images and videos with specific designs.
We also introduce a cheap but effective operation called \textbf{Skiparse Attention} for further reducing computation. 
% More details are presented in Section~\ref{sec:diff_model}.

\item \textbf{Condition Controllers.}  
We design a frame-level image condition controller to introduce image conditions into the basic model for supporting various tasks including Image-to-Video, Video Transition, and Video Continuation in one framework.
% For details, see Sec.~\ref{sec:i2v_model}. 
Additionally, we develop a novel network architecture to introduce structure conditions into our base model for controllable generation.
% Specifically, it contains two light components: an encoder focuses on extracting a high-level representation from the structural signals, and a projector transforms the representation into the injection feature directly added to the base model.
% We detailedly introduce it in Sec. \ref{Controllner sec:5}.

\end{itemize}

In addition, we carefully design a series of assistant strategies during all stages for training more efficiently and achieving more appreciated results in inference:

\begin{itemize}
\item \textbf{Min-Max Token Strategy.} 
The Open-Sora Plan uses min-max tokens for training, which aggregates data of different resolutions and durations within the same bucket. This strategy unlocks efficient NPUs/GPUs computation and maximizes the effective usage of data. 
% Additionally, we carefully design multiple training phases, allowing the model to progressively learn video generation. 
% The training strategies are outlined in Section~\ref{sec:train_st}. 

\item \textbf{Adaptive Gradient Clipping Strategy.} 
We propose an adaptive gradient clipping strategy that detects outlier data based on the gradient norm at each step, preventing outliers from skewing the model's gradient direction. 
% This adaptive approach generalizes well across various large models. 
% We show the details in Section~\ref{sec:ema_gn}

\item \textbf{Prompt Refinement Strategy.} 
We develop a prompt refiner that enables the model to reasonably expand input prompts while following semantics. Prompt refiner alleviates the issue of inconsistencies in prompt length and descriptive granularity during training and generation, significantly enhancing the stability of video motion and enriching details. 
% We fine-tune a large language model with our custom-built dataset of 11k text instructions. 
% We outline the model details for the prompt refiner in Section~\ref{sec:refiner}.
\end{itemize}

Moreover, we propose an efficient data curation pipeline to automatically filter and annotate visual data from uncleaned datasets:
%in Section~\ref{sec:data}

\begin{itemize}
\item \textbf{Multi-dimensional Data Processor.} 
Our data curation pipeline includes detecting jump cuts, clipping videos, filtering out fast or slow motion, cropping edge subtitles, filtering aesthetic scores, assessing video technical quality, and annotating captions. 

\item \textbf{LPIPS-Based Jump Cuts Detection.} 
We implement a video cut detection method based on Learned Perceptual Image Patch Similarity (LPIPS)~\citep{Zhang_Isola_Efros_Shechtman_Wang_2018} to prevent incorrect segmentation of fast-motion shots. 
% Manual inspection validates that our method accurately identifies video transitions.
\end{itemize}

We notice that our Open-Sora Plan is an underway open-source project and we will make continuous efforts towards high-quality video generation.
All latest news, codes, and model weights will be publicly updated at \textcolor{blue}{\url{https://github.com/PKU-YuanGroup/Open-Sora-Plan}}.

%% file: sec/3_model.tex
\section{Core Models of Open-Sora Plan}
\begin{figure*}[t]
\centering
    \includegraphics[width=\linewidth]{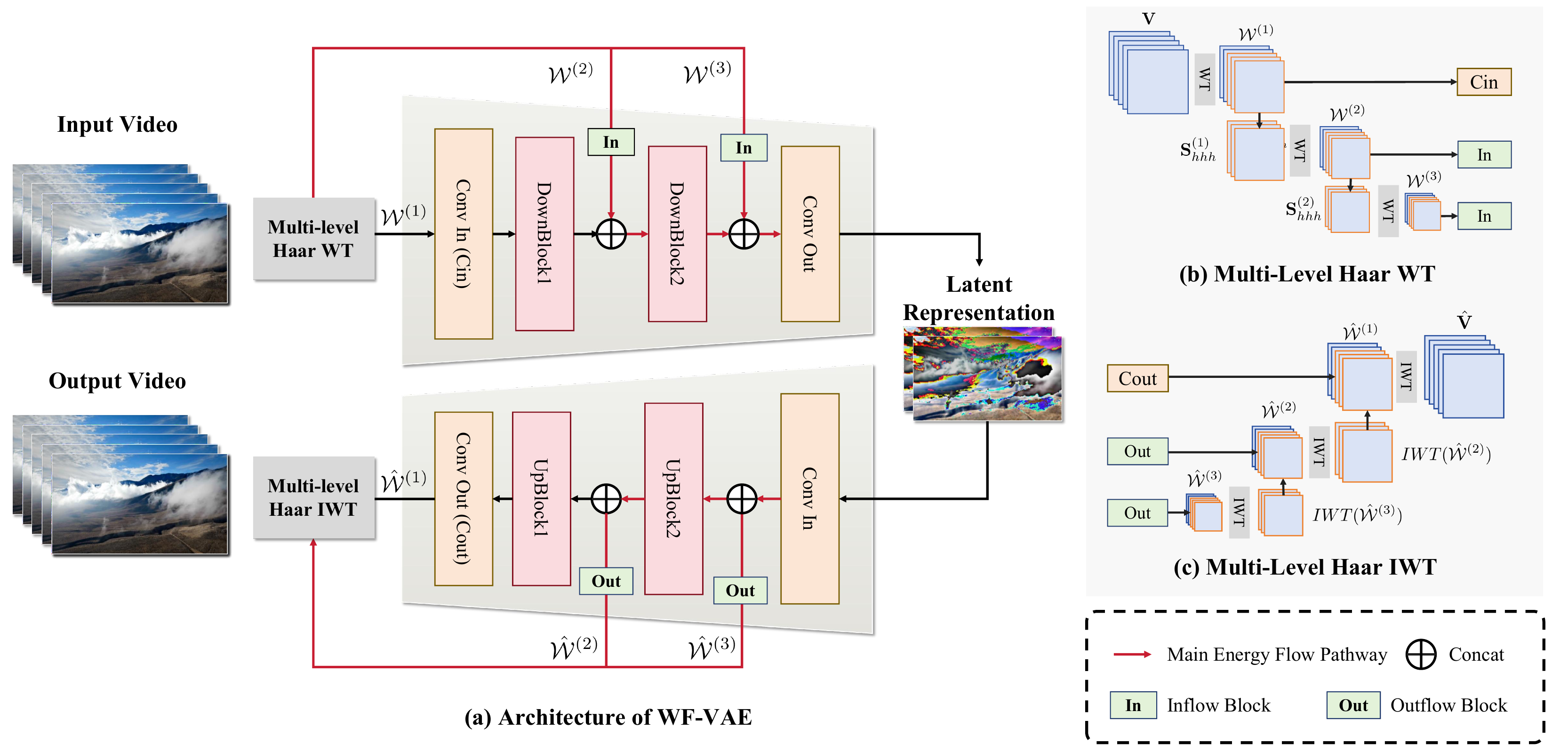}
    \vspace{-10pt}
\caption{\textbf{Overview of WF-VAE}. WF-VAE~\citep{li2024wfvaeenhancingvideovae} consists of a backbone and a main energy path, with such a path injecting the main flow of video energy into the backbone through concatenations.}
\label{vae_design}
\end{figure*}

\subsection{Wavelet-Flow VAE}
\label{sec:vae}

\noindent\textbf{Preliminary.}  The multi-level Haar wavelet transform decomposes video signals by applying scaling filter $\mathbf{h} = \frac{1}{\sqrt{2}}[1, 1]$ and wavelet filter $\mathbf{g} = \frac{1}{\sqrt{2}}[1, -1]$ along temporal and spatial dimensions. For a video signal $\mathbf{V} \in \mathbb{R}^{C \times T \times H \times W}$, where $C$, $T$, $H$, and $W$ correspond to the number of channels, temporal frames, height, and width, the 3D Haar wavelet transform at layer $l$ is defined as:
\begin{equation}
\mathbf{S}^{(l)}_{ijk} = \mathbf{S}^{(l-1)} * (f_i \otimes f_j \otimes f_k),
\end{equation}
where $f_i, f_j, f_k \in {\mathbf{h}, \mathbf{g}}$ represent the filters applied along each dimension, and $*$ represents the convolution operation. The transform begins with $\mathbf{S}^{(0)} = \mathbf{V}$, and for subsequent layers, $\mathbf{S}^{(l)} = \mathbf{S}_{hhh}^{(l-1)}$, indicating that each layer operates on the low-frequency component from the previous layer. At each decomposition layer $l$, the transform produces eight sub-band components: $\mathcal{W}^{(l)} = \{ \mathbf{S}_{hhh}^{(l)}, \mathbf{S}_{hhg}^{(l)}, \mathbf{S}_{hgh}^{(l)}, \mathbf{S}_{ghh}^{(l)}, \mathbf{S}_{hgg}^{(l)}, \mathbf{S}_{ggh}^{(l)}, \mathbf{S}_{ghg}^{(l)}, \mathbf{S}_{ggg}^{(l)} \}$. Here, $\mathbf{S}_{hhh}^{(l)}$ represents the low-frequency component across all dimensions, while $\mathbf{S}_{ggg}^{(l)}$ captures high-frequency details. To implement different downsampling rates in the temporal and spatial dimensions, a combination of 2D and 3D wavelet transforms can be implemented. Specifically, to obtain a compression rate of 4$\times$8$\times$8 (temporal$\times$height$\times$width), we can employ a combination of two-layer 3D wavelet transform followed by one-layer 2D wavelet transform.

\noindent\textbf{Training Objective.} Building upon the training strategies outlined in \cite{rombach2022high}, the proposed loss function integrates several components: reconstruction loss (including both L1 and perceptual losses~\citep{Zhang_Isola_Efros_Shechtman_Wang_2018}), adversarial loss, and KL divergence regularization. As illustrated in Fig.~\ref{vae_design}, our model architecture emphasizes a low-frequency energy flow and enforces symmetry between the encoder and decoder. To preserve this architectural principle, we introduce a novel regularization term, denoted as $\mathcal{L}_{W\!L}$ (WL loss), which ensures structural consistency by penalizing deviations from the expected energy flow:
\begin{equation}
\mathcal{L}_{W\!L} = |\hat{\mathcal{W}}^{(2)} - \mathcal{W}^{(2)}| + |\hat{\mathcal{W}}^{(3)} - \mathcal{W}^{(3)}|.
\end{equation}
The overall loss function is defined as:
\begin{equation}
\begin{split}
    \mathcal{L} &= \mathcal{L}_{recon} +  \lambda_{adv} \mathcal{L}_{adv} + \lambda_{K\!L}\mathcal{L}_{K\!L} + \lambda_{W\!L} \mathcal{L}_{W\!L}.
\end{split}
\end{equation}
where $\lambda_{adv}$, $\lambda_{KL}$, and $\lambda_{W\!L}$ are weighting coefficients for the corresponding loss components. Following~\citep{Esser_2021_CVPR}, we adopt dynamic adversarial loss weighting to balance the relative gradient magnitudes of the adversarial and reconstruction losses:
\begin{equation}
   \lambda_{\mathrm{adv}}=\frac{1}{2}\bigg(\frac{\|\nabla_{G_L}[\mathcal{L}_{\mathrm{recon}}]\|}{\|\nabla_{G_L}[\mathcal{L}_{\mathrm{adv}}]\|+\delta}\bigg),
\end{equation}
where $\nabla{G_L}[\cdot]$ represents the gradient with respect to the final layer of the decoder, and $\delta=10^{-6}$ is introduced for numerical stability.

% \begin{figure}[t]
% \begin{minipage}[c]{0.6\linewidth}
%     \includegraphics[width=\linewidth]{fig/Report-WFVAE.pdf}
%     \label{fig:vae_design}
% \end{minipage}
% \hfill
% \begin{minipage}[c]{0.4\linewidth}
% \vspace{20pt}
%   \includegraphics[width=\linewidth]{fig/Causal_chunk.pdf}
%     \label{fig:causalcache}
% \end{minipage}
% \caption{\textbf{\textcolor{red}{Left:} Overview of WF-VAE}. WF-VAE consists of a backbone and a main energy path, with such a path injecting the main flow of video energy into the backbone through concatenations. \textbf{\textcolor{red}{Right:} Illustration of Causal Cache}. We demonstrate an example with $k_t=3$, $s_t=1$, $T_{chunk}=4$ and $T_{cache}(m)=2$, where the convolution maintains continuous sliding.}
% \vspace{-10pt}
% \end{figure}
% \begin{figure}[t]
% \centering
% \includegraphics[width=0.46\linewidth]{fig/Report-WFVAE.pdf}
% \caption{\textbf{Framework of WF-VAE}. Our architecture consists of a backbone and a main energy path. The main energy path functions as a “highway” for the main flow of video energy, channeling this energy into the backbone through concatenations. This design ensures that essential video components retain more information within the latent representation.}
% \hspace{1in}
% \includegraphics[width==0.46\linewidth]{fig/Causal_chunk.pdf}
% \caption{\textbf{Illustration of Causal Cache}. We demonstrate an example with $k_t=3$, $s_t=1$, $T_{chunk}=4$ and $T_{cache}(m)=2$, where the convolution maintains continuous sliding.}
% \label{fig:vae_design}
% \end{figure}
\begin{wrapfigure}{r}{0.5\linewidth}
\vspace{-10pt}
\includegraphics[width=\linewidth]{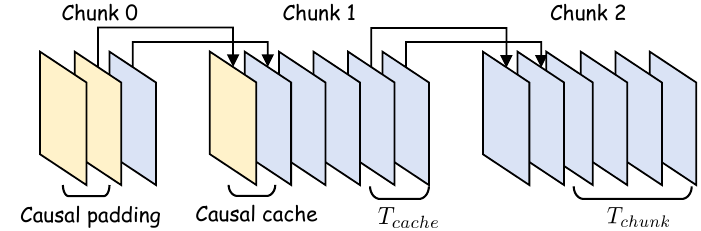}
\caption{Illustration of Causal Cache.}
\label{fig:causalcache}
\end{wrapfigure}

\noindent\textbf{Causal Cache.}
We substitute regular 3D convolutions with causal 3D convolutions~\citep{yu2024languagemodelbeatsdiffusion} in WF-VAE with $k_t-1$ temporal padding at the start, enabling unified processing of images and videos. We extract the first frame and process the remaining frames in chunks of size $T_{chunk}$ for efficient inference of T-frame videos. We cache $T_{cache}(m)$ tail frames between chunks, where:
\begin{equation}
    T_{cache}(m) = k_t + mT_{chunk} - s_t \lfloor \frac{mT_{chunk}}{s_t} + 1 \rfloor.
\end{equation}
This method necessitates that $(T - k_t)$ is divisible by $s_t$ and $(T - 1) \bmod s_t = 0$. We given a  illustrated sample for understanding in Fig.~\ref{fig:causalcache}, with $k_t=3, s_t=1, T_{chunk}=4$, $T_{cache}(m)=2$ frames are cached.

\noindent\textbf{Training Details.}  We utilize the AdamW optimizer~\citep{Kingma_Ba_2014,loshchilov2019decoupledweightdecayregularization} with parameters $\beta_1 = 0.9$ and $\beta_2 = 0.999$, maintaining a fixed learning rate of $1 \times 10^{-5}$. Our training process consists of three stages: \textbf{(i)} In the first stage, following the methodology of~\citep{chen2024od}, we preprocess videos to contain 25 frames at a resolution of $256 \times 256$, with a total batch size of 8. \textbf{(ii)} We update the discriminator, increase the number of frames to 49 and halve the frames per second (FPS) to enhance motion dynamics. \textbf{(iii)} We find that a large $\lambda_{lpips}$ adversely affects video stability; hence, we update the discriminator again and set $\lambda_{lpips}$ to $0.1$.
The initial stage is trained for 800,000 steps, and the subsequent stages are each trained for 200,000 steps. The training process is conducted on 8 NPUs~\citep{liao2021ascend}/GPUs. We employ a 3D discriminator and initiate GAN training from the beginning.

% We leverage this property by creating an energy flow pathway outside the backbone network. The pathway processes wavelet pyramid features ${\mathcal{W}^{(1)},\mathcal{W}^{(2)},\mathcal{W}^{(3)}}$, where $\mathcal{W}^{(1)}$ serves as encoder input and decoder target. Features from both branches are concatenated at corresponding scales, with InflowBlock and OutflowBlock managing the feature transformations. This design enables efficient processing of low-frequency information while maintaining reconstruction quality.

\subsection{Joint Image-Video Skiparse Denoiser}
\label{sec:diff_model}

\begin{figure*}[t]
\includegraphics[width=\textwidth]{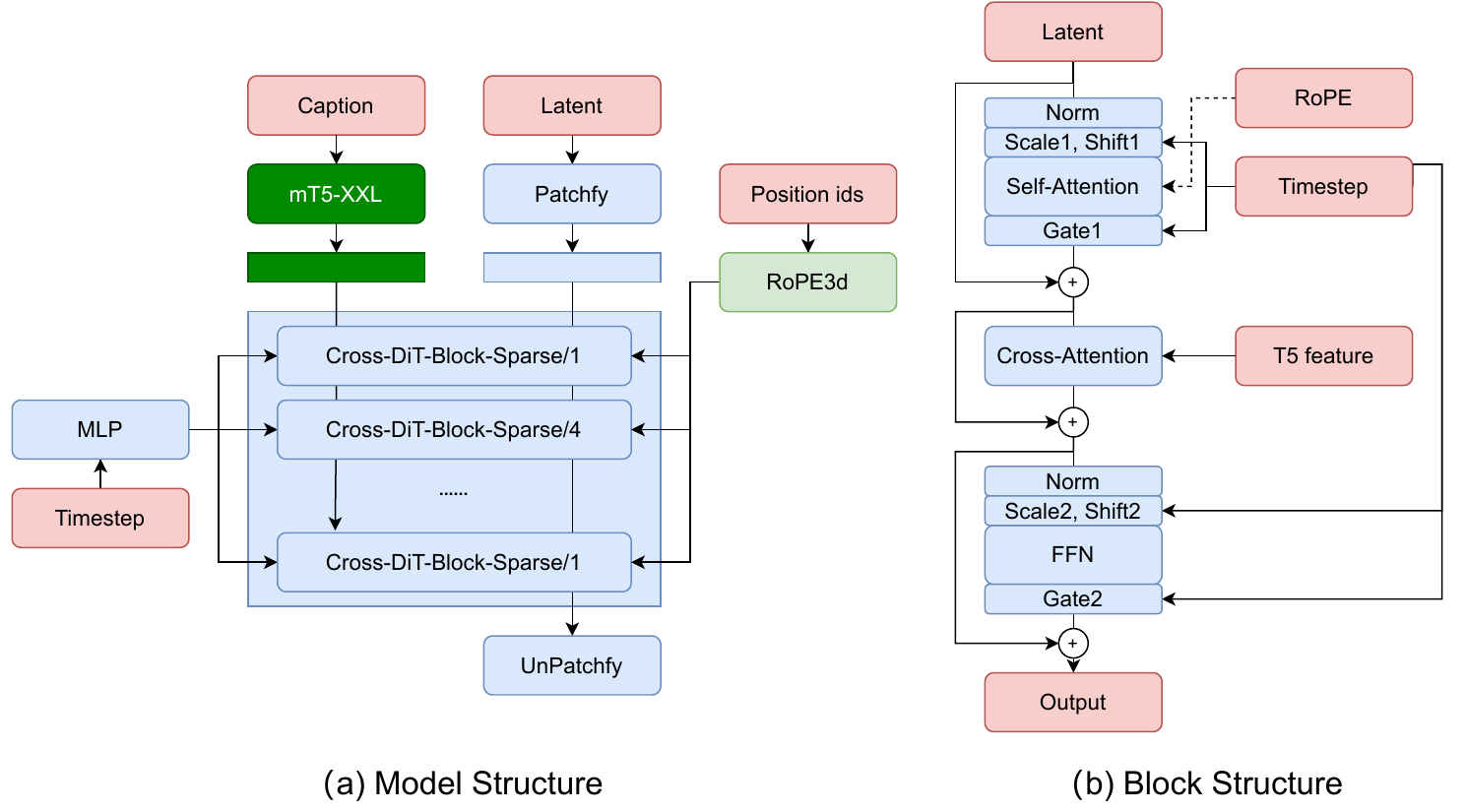}
\caption{\textbf{Overview of the Joint Image-Video Skiparse Denoiser}. The model learns the denoising process in a low-dimensional latent space, which is compressed from input videos via our Wavelet-Flow VAE. Text prompts and timesteps are injected into each Cross-DiT block layer equipped with 3D RoPE. Our Skiparse attention is applied to every layer except the first and last two layers.}
\label{fig:model}
\end{figure*}

\subsubsection{Model Overview} 
\label{sec:model_overview}
As shown in Fig.~\ref{fig:model}, we compress input images or videos from pixel space to latent space for denoising training with the diffusion model. Given an input latent $x \in \mathbb{R}^{B \times C \times T \times H \times W}$, we first split latent into small tokens by a 3D convolutional layer and flattened into a 1D sequence, with converting the latent dimension $C$ to dimension $D$. We use kernel sizes $k_t=1, k_h=2$ and $k_w=2$, with strides matching the kernel sizes, resulting in a total of $L=\frac{T H W}{k_t k_h k_w}$ tokens. We further use mT5-XXL~\citep{xue2020mt5} as the text encoder to map text prompts to a high-dimensional feature space, and we also convert text feature to dimension $D$ through a single MLP layer.

\noindent\textbf{3D RoPE.} We employ 3D rotational position encoding, which allows the model to directly compare relative differences between positions rather than relying on absolute positions. We define the computation process of $nD$ RoPE. After ``patchifying'' operation, the latent $\mathbf{X} \in \mathbb{R}^{B \times L \times D}$ is divided into $n$ parts along the $D$ dimension, \textit{e.g.}, $\mathbf{X}=\left[\mathbf{X}_{\mathbf{1}}, \ldots, \mathbf{X}_{\mathbf{n}}\right]$, where $\mathbf{X}_{\mathbf{i}} \in \mathbb{R}^{B \times L \times \frac{D}{n}}$, $ i \in[1, \ldots, n]$, and we apply RoPE on partitioned tensor $\mathbf{X}_{\mathbf{i}}$. Assuming the RoPE operation~\citep{su2024roformer} is denoted as $\operatorname{RoPE}\left(\mathbf{X}_{\mathbf{i}}\right)$, we inject the relative position encoding of the i-th dimension into tensor $\mathbf{X}_{\mathbf{i}}$, and concatenate processed tensors along the $D$ dimension to obtain the final result:

\begin{align}
\mathbf{X}_{\mathbf{i}}^{\text {rope }}&=\operatorname{RoPE}\left(\mathbf{X}_{\mathbf{i}}\right), \\
\mathbf{X}_{\text {final }}&=\operatorname{Concat}\left(\left[\mathbf{X}_{\mathbf{1}}^{\text {rope }}, \ldots, \mathbf{X}_{\mathbf{n}}^{\text {rope }}\right]\right),
\end{align}

where $\operatorname{Concat}(\cdot)$ denotes the concatenate operation and $\mathbf{X}_{\text {final }} \in \mathbb{R}^{B \times L \times D}$. When $n=1$, it is equivalent to applying RoPE on a 1D sequence in large language models. When $n=2$, it can be viewed as 2D RoPE applied along the height and width directions of an image. When $n=3$, RoPE is successfully applied to video data by incorporating relative position encoding in both the temporal and spatial dimensions to enhance the representation of sequences.

\noindent\textbf{Block Design.} Inspired by large language model architectures~\citep{dubey2024llama,yang2024qwen2,jiang2023mistral,young2024yi}, we adopt a pre-norm transformer block structure primarily comprising self-attention, cross-attention, and a feedforward network. 
Following~\citep{peebles2023scalable,chen2023pixartalpha}, we map timesteps to two sets of scale, shift, and gate parameters through \textit{adaLN-Zero}~\citep{peebles2023scalable}.
We then inject such two sets of values to self-attention and the FFN separately, and 3D RoPE is employed in self-attention layers. 
In version 1.2, we start to introduce Full 3D Attention instead of 2+1D Attention for significantly enhancing video motion smoothness and visual quality. 
However, the quadratic complexity of Full 3D Attention requires substantial computational resources, thus we propose a novel sparse attention mechanism. 
To ensure direct 3D interaction, we retain Full 3D Attention in the first and last two layers.

\subsubsection{Skiparse Attention} 
\label{sec:skiparse}
The 2+1D Attention widely leveraged by former video generation methods calculates frame interactions only along the temporal dimension, theoretically and practically limiting video generation performance. 
% In real-world scenarios, any event is not only affected by events at the same spatial position but different moments or at the same moment but different spatial positions. 
% It is more often influenced by events at different positions and moments. 
% Here’s a simple example: In a boxing match, except for straight punches, every punch comes from different spatial positions over time. 
% This helps boxers surprise their opponents and win the match. This constraint makes it challenging for 2+1D DiT to model complex physical dynamics accurately.
Compared to 2+1D Attention, Full 3D Attention represents global calculation for allowing content from arbitrarily spatial and temporal positions to interact, which approach aligns well with real-world physics. 
However, Full 3D Attention is time-consuming and inefficient, as visual information often contains considerable redundancy, making it unnecessary to establish attention across all spatiotemporal tokens.
An ideal spatiotemporal modeling approach should employ attention that minimizes the overhead from redundant visual information while capturing the complexities of the dynamic physical world. Reducing redundancy requires avoiding connections among all tokens, yet global attention remains essential for modeling complex physical interactions.

\begin{figure}[t]
\centering
\includegraphics[width=\linewidth]{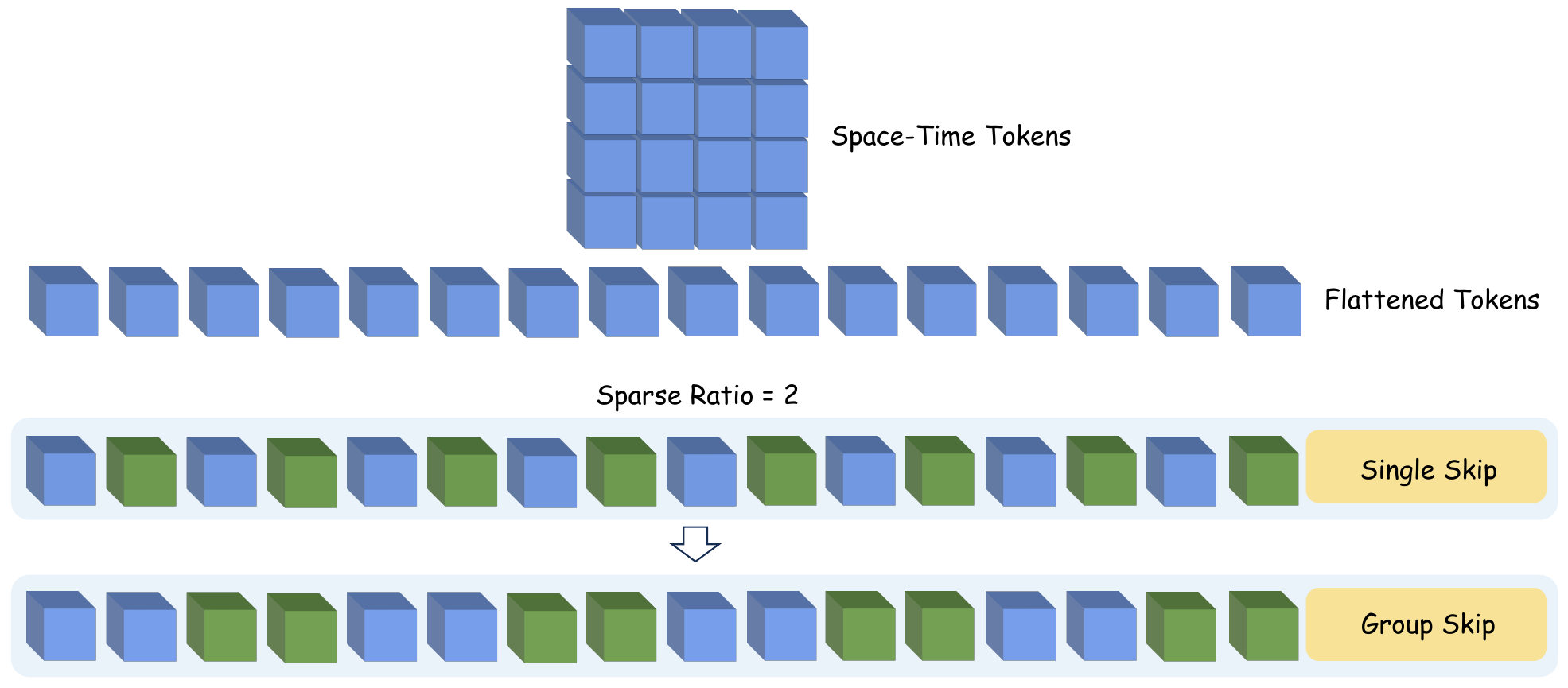}
\caption{\textbf{Calculation process of Skiparse Attention} with sparse ratio $k=2$ for example. In our Skiparse Attention operation, we alternately perform the Single Skip and the Group Skip operations, reducing the sequence length to $1/k$ compared to the original size in each operation.}
\label{fig:model_skiparse_attention}
\end{figure}

\begin{figure*}[t]
    \centering
    \includegraphics[width=0.9\textwidth]{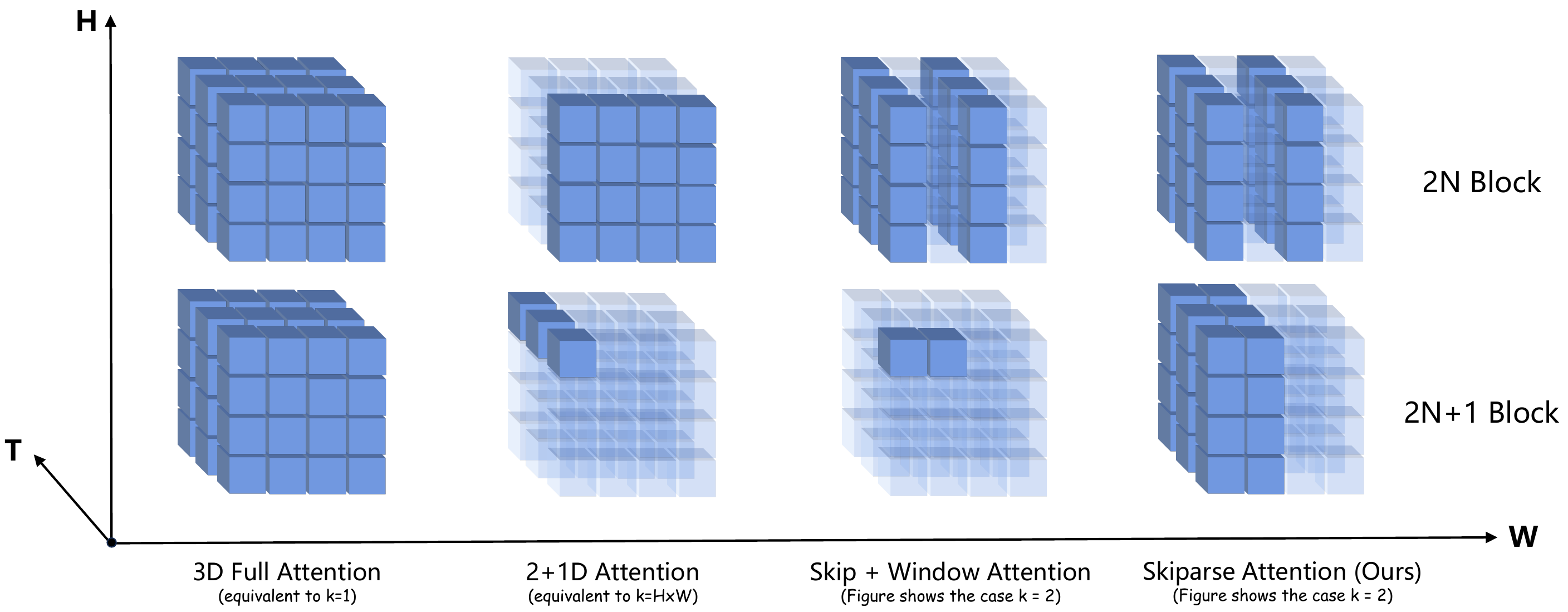}
    \caption{\textbf{The interacted sequence scope of different attention mechanisms.} Various attention mainly differ in the number and position of selected tokens during attention computations.}
    \label{fig:attention_comparison}
\end{figure*}
% Each mechanism defines unique patterns for which tokens are included in the attention process.

To balance the computation efficiency and spatiotemporal modeling ability, we propose a \textbf{Skiparse} (Skip-Sparse) Attention mechanism.
Denoiser with Skiparse Attention only modifies the original attention layers to two alternating sparse attention operations named \textbf{Single Skip} and \textbf{Group Skip} in Transformer Blocks.
Giving a sparse ratio $k$, the sequence length in the attention operation reduces to $\frac{1}{k}$ compared to the original, and batch size increases by $k$-fold, lowering the theoretical complexity of self-attention to $\frac{1}{k}$, while cross attention complexity remains unchanged.

The Calculation process of two skip operations is shown Fig.~\ref{fig:model_skiparse_attention}.
In \textbf{Single Skip} operation, the elements located at positions $[0, k, 2k, 3k, ...]$,  $[1, k+1, 2k+1, 3k+1, ...]$, ..., $[k-1, 2k-1, 3k-1, ...]$ are bundled into a sequence, \textit{e.g.}, each token performs attention with tokens spaced $k-1$ apart. 

In \textbf{Group Skip} operation, the elements at positions $[(0, 1, ..., k-1), (k^2, k^2+1, ..., k^2+k-1), (2k^2, 2k^2+1, ..., 2k^2+k-1), ...]$, $[(k, k+1, ..., 2k-1), (k^2+k, k^2+k+1, ..., k^2+2k-1), (2k^2+k, 2k^2+k+1, ..., 2k^2+2k-1), ...]$, ..., $[(k^2-k, k^2-k-1, ..., k^2-1), (2k^2-k, 2k^2-k-1, ..., 2k^2-1), (3k^2-k, 3k^2 -k-1, ..., 3k^2-1), ...]$ are bundled as a sequence. 
Concretely, we first \textit{group adjacent tokens} in segments of length $k$, then \textit{bundle these groups} with other groups that are spaced $k-1$ groups apart into a sequence. 
For instance, in $[(0, 1, ..., k-1), (k^2, k^2+1, ..., k^2+k-1), (2k^2, 2k^2+1, ..., 2k^2+k-1), ...]$, each set of indices in parentheses represents a group, and each group is then connected with another group offset by $k-1$ groups to form one sequence. 
We notice that the main difference between the Group Skip operation and traditional Skip + Window Attention is our operation involves not only grouping but also skipping, which is ignored by previous attempts.
Concretely, Window Attention only groups adjacent tokens without connecting skipped groups into one sequence.
The distinctions among these attention methods are illustrated in Fig.~\ref{fig:attention_comparison}, with dark tokens representing the tokens involved in one attention calculation.

We further notice that the attention in 2+1D DiT corresponds to $k=HW$  (Skip operation in Group Skip has no effect when $T \ll HW$), while Full 3D DiT corresponds to $k=1$. 
In Skiparse Attention, $k$ is typically chosen to be close to 1, yet far smaller than $HW$, making the Skiparse Attention approach the effectiveness of Full 3D Attention while decreasing the computation cost.

% the Single Skip operation is intuitively, and easily understood by most. 
% Within the Group Skip operation, the Group operation is also intuitive, serving as a means to model local information. 
% However, the Group Skip operation involves not only grouping but also skipping, particularly between groups, which is often ignored. 
% This oversight frequently leads researchers to confuse Skiparse Attention with a Skip + Window Attention approach. 
% The key difference lies in even-numbered blocks: Window Attention only groups tokens without skipping between groups. 

Additionally, we propose the concept of \textbf{Average Attention Distance} ($\mathrm{AD_{avg}}$) to quantify how closely a given attention aligns with Full 3D Attention. This concept is defined as follows: 
If at least $m$ attention calculations are required to establish a connection between any two tokens A\&B, the attention distance A$\rightarrow$B is $m$ (Noticing that the attention distance between a token and itself is zero).
Thus the $\mathrm{AD_{avg}}$ for an attention mechanism is the mean of the attention distances across all token directions in input sequences, and $\mathrm{AD_{avg}}$ reflects the modeling efficiency among all tokens for the corresponding attention method.
To calculate the specific $\mathrm{AD_{avg}}$ of different attention methods, we can first identify which tokens have an attention distance of 1, and tokens with an attention distance of 2 can be determined.
Therefore, we give the $\mathrm{AD_{avg}}$ and calculation process following:

For Full 3D Attention, each token can interact with any other token in one attention calculation, resulting in the $\mathrm{AD_{avg}}=1$.

For 2+1D Attention, any two tokens can be directed with an attention distance between 1 and 2.
In the $2N$ Block, attention operates over the $(H, W)$ dimensions, where tokens within this region have an attention distance of 1. In the $2N+1$ Block, attention operates along the $T$ dimension, and attention distance is also 1 for these tokens. The total number of tokens with an attention distance of 1 is  $(HW + T - 1) - 1 = HW + T - 2$.
Therefore, $\mathrm{AD_{avg}}$ of 2+1D Attention is:
\begin{equation}
    \begin{aligned}
        \mathrm{AD_{avg}} &= \frac{1}{THW} \left[ 1 \times 0 + (HW + T - 2) \times 1 \right. \\
        & \quad \left. + \left( THW - (HW + T - 1) \right) \times 2 \right] \\
        & = 2 - \left( \frac{1}{T} + \frac{1}{HW} \right).
    \end{aligned}    
\end{equation}
For Skip + Window Attention, aside from the token itself, there are $\frac{THW}{k} - 1$ tokens with an attention distance of 1 in the $2N$ Block, and $k - 1$ tokens with an attention distance of 1 in the $2N+1$ Block. Thus, the total number of tokens with an attention distance of 1 is $\frac{THW}{k} + k - 2$.
Therefore, $\mathrm{AD_{avg}}$ of Skip + Window Attention is:
\begin{equation}
    \begin{aligned}
    \mathrm{AD_{avg}} &= \frac{1}{THW} \left[ 1 \times 0 + \left( \frac{THW}{k} + k - 2 \right) \times 1 \right.  \\
      &\quad \left. + \left( THW - \left( \frac{THW}{k} + k - 1 \right) \right) \times 2 \right]  \\
      &= 2 - \left( \frac{1}{k} + \frac{k}{THW} \right).
    \end{aligned}    
\end{equation}
In Skiparse Attention, aside from the token itself, $\frac{THW}{k} - 1$ tokens have an attention distance of 1 in the $2N$ Block, and $\frac{THW}{k} - 1$ tokens have an attention distance of 1 in the $2N+1$ Block. Notably, $\frac{THW}{k^2} - 1$ tokens can establish an attention distance of 1 in both blocks and should not be counted twice. Therefore, $\mathrm{AD_{avg}}$ in Skiparse Attention is:
\begin{equation}
    \begin{aligned}
        \mathrm{AD_{avg}} &= \frac{1}{THW} \left[ 1 \times 0 + \left( \frac{2THW}{k} - 2 - \left( \frac{THW}{k^2} - 1 \right) \right) \times 1 \right. \\
  &\quad \left. + \left( THW - \left( \frac{2THW}{k} - \frac{THW}{k^2} \right) \right) \times 2 \right]  \\
  &= 2 - \frac{2}{k} + \frac{1}{k^2} - \frac{1}{THW} \ \approx \ 2 - \frac{2}{k} + \frac{1}{k^2}.
    \end{aligned}
\end{equation}
We notice that the actual sequence length is $k\lceil \frac{THW}{k^2} \rceil$ rather than $\frac{THW}{k}$ in the Group Skip of the $2N+1$ Block. Our calculation assumes the ideal case where $k \ll THW$ and $THW\mod k=0$, yielding $k\lceil \frac{THW}{k^2} \rceil = k \cdot \frac{THW}{k^2} = \frac{THW}{k}$. In practical applications, excessively large $k$ values are typically avoided, making this derivation a reasonably accurate approximation for general usage.

% Specifically, when $k = HW$ and padding is disregarded, the Group Skip operation reduces to window attention with a window size of $HW$ because of $T \ll HW$. Given that padding does not affect the final computation, Skiparse Attention is equivalent to 2+1D Attention when $k = HW$.

For the commonly used resolution of $93 \times 512 \times 512$, using a causal VAE with a $4 \times 8 \times 8$ compression rate and a convolutional layer with a $1 \times 2 \times 2$ kernel for patch embedding, we obtain a latent shape of $24 \times 32 \times 32$ as input sequence for attention calculations. We summarize the characteristics of these attention types in Tab.~\ref{tab:four_attention_comparison}, and $\mathrm{AD_{avg}}$ for different attention methods when latent shape is $24 \times 32 \times 32$ in Tab.~\ref{tab:avg_attention_distance}.
Considering the balance between computational load and Average Attention Distance, we use Skiparse Attention with $k = 4$ in our implementations.

\begin{table*}[t]
    \setlength\tabcolsep{2pt}
    \renewcommand{\arraystretch}{1.1}
    \footnotesize
\caption{\textbf{Comparison of the different attention mechanisms.} Across multiple comparison metrics, Skiparse Attention is closer to Full 3D Attention, giving it the best spatiotemporal modeling capability apart from Full 3D Attention.}
\label{tab:four_attention_comparison}
    \begin{tabular}{l|c|c|c|c|c}
        \toprule
         \multirow{2}{*}{Attention Mechanisms} & \multirow{2}{*}{Speed} & Modeling  & \multirow{2}{*}{Global Attention} & Block & \multirow{2}{*}{Average Attention Distance} \\
         & & Capability & & Computation & \\
         \midrule
         Full 3D Attention & Slow & Strong & All blocks & Equal & 1\\
         2+1D Attention & Fast & Weak & None block & Not Equal & $2-(\frac{1}{T}+\frac{1}{HW})$ \\
        Skip + Window Attention & Middle & Weak & Half blocks & Not Equal & $2-(\frac{1}{k}+\frac{k}{THW})$ \\
        \textbf{Skiparse Attention} & Middle & Strong & All blocks & Equal & $2-\frac{2}{k}+\frac{1}{k^2}, 1<k\ll THW$ \\
        
         \bottomrule
    \end{tabular}
\end{table*}

\begin{table*}[t]
    \setlength\tabcolsep{25pt}
    \renewcommand{\arraystretch}{1}
    \centering
\caption{\textbf{The average attention distance $\mathrm{AD_{avg}}$ of different attention mechanisms.} Results are calculated when the latent shape is $24\times32\times32$.}
\label{tab:avg_attention_distance}
    \begin{tabular}{l|c}
        \toprule
         Attention Mechanisms & $\mathrm{AD_{avg}}$\\
         \midrule
         Full 3D Attention& \textbf{1.000}\\
         2+1D Attention& 1.957\\
         Skip + Window Attention ($k=2$)& 1.500\\
         Skip + Window Attention ($k=4$)& 1.750\\
         Skip + Window Attention ($k=8$)& 1.875\\
         \textbf{Skiparse Attention ($k=2$)}& \textbf{1.250}\\
         \textbf{Skiparse Attention ($k=4$)}& 1.563\\
         \textbf{Skiparse Attention ($k=8$)}& 1.766\\
         \bottomrule
    \end{tabular}
\vspace{-10pt}
\end{table*}

\subsubsection{Training Details}
\label{sec:t2v_train}
Similar to previous works~\citep{opensora,chen2024pixart,blattmann2023stable}, we use a multi-stage approach for model training. 
Starting with training an image model, our joint denoiser learns a rich understanding of static visual features, as many effective visual patterns in images also apply to videos. 
Benefiting from the 3D DiT architecture, all parameters transfer seamlessly from images to videos. 
Thus, we adopt a progressive training strategy from images to videos. 
For all training stages, we use v-prediction diffusion loss with zero terminal SNR~\citep{lin2024common}. We use min-snr weighting strategy~\citep{hang2023efficient} with $\gamma=5.0$ to accelerate the convergence process. The text encoder has a maximum input length of 512. 
We use AdamW \cite{Kingma_Ba_2014,loshchilov2019decoupledweightdecayregularization} optimizer with parameters $\beta_1 = 0.9$ and $\beta_2 = 0.999$. Details of leveraged datasets in training stages are shown in Sec.~\ref{sec:data}

\noindent\textbf{Text-to-Image Pretraining.} The objective of this stage is to learn a visual prior that enables fast convergence when training on videos, reducing dependency on large-scale video datasets. Since the weights of Full 3D Attention can efficiently transfer to Skiparse Attention, we first train a Full 3D Attention model on $256\times256$ images to generate text-conditioned images, for approximately 150k steps. We then inherit the model weights and replace Full 3D Attention with Skiparse Attention, allowing tuning from a 3D dense attention model to a sparse attention model. The tuning process involves around 100k steps, a batch size of 1024, and a learning rate of 2e-5.
Image datasets includes SAM, Anytext, and Human-images.

\noindent\textbf{Text-to-Image\&Video Pretraining.} We jointly train on images and videos, with a maximum shape of $93\times640\times640$. The pretraining process includes approximately 200k steps, a batch size of 1024, and a learning rate of 2e-5. Image data consists almost entirely of SAM from version 1.2.0, and the leveraged video dataset is the original Panda70M.

\noindent\textbf{Text-to-Video Fine-tuning.} The model nearly converges around 100k steps, with no substantial gains observed by 200k steps. Following the procedures in Sec.~\ref{sec:data}, we refine the data by cleaning and re-captioning. Fine-tuning is conducted with the filtered Panda70M and additional high-quality data at a fixed resolution of $93\times352\times640$. This process runs for 30k steps with a learning rate of 1e-5, utilizing 256 NPUs/GPUs with a total batch size of 1024.

\begin{figure*}[t]
\includegraphics[width=\linewidth]{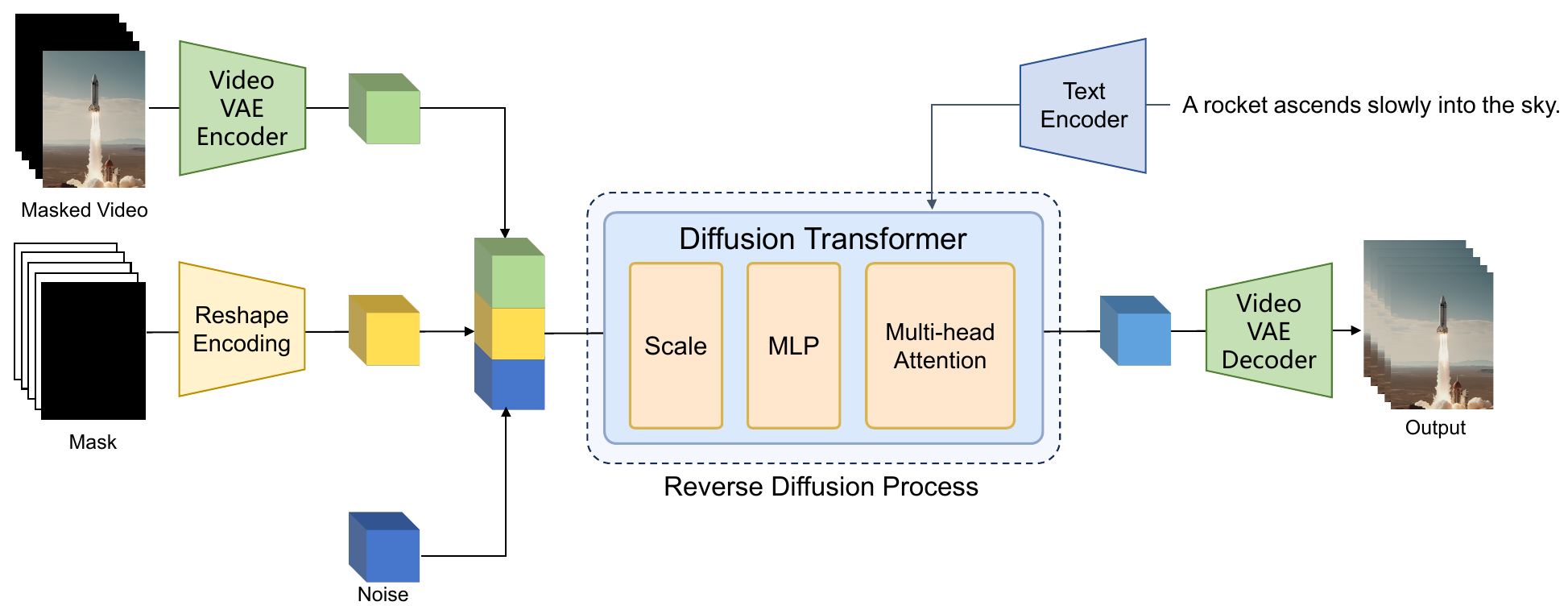}
\caption{\textbf{Overview of our Image Condition Controller.} Our Controller unifies multiple image conditional tasks including image-to-video, video transition, and video continuation in one framework when giving masks are changed.}
\label{fig:opensora_inpaint}
\end{figure*}

% \begin{figure}[t]
%     \centering
%     \includegraphics[width=\linewidth]{fig/opensora_inpaint.pdf}
%     \caption{\textbf{Overview of our Inpainting Model.} When the first frame is presented while the remaining frames are masked, the model performs Image-to-Video generation.}
%     \label{fig:opensora_inpaint}
% \end{figure}
% \begin{figure}[t]
%     \centering
%     \includegraphics[width=\linewidth]{fig/inpaint_type.pdf}
%     \caption{\textbf{Different types of masks in the inpainting model.} Each mask type corresponds to a specific inpainting task in the frame dimension. Black frames indicate they are retained, while white frames indicate they are masked.}
%     \label{fig:inpaint_type}
% \end{figure}

\begin{figure*}[t]
\centering
\includegraphics[width=1.0\textwidth]{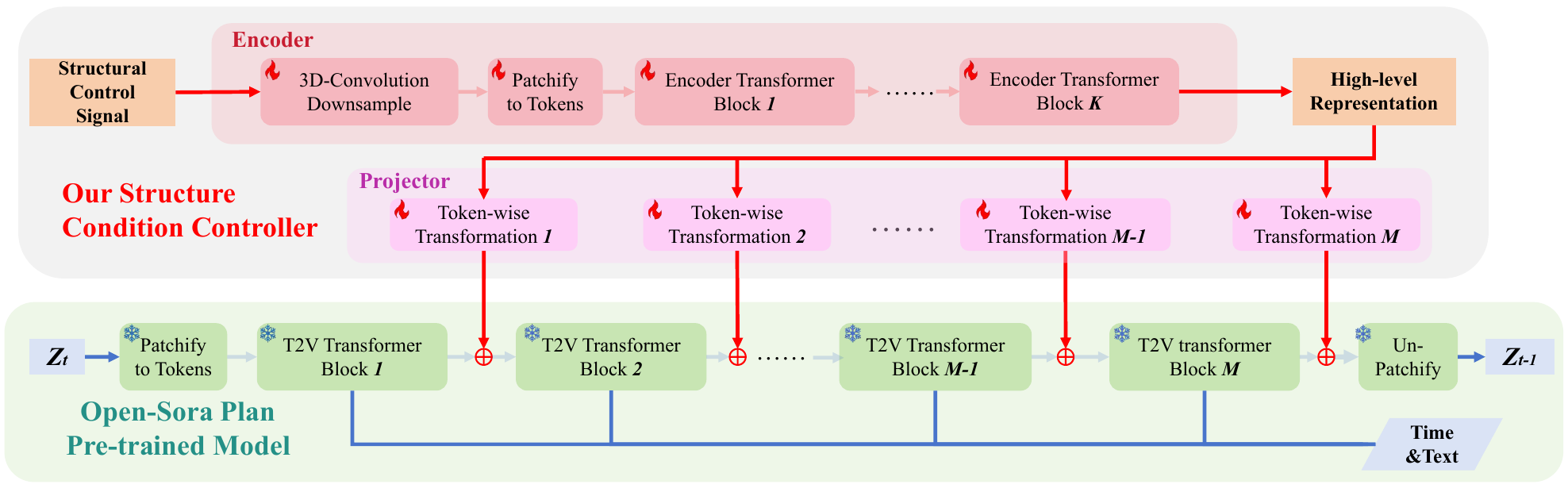}
\caption{\textbf{Overview of our Structure Condition Controller.}
The structure Controller contains two light components including an encoder that focuses on extracting a high-level representation from the structural signals and a projector that transforms such representation into injection features.
Finally, we directly add obtained injection features to the pre-trained model for structure control.}
\label{control_pipline}
\end{figure*}

\subsection{Conditional Controllers}
\label{sec:x2v_model}
% We present the image-to-video architecture and training details in Sec.~\ref{sec:i2v_model}, with qualitative results shown in Sec.~\ref{sec:res_i2v}. Additionally, we introduce a structured-conditional-to-video architecture, detailed in Sec.~\ref{Controllner sec:5}, with visualizations provided in Sec.~\ref{sec:res_x2v}.

\subsubsection{Image Condition Controller} 
\label{sec:i2v_model}
Inspired by Stable Diffusion Inpainting~\citep{stable_diffusion}, we regard the image conditional tasks as an inpainting task in the temporal dimension for a more flexible training paradigm. 

The image condition model is initialized by our text-to-video weights. As shown in Fig.~\ref{fig:opensora_inpaint}, it adds two additional inputs including \textbf{given mask} and \textbf{masked video}, which are concatenated with the latent noise and then fed into the Denoiser. 
For the given mask, instead of employing VAE for encoding, we adopt the ``reshape'' operation to align latent dimensions due to the temporal down-sampling in VAE will damage the control accuracy of masks.
For the masked video, we multiply the original video by the given mask and then input the multiplied video into VAE for encoding.

Unlike previous works based on 2+1D Attention, which inject semantic features of images (usually extracted via CLIP~\citep{clip}) into the UNet or DiT to enhance cross-frame stability~\citep{blattmann2023stable, dynamicrafter, easyanimate}, we simply alter the input channels of the DiT without incorporating semantic features for control. 
% With Skiparse Attention, pixels at any spatial and temporal position can connect with certain pixels in conditional images in a single attention operation, sufficiently ensuring motion consistency.
We observe that leveraging various semantic injection methods can not noticeably improve the generated results while instead limiting the range of motion, thus we discard the image semantic injection module in our experiments.

\begin{wrapfigure}{l}{0.4\linewidth}
\vspace{-3pt}
\includegraphics[width=\linewidth]{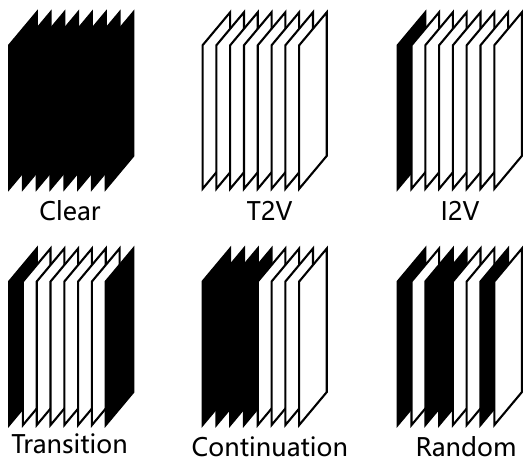}
\caption{\textbf{Different types of masks for image-conditioned generation.} Black masks indicate corresponding frames are retained, while white masks indicate frames are masked.}
\label{fig:inpaint_type}
\vspace{-15pt}
\end{wrapfigure}

\noindent\textbf{Training Details.} For training configuration, we adopt the same settings as the text-to-video model, including v-prediction, zero terminal SNR, and min-snr weighting strategy, with parameters consistent with the text-to-video model. We also use the AdamW optimizer with a constant learning rate of 1e-5 and utilize 256 NPUs a batch size fixed at 512.

Thanks to the flexibility of different mask types in our inpainting framework, we design a progressive training strategy that gradually increases the difficulty of training tasks as shown in Fig.~\ref{fig:inpaint_type}, which strategy can lead to smoother training curves and improve motion consistency.
The masks used during training are set as follows:
(\textbf{1}) \textbf{Clear:} Retain all frames.
(\textbf{2}) \textbf{T2V:} Discard all frames.
(\textbf{3}) \textbf{I2V:} Retain only the first frame but discard the rest.
(\textbf{4}) \textbf{Transition:} Retain only the first and last frames but discard the rest.
(\textbf{5}) \textbf{Continuation:} Retain the first $n$ frames but discard the rest.
(\textbf{6}) \textbf{Random:} Retain $n$ randomly selected frames but discard the rest.
Concretely, Our progressive training strategy includes two stages. In Stage 1, we train on multiple simple tasks at a low resolution. In Stage 2, we train the image-to-video and video transition tasks at a higher resolution.

\noindent\textbf{Stage 1:} Any resolution and duration within $93 \times 102400$ ($320 \times 320$), using unfiltered motion and aesthetic low-quality data. The task ratios at different steps are as follows:
\begin{enumerate}
    \item T2V 10\%, Continuation 40\%, Random 40\%, Clear 10\%. Ensure that at least 50\% of the frames are retained during continuation and random mask, training with 4 million samples.
    \item T2V 10\%, Continuation 40\%, Random 40\%, Clear 10\%. Ensure that at least 25\% of the frames are retained during continuation and random mask, training with 4 million samples.
    \item T2V 10\%, Continuation 40\%, Random 40\%, Clear 10\%. Ensure that at least 12.5\% of the frames are retained during continuation and random mask, training with 4 million samples.
    \item T2V 10\%, Continuation 25\%, Random 60\%, Clear 5\%. Ensure that at least 12.5\% of the frames are retained during continuation and random mask, training with 4 million samples.
    \item T2V 10\%, Continuation 25\%, Random 60\%, Clear 5\%, training with 8 million samples.
    \item T2V 10\%, Continuation 10\%, Random 20\%, I2V 40\%, Transition 20\%, training with 16 million samples.
    \item T2V 5\%, Continuation 5\%, Random 10\%, I2V 40\%, Transition 40\%, training with 10 million samples.
\end{enumerate}

\noindent\textbf{Stage 2:} Any resolution and duration within $93 \times 236544$ (e.g., $480 \times 480$, $640 \times 352$, $352 \times 640$), using filtered motion and aesthetic high-quality data, ratios of different tasks are
T2V 5\%, Continuation 5\%, Random 10\%, I2V 40\%, Transition 40\%, training with 15 million samples.

After completing the two-stage training, we draw on the approach mentioned in \cite{yang2024cogvideox}, adding slight Gaussian noise to the conditional images to enhance generalization during fine-tuning, with utilizing 5 million filtered motion and aesthetic high-quality data. 

\subsubsection{Structure Condition Controller}
\label{Controllner sec:5}

When imposing structural control on our retained text-to-image model, an intuitive idea is to use previous control methods \cite{controlnet, t2iadapter, controlnet_plus_plus, sparsectrl} specified for the U-net-based base models.
However, most of these methods are based on ControlNet \cite{controlnet}, which copies half of the base model to process the control signals and will increase the hardware consumption by nearly 50\%.
The additional consumption is immense, as the original expense of our Open-Sora Plan base model is already extremely high.
Although some works \cite{t2iadapter, controlnext} try to replace the heavy copy of the base model with a lighter network at the sacrifice of controllability, these will probably lead to poor alignment with the input structural signals and the generated video when used for our base model.

To more efficiently add structural control to our base model, we propose a novel Structure Condition Controller, as shown in Fig.~\ref{control_pipline}.
Specifically, we suppose the denoiser of our base model contains $M$ transformer blocks.
For the $j$-th $1\leq j\leq M$ transformer block $\mathcal{T}_j$ in the base model, its output is a series of tokens $\boldsymbol{X}_j$, which can be expressed as:
\begin{equation}
\boldsymbol{X}_{j}=\mathcal{T}_j(\boldsymbol{X}_{j-1}).
\end{equation}
Given a structural signal $\boldsymbol{C}_S$, the encoder $\mathcal{E}$ extracts the high-level representation $\boldsymbol{R}$ from $\boldsymbol{C}_S$:
\begin{equation}
\label{Eq8}
    \boldsymbol{R}=\mathcal{E}(\boldsymbol{C}_S).
\end{equation}
Then, the projector $\mathcal{P}$, containing $M$ transformations with the same process, transforms $\boldsymbol{R}$ into the injection feature $\boldsymbol{F}$, including $M$ elements, which can be expressed as:
\begin{equation}
    \mathcal{P}=[\mathcal{P}_1,\mathcal{P}_2,...\mathcal{P}_M],
\end{equation}
\begin{equation}
    \boldsymbol{F}=[\boldsymbol{F}_1,\boldsymbol{F}_2,..., \boldsymbol{F}_M],
\end{equation}
\begin{equation}
    \boldsymbol{F}_j=\mathcal{P}_j(\boldsymbol{R}).
\end{equation}
Here $\mathcal{P}_j$ denotes the $j$ transformation of $\mathcal{P}$ that transform $\boldsymbol{R}$ to $\boldsymbol{F}_j$, the $j$-th element of $\boldsymbol{F}$.
To impose structural control on the base model, we can directly add $\boldsymbol{F}_j$ to $\boldsymbol{X}_j$:
\begin{equation}
\label{Eq12}
    \boldsymbol{X}_{j} = \boldsymbol{X}_j + \boldsymbol{F}_j.
\end{equation}
To satisfy the above equation, we should ensure the shape of $\boldsymbol{F}_j$ equals $\boldsymbol{X}_j$.
To achieve this, we use the following design of our encoder $\mathcal{E}$ and projector $\mathcal{P}$.
Specifically, in the encoder $\mathcal{E} $, we first downsample $C_S$ to make its shape the same as $\boldsymbol{Z_t}$ with a tiny 3D convolution-based network.
Then, we flatten $C_S$ to tokens with the same shape as $X_j(1\leq j\leq M)$.
After that, to obtain $\boldsymbol{R}$, these tokens are processed by $K$ transformer blocks, which maintain the token's shape.
For the projector $\mathcal{P}$, we only need to promise $\mathcal{P}_j$ will not change the token shape of $\boldsymbol{R}$.
Thus, we design $\mathcal{P}_j$ as a token-wise transformation with the same input and output shape, such as a linear FC-layer or two-layer MLP, which is efficient and can maintain the token shape.

\noindent\textbf{Training Details.}
We utilize the Panda70M dataset to train our Structure Controller.
Given a video clip, we use the specified signal extractors to extract the corresponding structural control signals.
Specifically, we extract the canny, depth, and sketch, by canny detector \cite{canny}, Midas \cite{midas}, and PiDiNet \cite{pidinet}, respectively.
We train our Structure Controller for 20k steps, on 8 NPUs/GPUs, with a total batch size of 16, and a learning rate of 4e-6. 

%% file: sec/5_strategy.tex
\section{Assistant Strategies} 
\label{sec:train_st}

\subsection{Min-Max Token Strategy}
\label{sec:mmtt}

To achieve efficient processing on hardware, deep neural networks are typically trained with batched inputs, meaning the shape of the training data is fixed. Traditional methods adopt two approaches including resizing images or padding images to a fixed size. However, both approaches have drawbacks, \textit{e.g.}, the former loses useful information, while the latter has low computational efficiency. 
Generally, there are three methods for training with variable token counts: Patch n' Pack~\citep{dehghani2024patch,yang2024cogvideox}, Bucket Sampler~\citep{chen2023pixartalpha,chen2024pixart,opensora}, and Pad-Mask~\citep{Lu2024FiT,wang2024fitv2}.

\noindent\textbf{Patch n' Pack.} By packing multiple samples, this method addresses the fixed sequence length limitation. Patch n' Pack defines a new maximum length, and tokens from multiple data instances are packed into this new data. As a result, the original data is preserved while enabling training with arbitrary resolutions. However, this method introduces significant intrusion into the model code, making it difficult to adapt in fields where the model architecture is not yet stable.

\noindent\textbf{Bucket Sampler.} This method packs data of different resolutions into buckets and samples batches from the buckets to ensure all data in a batch have the same resolution. It incurs minimal intrusion into the model code, primarily requiring modifications to the data sampling strategy.

\noindent\textbf{Pad-Mask.} This method sets a maximum resolution, pads all data to this resolution, and generates a corresponding mask to exclude loss from the masked areas. While conceptually simple, it has low computational efficiency.

We believe current video generation models are still in an exploratory phase. Patch n' Pack incurs significant intrusion into the model code, leading to unnecessary development costs. Pad-mask has low computational efficiency, which wastes resources in dense computations like video. The bucket strategy, while requiring no changes to the model code, leads to greater loss oscillation as token count variation increases (with more resolution types), indicating higher training instability. 
Given a maximum token $m$, resolution stride $s$, and a set of possible resolution ratios $\mathcal{R}=\left\{\left(r_1^h, r_1^w\right),\left(r_2^h, r_2^w\right), \ldots,\left(r_n^h, r_n^w\right)\right\}$, we propose the \textbf{Min-Max Token} strategy for tacking mentioned issues.
We notice that $s=8\times2$ is the multiples of spatial downsampling rate in VAE and convolution stride in denoiser, and there are five common resolutions: $\frac{1}{1}$, $\frac{3}{4}$, $\frac{4}{3}$, $\frac{9}{16}$ and $\frac{16}{9}$ in practical needs. 
For each ratio $\left(r_i^h, r_i^w\right)$ in $\mathcal{R}$, 
$r_i^h$ and $r_i^w$ are required to be \textbf{coprime positive integers}. The height $h$ and width $w$ are defined as $h=r_i^h \cdot k \cdot s$ and $w=r_i^w \cdot k \cdot s$, where is the scaling factor $k$ to be determined. The total token count $n$ satisfies the constraint $n=h \cdot w \leq m$. Substituting the expressions for $h$ and $w$, we get:
\begin{equation}
n_i=\left(r_i^h \cdot k \cdot s\right) \cdot\left(r_i^w \cdot k \cdot s\right)=r_i^h \cdot r_i^w \cdot k^2 \cdot s^2,
\end{equation}
so the constraint becomes:
\begin{equation}
r_i^h \cdot r_i^w \cdot k^2 \cdot s^2 \leq m .
\end{equation}
Taking the square root of both sides, to ensure 
$k$ is an integer, we obtain the upper bound result for $k$:
\begin{equation}
k_i=\left\lfloor\sqrt{\frac{m}{r_i^h \cdot r_i^w \cdot s^2}}\right\rfloor .
\end{equation}
The set of minimum token $n$ is then expressed as:
\begin{equation}
n=\min \left(\left\{r_i^h \cdot r_i^w \cdot k_i^2 \cdot s^2 \mid\left(r_i^h, r_i^w\right) \in \mathcal{R}\right\}\right) .
\end{equation}
For example, the max token $m$ is typically set as a square rootable number, such as 65536 ($256 \times 256$), as it reliably supports a 1:1 aspect ratio. Given this, we configure $s=16$, and aspect ratios of 3:4 and 9:16. The resulting min token $n$ is 36864 ($144\times256$).

As discussed above, we implement the Min-Max Token Training combined with the Bucket Sampler using a custom data sampler to maintain a consistent token count per global batch, though token counts vary across global batches. This approach allows NPUs/GPUs to maintain nearly identical compute times, reducing synchronization overhead. The method fully decouples data sampling code from model code, providing a plug-and-play sampling strategy for multi-resolution, multi-frame data.

\subsection{Adaptive Gradient Clipping Strategy}
\label{sec:ema_gn}

% During model training, we observed the loss spike phenomenon mentioned in the Playground paper. Although these spikes don't lead to NaN values and quickly return to normal loss, they significantly affect the output image quality. The spikes are caused by issues in multi-node, multi-NPU/GPU communication, leading to abnormally large gradient norms on certain NPUs/GPUs. Therefore, this issue occurs more frequently in large-scale models requiring more training resources. To address this, Playground uses a pre-defined threshold: if the gradient norm exceeds this value, the current training iteration is discarded. However, since gradient norms decrease as training progresses, a fixed threshold is less effective for spike detection in later stages of training. To improve this, we track the exponential moving average (EMA) of the max gradient norm as $mgn_{ema}$ and the EMA of the standard deviation of the variance $mgn - mgn_{ema}$ as $mgnstd_{ema}$. When $|mgn - mgn_{ema}| > 3*mgnstd_{ema}$, we flag it as an anomaly. Additionally, we found that spikes usually affect only one NPU/GPU, so instead of discarding the entire iteration, we only discard the batch from the affected rank, minimizing the loss of valid batches.

\begin{wrapfigure}{r}{0.48\linewidth}
\vspace{-20pt}
\centering
\includegraphics[width=\linewidth]{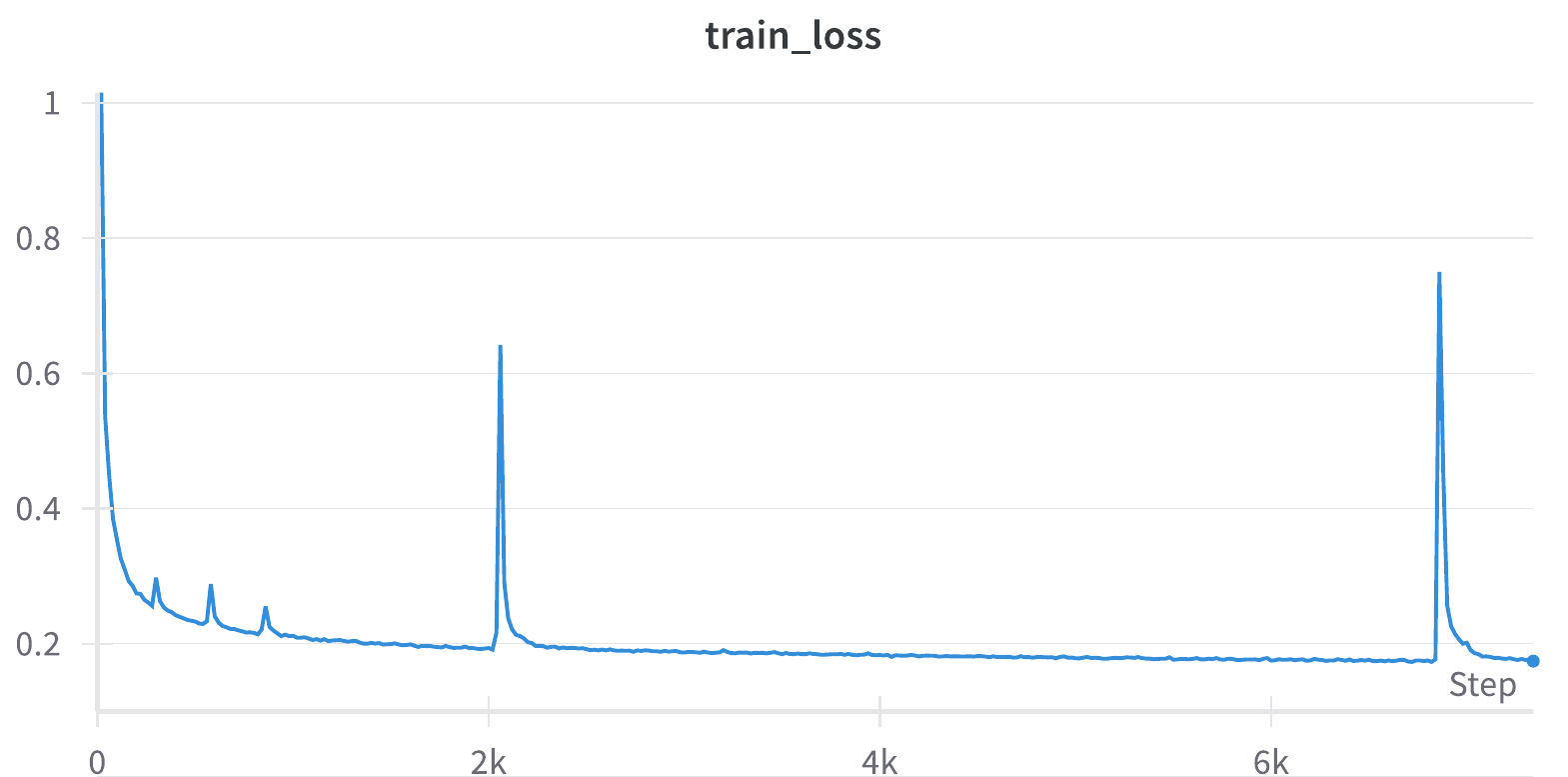}
\caption{\textbf{Plot of spikes in training loss.} We observe loss spikes during training that could not be reproduced with a fixed seed.}
\label{fig:loss_spike}
\vspace{-10pt}
\end{wrapfigure}

In distributed model training, we often observe loss spikes as shown in Fig.~\ref{fig:loss_spike}, significantly degrade output quality without causing NaN errors. 
Unlike typical NaN errors that disrupt training, these spikes temporarily increase loss values and are followed by a return to normal levels, which occur sporadically and adversely impact model performance. 
These spikes arise due to various issues, including abnormal outputs from the VAE encoder, desynchronization in multi-node communication, or outliers in training data leading to large gradient norms.

\begin{figure}[t!]
    \begin{minipage}{0.48\textwidth}
        \includegraphics[width=\linewidth]{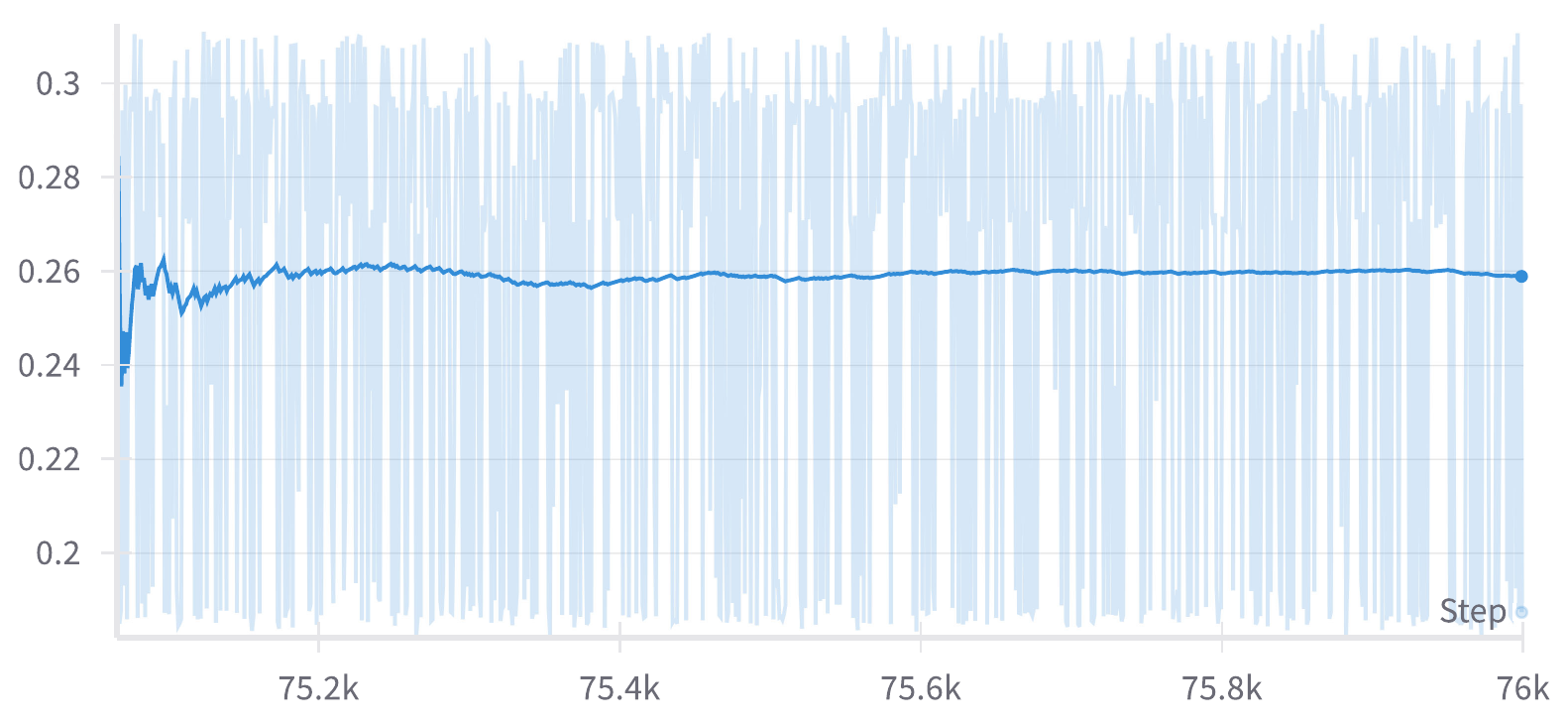}
        \caption*{(a)}
    \end{minipage}
    \hfill
    \begin{minipage}{0.48\textwidth}
        \includegraphics[width=\linewidth]{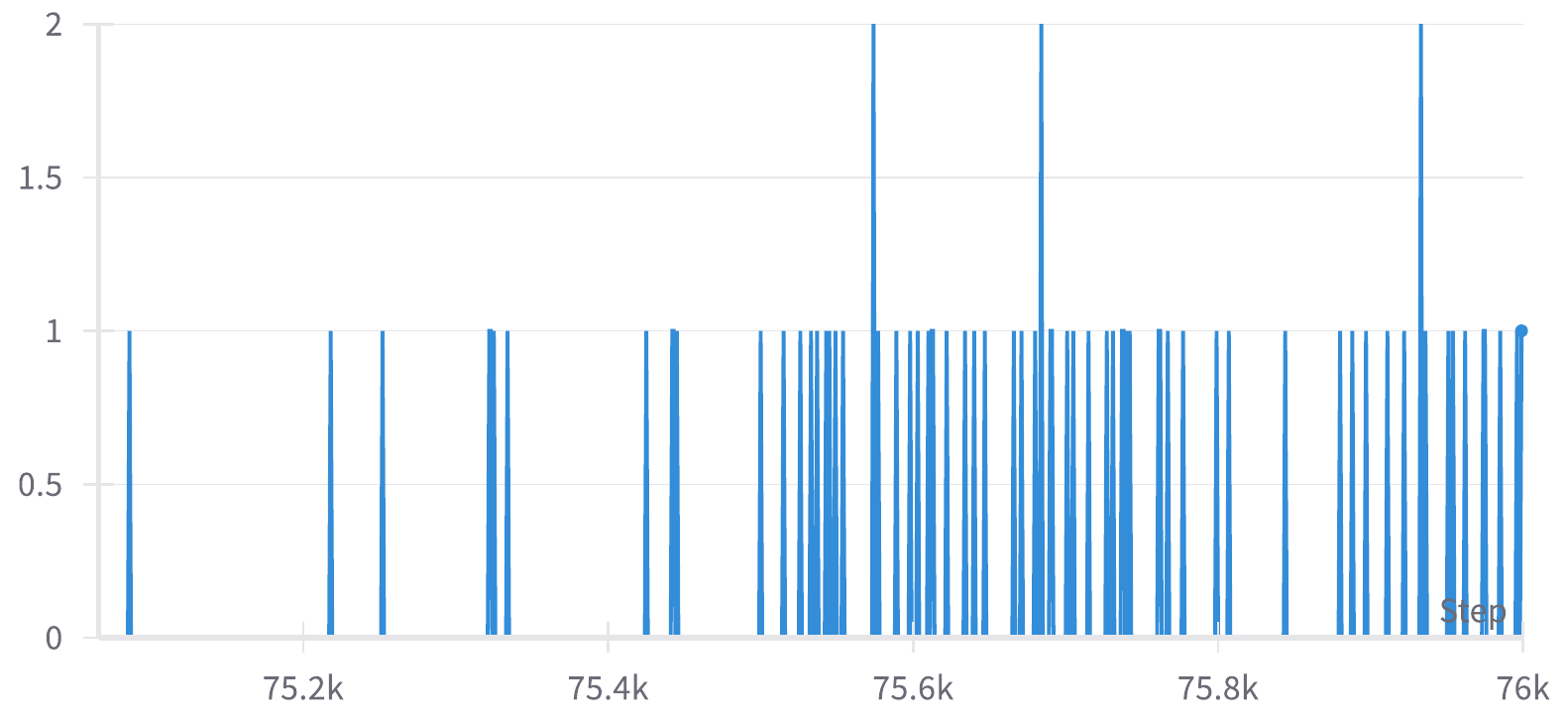}
        \caption*{(b)}
        % \label{fig:num_zero_grad}
    \end{minipage}
    \begin{minipage}{0.48\textwidth}
        \includegraphics[width=\linewidth]{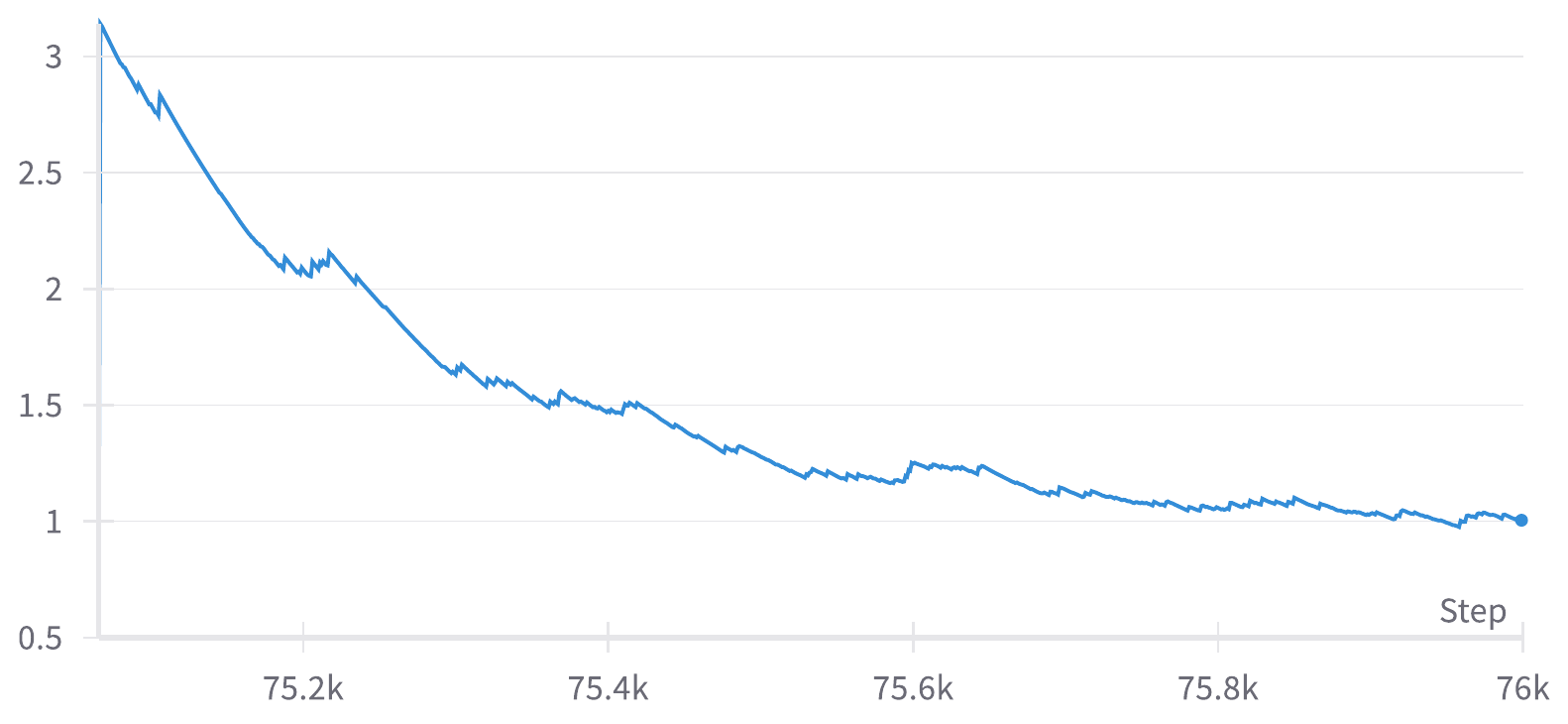}
        \caption*{(c)}
        % \label{fig:max_norm}
    \end{minipage}
    \hfill
    \begin{minipage}{0.48\textwidth}
        \includegraphics[width=\linewidth]{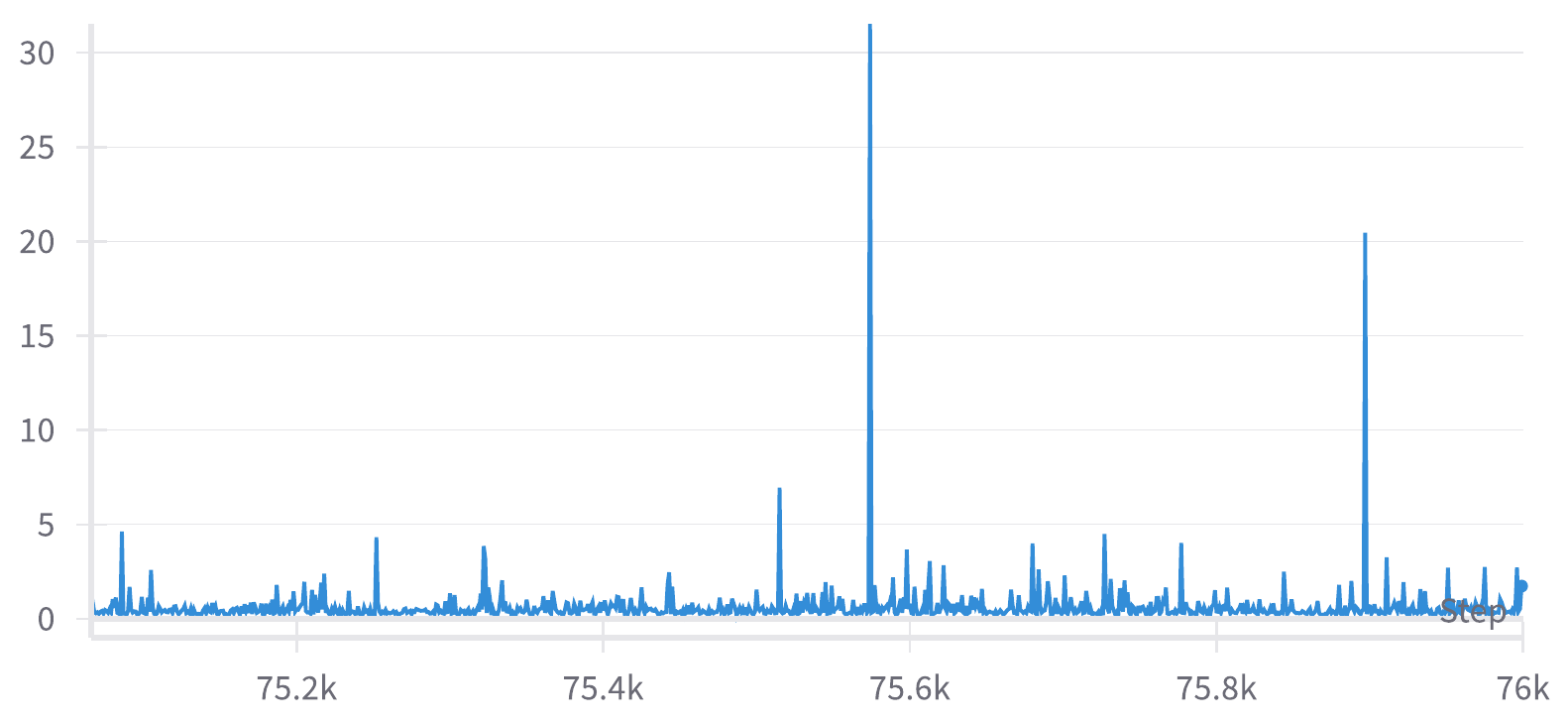}
        \caption*{(d)}
        % \label{fig:max_grad_norm}
    \end{minipage}
    \begin{minipage}{0.48\textwidth}
        \includegraphics[width=\linewidth]{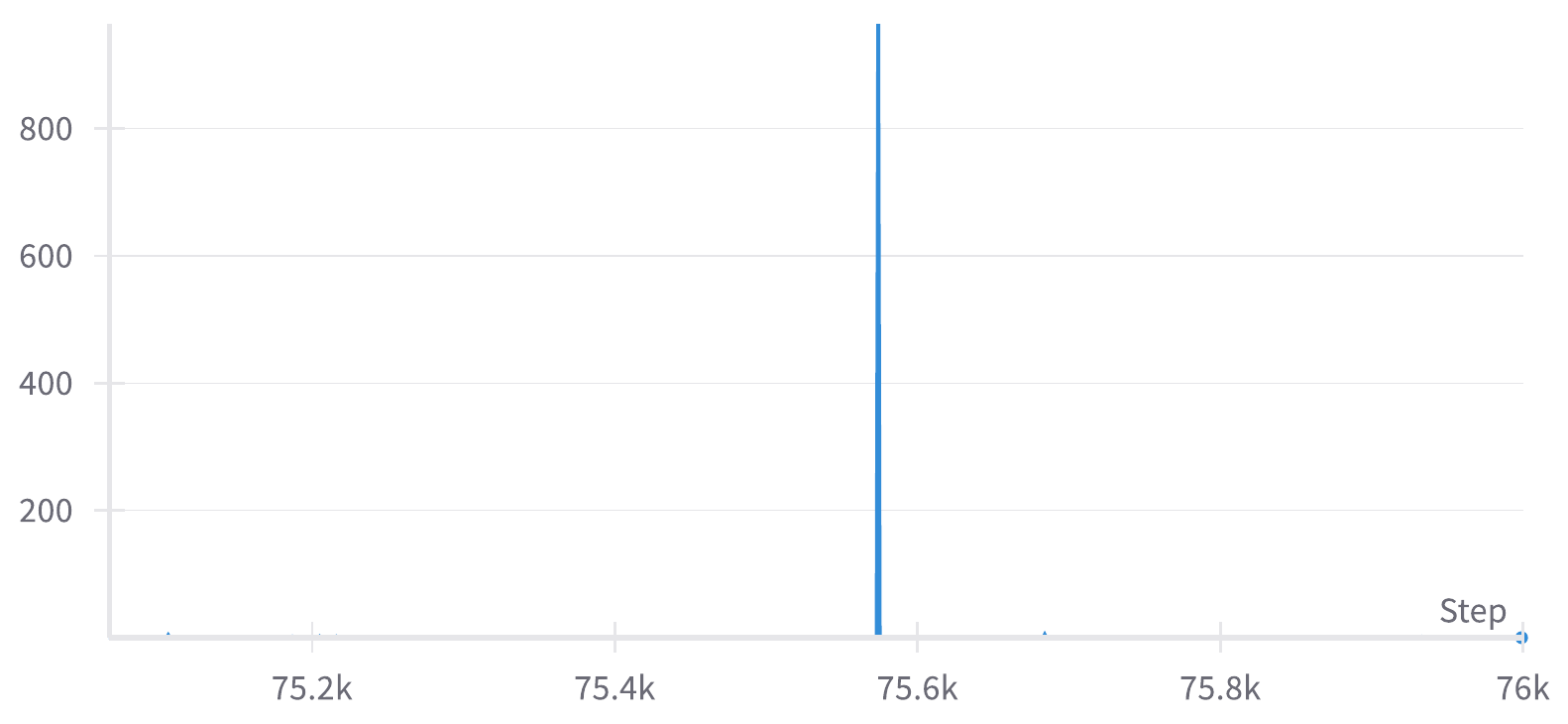}
        \caption*{(e)}
        % \label{fig:max_grad_norm_var}
    \end{minipage}
    \hfill
    \begin{minipage}{0.48\textwidth}
        \includegraphics[width=\linewidth]{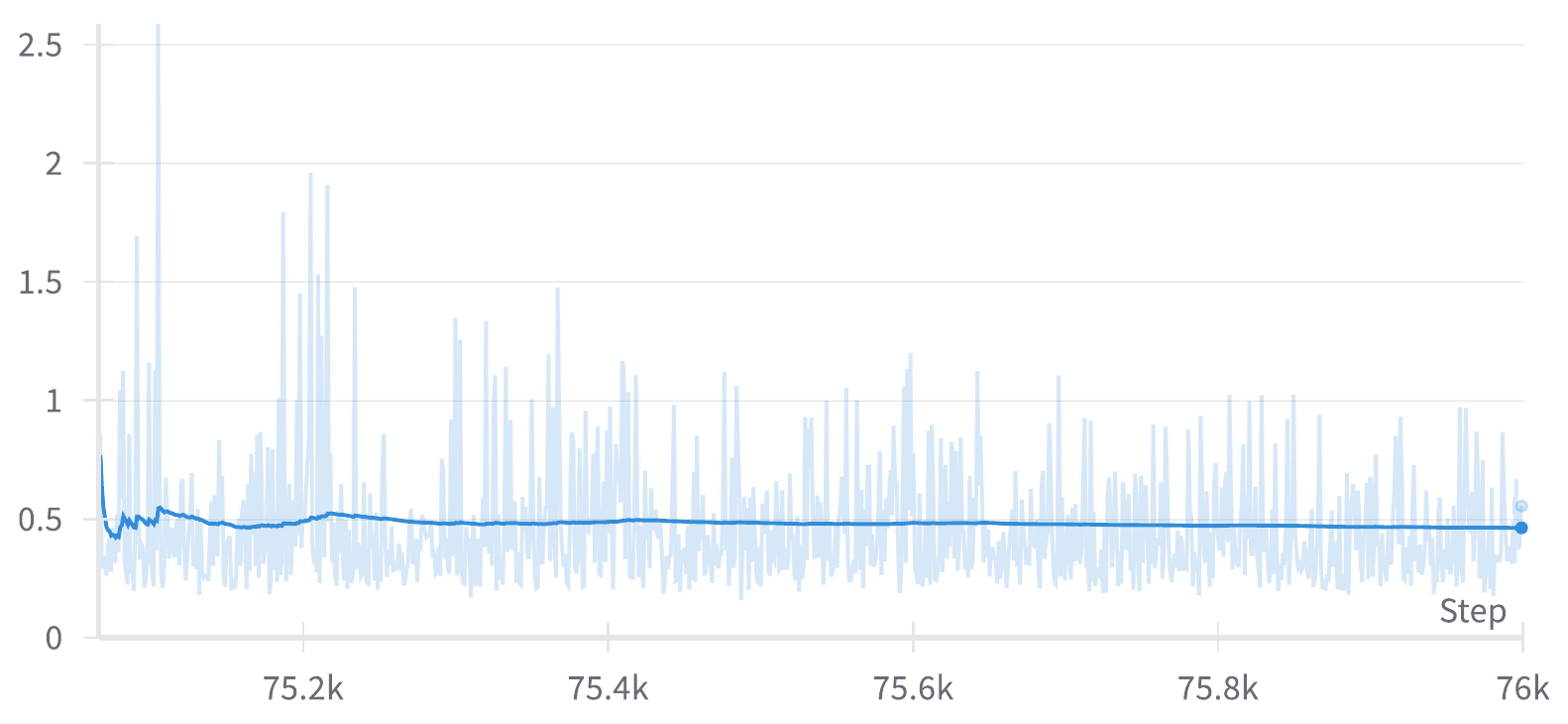}
        \caption*{(f)}
        % \label{fig:max_grad_norm_clip}
    \end{minipage}
    \begin{minipage}{0.48\textwidth}
        \includegraphics[width=\linewidth]{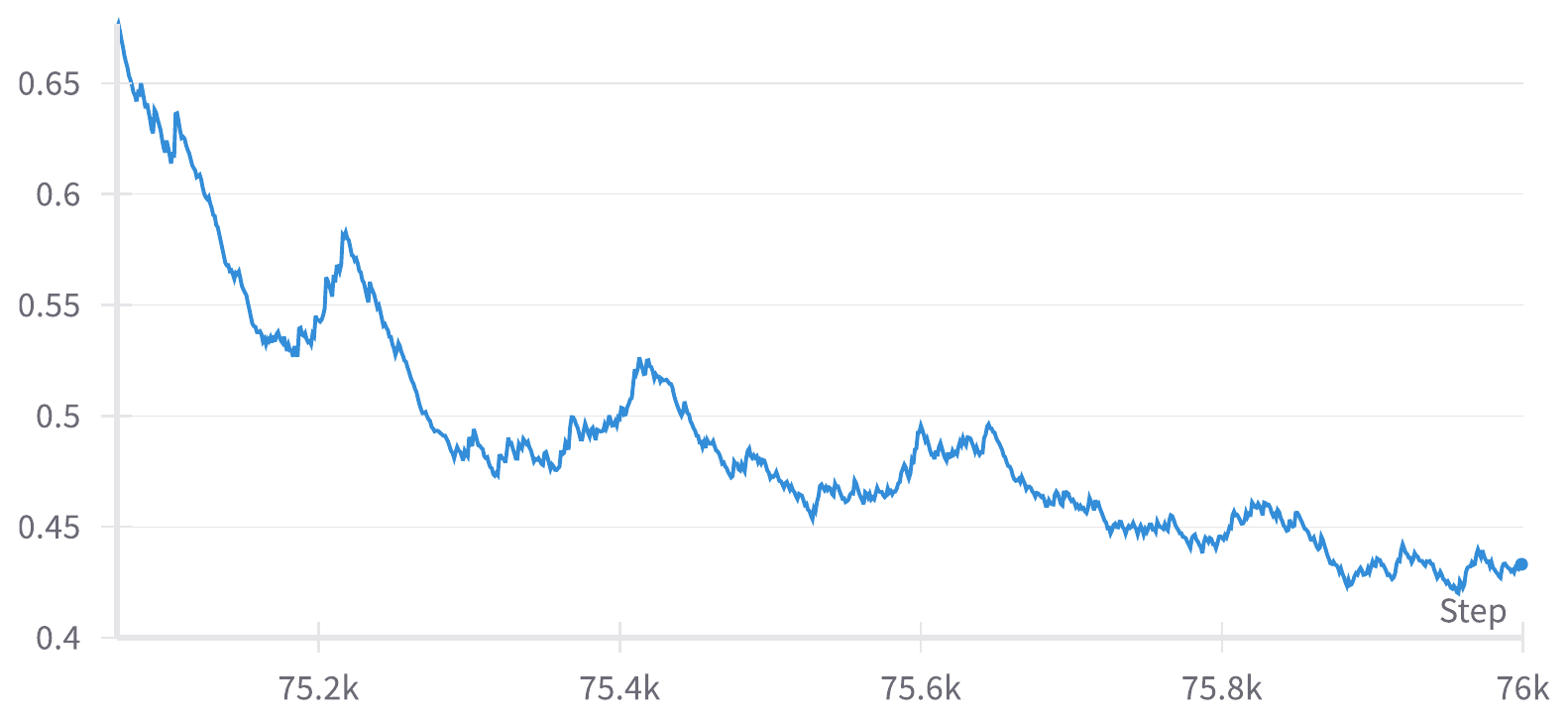}
        \caption*{(g)}
        % \label{fig:moving_avg_max_grad_norm}
    \end{minipage}
    \hfill
    \begin{minipage}{0.48\textwidth}
        \includegraphics[width=\linewidth]{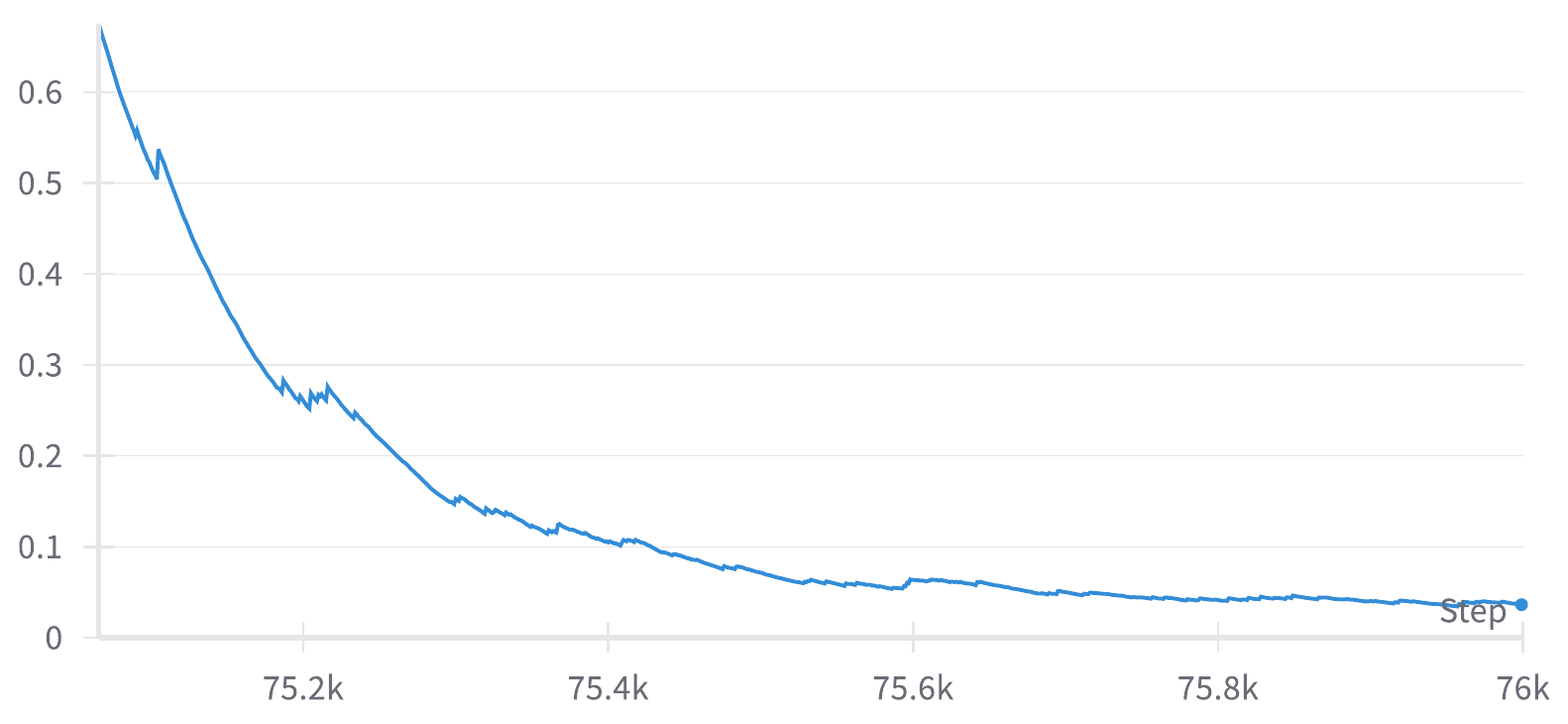}
        \caption*{(h)}
        % \label{fig:moving_avg_max_grad_norm_var}
    \end{minipage}

    \caption{\textbf{Logging abnormal iterations during training.} We resume training at step 75k and display logs from step 75k to 76k, noting an anomaly around step 75.6k. \textbf{(a)} Diffusion model loss during training. \textbf{(b)} Abnormal local batches discarded per step. \textbf{(c)} Gradient norm upper bound plotted based on a 3-sigma criterion. \textbf{(d)} Maximum gradient norm among all local batches. \textbf{(e)} Variance of the maximum gradient norm. Note that most steps involve values close to 0. \textbf{(f)} Maximum value of all processed gradient norms. \textbf{(g)} EMA of the maximum gradient norm. \textbf{(h)} EMA of the variance of the maximum gradient norm.}
    \label{fig:logging}
    \vspace{-10pt}
\end{figure}
% (d) and (e) reflect the state before handling anomalies; 
% (f), (g) and (h) show normal maximum gradient norm statistics, where we do not update the moving average with abnormal values.

We attempt many methods including applying gradient clipping, adjusting the $\beta_2$ in optimizer, and reducing the learning rate, but none of these approaches resolve the issue, which appears randomly and cannot be reproduced even with a fixed seed. Playground v3~\citep{liu2024playground} encounters the same issue and involves discarding an iteration if the gradient norm exceeds a fixed threshold. However, fixed thresholds may fail to adapt to decreasing gradient norms as training progresses. Therefore, we introduce an adaptive thresholding mechanism that leverages Exponential Moving Averages (EMA) for effective anomaly detection. Our approach mitigates the effects of spikes while preserving training stability and output quality.

Let $\mathrm{gn}_i$ denote the gradient norm on NPU/GPU$_i$ for $i=1,2, \ldots, N$, where $N$ is the total number of NPUs/GPUs. We define the maximum gradient norm across all NPUs/GPUs as:
\begin{equation}
\mathrm{gn}_{\max }=\max _{i=1}^N \mathrm{gn}_i.
\end{equation}
To ensure the threshold adapts to the training dynamics, we use the EMA of the maximum gradient norm $\mathrm{ema}_{\mathrm{gn}}$ and its variance-based EMA $\mathrm{ema}_{\mathrm{var}}$, which updated as follows:
\begin{align}
\mathrm{ema}_{\mathrm{gn}} &= \alpha \cdot \mathrm{ema}_{\mathrm{gn}} + (1-\alpha) \cdot \mathrm{gn}_{\max }, \\
\mathrm{ema}_{\mathrm{var}} &= \alpha \cdot \mathrm{ema}_{\mathrm{var}}+(1-\alpha) \cdot \left(\mathrm{gn}_{\max }-\mathrm{ema}_{\mathrm{gn}}\right)^2,
\end{align}
where $\alpha$ is the update rate for EMA, we set it to 0.99. We can record whether each gradient norm is abnormal based on the 3-sigma rule, denoted as $\delta_i$:
\begin{equation}
\delta_i= \begin{cases}0, & \text { if }\mathrm{gn}_i-\mathrm{ema}_{\mathrm{gn}} > 3 \cdot \sqrt{\mathrm{ema}_{\mathrm{var}}} \\ 1, & \text { otherwise }\end{cases}.
\end{equation}
Then, the number of normal gradient norm $M$ can be obtained by summing the indicator functions of all NPUs/GPUs:
\begin{equation}
M=\sum_{i=1}^N \delta_i.
\end{equation}
For each NPU/GPU, we define the final gradient update rule based on the detection result. If an anomaly is detected for NPU/GPU$_i$, the gradient for that NPU/GPU is set to zero, or it will be multiplied by $\frac{N}{M}$ otherwise:
\begin{equation}
g_i^{\text {final }}= \begin{cases}0, & \text { if }\mathrm{gn}_i-\mathrm{ema}_{\mathrm{gn}}>3 \cdot \sqrt{\mathrm{ema}_{\mathrm{var}}} \\   \frac{N}{M} \cdot g_i, & \text { otherwise }\end{cases}.
\end{equation}
After adjusting the gradients, we apply an all-reduce operation across NPUs/GPUs to synchronize the remaining non-zero gradients. In Fig.~\ref{fig:logging}, we illustrate how the moving average gradient norm addresses abnormal data. Fig.~\ref{fig:logging} (d) and Fig.~\ref{fig:logging} (e) show a sudden increase in gradient norm on a specific NPU/GPU near step 75.6k, exceeding the moving average of the maximum gradient norm (seen in Fig.~\ref{fig:logging} (c)). Consequently, the gradient for this local batch is set to zero (logged in Fig.~\ref{fig:logging} (b)). We also record the post-discard maximum gradient to confirm successful handling. Finally, the processed maximum gradient norm (logged in Fig.~\ref{fig:logging} (f)) updates the moving average of the maximum gradient norm and its variance in Fig.~\ref{fig:logging} (g) and Fig.~\ref{fig:logging} (h). As shown in Fig.~\ref{fig:logging} (a), the training loss remains stable without spikes, demonstrating that this approach effectively prevents anomalous batches from affecting the training process without discarding entire iterations.

\subsection{Prompt Refiner}
\label{sec:refiner}
The training dataset for the video generation model is annotated by Vision Language Models~\citep{chen2024far, wang2024qwen2}, providing highly detailed descriptions of scenes and themes, with most annotations consisting of lengthy texts that differ substantially from typical user input. User input is generally less detailed and concise, containing fewer words (\textit{e.g.}, in VBench~\citep{vbench}, most test texts contain fewer than 30 words, sometimes no more than 5 words). This discrepancy results in a significant gap compared to the textual conditions used in model training, leading to reduced video quality, semantic fidelity, and motion amplitude. To address this gap and enhance the model performance when facing shorter texts, we introduce an LLM to leverage its text expansion and creation capabilities to transform short captions into more elaborate descriptions. 
% The quantifiable results can be found in Section~\ref{sec:res_refiner}.

% We develop a prompt refiner that enables the model to reasonably expand based on the input prompt while following it. This alleviates the issue of inconsistencies in sentence length and descriptive granularity during training and generation, significantly enhancing the stability of generated video motion and enriching details. We fine-tune a large language model with our custom-built dataset of 11k text instructions. We outline the model details for the prompt refiner in Section~\ref{sec:refiner}.
% 视频生成模型的训练集由VLM标注, 他们对画面环境和主题的描写非常细致, 且大多数都为长文本, 对比用户输入有很大的差距. 用户输入通常描述不够细致精确、且文字量较少 (如在 VBench 中, 多数测试文本只有30个以内的单词, 甚至只有5个单词), 与模型训练时的文本 condition 有较大的gap, 会导致生成视频画面质量降低、语义遵循度变差, 运动幅度降低. 为了降低模型对短文本的理解难度, 我们通过引入LLM来bridge the gap.旨在让 LLM 通过他的文本续写和创作能力先将短的caption补充为长的caption.我们使用 gpt4o 来生成训练文本对, 通过特定的prompt, 我们让LLM补充更加详细的动作和场景描述, 丰富的镜头语言、光影细节、环境氛围. 用原始和gpt润色后的文本对训练refiner模型
%在 human evaluation 和 VBench score 中都取得了很好的效果. 由于 gpt4 成本太高, 且无法合并为 end to end 的生成pipeline, 因此我们选择用gpt4生成refine 后的caption, 再用这些 caption来finetune语言模型.

%  因此我们使用coco人工标注的12k短cap, 6k条diffisondb 的特色tag, 3k条视觉语言模型标注的jounerydb的短caption 和 400+个dense caption作为原始caption, 让 gpt4 生成对应的训练集finetune llama3.1 8B.

%  在实践过程中, 发现所有的Vbench 结果都有较为明显的提升, 尤其在action 和 物品描述上成绩斐然. 通过观察prompt 也发现gpt没有修改prompt中的物体位置关系和物体数目. 同时可以将多语言的支持实现在 prompt refiner 中, 从而diffusion model 可以使用纯英文的tokenizer
\noindent\textbf{Data preparation.}
We use GPT-4o to generate paired training texts, using specific prompts to instruct the LLM to supplement detailed actions, scene descriptions, cinematic language, lighting nuances, and environmental atmosphere. These original and LLM-augmented text pairs are then used to train the refiner model. Concretely, the instruct prompt is:
\ul{\textit{rewrite the prompt:``{prompt}'' to contain subject description action, scene description. (Optional: camera language, light and shadow, atmosphere) and conceive some additional actions to make the prompt more dynamic, making sure it’s a fluent sentence.}} Our data composition for fine-tuning LLM is shown in Tab.~\ref{table:annotated_data}. Specifically, COCO \cite{lin2014microsoft} consists of manually annotated data, while JourneyDB \cite{sun2024journeydb} contains labels generated by a visual language model (VLM).
% 在推理的时候prompt refiner也不需要放置在显存中, 因此以最小的成本对模型能力进行了优化
\begin{table*}[t]
\setlength\tabcolsep{5.0mm}
\centering
\caption{Overview of utilized datasets for fine-tuning prompt refiner.}
\begin{tabular}{l|cccc}
\toprule
\textbf{Source}   & Year & \textbf{Length} & \textbf{Manual} & \textbf{\# Num}\\ 
\midrule
COCO \cite{lin2014microsoft}     &   2014  &  Short & Yes & 12k \\ 
DiffusionDB \cite{wang2022diffusiondb}  &   2022  &  Tags & Yes & 6k\\ 
JourneyDB \cite{sun2024journeydb}   &   2023   &  Medium  & No & 3k\\ 
Dense Captions (From Internet)  &   2024 &  Dense & Yes & 0.5k \\ 
\bottomrule
\end{tabular}
\label{table:annotated_data}
\end{table*}
% 数据，方法，训练

\noindent\textbf{Training Details.}
We perform LoRA fine-tuning using LLaMA 3.1 8B\footnote{\url{https://huggingface.co/meta-llama/Llama-3.1-8B}}, completing within 1 hour on a single NPU/GPU. Fine-tuning is conducted for just 1 epoch with a batch size of 32 and a LoRA rank of 64. The AdamW optimizer is used with $\beta_1 = 0.9$, $\beta_2 = 0.999$, and a learning rate of 1.5e-4.

%% file: sec/4_data.tex
\section{Data Curation Pipeline}
\label{sec:data}

Dataset quality is closely linked to model performance. However, some current open-source datasets, such as WebVid~\citep{bain2021frozen}, Panda70M~\citep{chen2024panda}, VIDAL~\citep{zhu2023languagebind} and HD-VILA~\citep{xue2022hdvila}, fall short in data quality. Excessive low-quality data in training disrupts the gradient direction of model learning. In this section, we propose an efficient, structured data-processing pipeline to filter high-quality video clips from raw data. We also present dataset statistics to provide reliable direction for further data enhancement.

\subsection{Training Data}
\label{sec:train_data}
% \begin{table}
%     \setlength\tabcolsep{2.0mm}
%     \caption{\textbf{Data card of Open-Sora Plan v1.3.}}
%     \label{tab:lvlm}
%     \centering
%     \begin{tabular}{l|clcc}
%     \toprule
%         \textbf{Dataset}  & \textbf{Method} & \textbf{Task}  & \textbf{Top-1} \\
%         \midrule
%         \multirow{4}{*}{MSR}  & \multirow{2}{*}{ImageBind} & V$\rightarrow$T  & 36.1$^*$ \\
%           &  & A+V$\rightarrow$T  & 36.8 \tiny{(+0.7)} \\
%           \cmidrule(r){3-4}
%           & \multirow{2}{*}{Ours} & V$\rightarrow$T  & 41.4 \\
%           &  & A+V$\rightarrow$T  & 42.0 \tiny{(+0.6)} \\
%         \midrule
%         \multirow{4}{*}{NYU}  & ImageBind & D$\rightarrow$T  & 54.0 \\
%           \cmidrule(r){3-4}
%           & \multirow{3}{*}{Ours} & D$\rightarrow$T & 65.1 \\
%           &  & RGB$\rightarrow$T  & 76.0 \\
%           &  & D+RGB$\rightarrow$T & 77.4 \tiny{(+1.4)} \\
%         \midrule
%         \multirow{2}{*}{LLVIP}  & \multirow{2}{*}{Ours} & RGB$^\dag$$\rightarrow$T & 62.4 \\
%           &  & I+RGB$^\dag$$\rightarrow$T & 79.3 \tiny{(+16.9)} \\
%     \bottomrule
%     \end{tabular}
% \end{table}

\begin{table*}
    \setlength\tabcolsep{1.6mm}
    \caption{\textbf{Data card of Open-Sora Plan v1.3.} ``*'' denotes that the original team employs multiple models, including OFA~\citep{wang2022ofa}, mPLUG-Owl~\citep{ye2023mplug}, and ChatGPT~\citep{openai2023gpt4} to refine captions. ``$\dag$'' indicates that while we do not release captions generated with QWen2-VL and ShareGPT4Video, the original team has made their generated captions publicly available.}
    \label{tab:data_card}
    \centering
    \begin{tabular}{c|ccc|ccc}
    \toprule
          \multirow{2}{*}{\textbf{Domain}} & \multirow{2}{*}{\textbf{Dataset}} & \multirow{2}{*}{\textbf{Source}} &  \multirow{2}{*}{\textbf{Captioner}} & \textbf{Data} &  \textbf{Caption}  & \multirow{2}{*}{\textbf{\# Num}} \\
          & & & & \textbf{Available} & \textbf{Available} & \\
        \midrule
        \multirow{4}{*}{Image}  & SAM & SAM  & LLaVA & Yes& Yes& 11.1M \\
          & Anytext & Anytext & InternVL2 & Yes & Yes & 1.8M \\
          & Human & LAION & InternVL2 & Yes& Yes& 0.1M \\
          & Internal & - & QWen2-VL & No& No& 5.0M \\
        \midrule
         \multirow{6}{*}{Video} & VIDAL & YouTube Shorts & Multi-model$^*$ & Yes& Yes & 2.8M \\
          \cmidrule(r){2-7}
          & \multirow{2}{*}{Panda70M} & \multirow{2}{*}{YouTube}  & QWen2-VL & \multirow{2}{*}{Yes}& \multirow{2}{*}{Yes$^\dag$} & \multirow{2}{*}{21.2M} \\
          &  &   & ShareGPT4Video & & & \\
          \cmidrule(r){2-7}
          & \multirow{3}{*}{StockVideo} & Mixkit$^\ddag$ & \multirow{2}{*}{QWen2-VL} & \multirow{3}{*}{Yes}& \multirow{3}{*}{Yes} & \\
          &  &  Pexels$^\curlywedge$ & \multirow{2}{*}{ShareGPT4Video} & & & 0.8M \\
          &  & Pixabay$^\curlyvee$ &  & & & \\
    \bottomrule
    \end{tabular}
    \begin{tablenotes}
     \small \item[1] $^\ddag$ \url{https://mixkit.co}, $^\curlywedge$ \url{www.pexels.com}, $^\curlyvee$ \url{https://pixabay.com}
    \end{tablenotes}
\vspace{-10pt}
\end{table*}

As shown in Tab.~\ref{tab:data_card}, we obtain 11 million image-text pairs from Pixart-Alpha~\citep{chen2023pixartalpha}, with captions generated by LLaVA~\citep{liu2024visual}. Additionally, we use the OCR dataset Anytext-3M~\citep{tuo2023anytext}, which pairs each image with corresponding OCR characters. We filter Anytext-3M for English data, constituting about half of the entire dataset. Since SAM~\citep{kirillov2023segment} data (as used in Pixart-Alpha) includes blurred faces, we selected 160k high-quality images from Laion-5B~\citep{schuhmann2022laion} to enhance the quality of person-related content in generation. The selection criteria include high resolution, high aesthetic scores, the absence of watermarks, and the presence of people in the images.

For videos, we download approximately 21M horizontal videos from Panda70M~\citep{chen2024panda} using our filtering pipeline. For vertical data, we obtain around 3M vertical videos from VIDAL~\citep{zhu2023languagebind}, sourced from YouTube Shorts. Additionally, we scrape high-quality videos from CC0-licensed websites, such as Mixkit, Pexels, and Pixabay. These open-source video sites contain no content-related watermarks.

\subsection{Data Filtering Strategy}

\begin{table*}
    \small
  \setlength\tabcolsep{1.9mm}
  \caption{Implementation details and discarded data number of different filtering steps.}
  \label{tab:lvlm}
  \centering
  \begin{tabular}{cccc}
    \toprule
    \textbf{Curation Step} & \textbf{Tools} & \textbf{Thresholds} & \textbf{Remaining} \\
    \midrule
    Video Slicing & - & Each video is clipped to 16s & 100\% \\
    Jump Cut & LPIPS~\citep{Zhang_Isola_Efros_Shechtman_Wang_2018} & 32 $\leq$ frames number $\leq$ 512 & 97\% \\
    Motion Calculation & LPIPS~\citep{Zhang_Isola_Efros_Shechtman_Wang_2018} & 0.001 $\leq$ motion score $\leq$ 0.3 & 89\% \\
    OCR Cropping & EasyOCR$^*$ & 0.20 $\leq$ edge  & 89\% \\
    Aesthetic Filtration& Laion Aesthetic Predictor v2$^\dag$ & 4.75 $\leq$ aesthetic score & 49\% \\
    Low-level Quality Filtration& DOVER~\citep{wu2023exploring} & 0 $\leq$ technical score & 44\% \\
    Motion Double-Checking & LPIPS~\citep{Zhang_Isola_Efros_Shechtman_Wang_2018} & 0.001 $\leq$ motion score $\leq$ 0.3 & 42\% \\
    
    \bottomrule
  \end{tabular}
  \begin{tablenotes}
     \small \item[1] $^*$ \url{https://github.com/JaidedAI/EasyOCR}
     \small \item[2] $^\dag$ \url{https://github.com/christophschuhmann/improved-aesthetic-predictor}
   \end{tablenotes}
\vspace{-10pt}
\end{table*}

\begin{enumerate}

\item \textbf{Video Slicing.}
Excessively long videos are not conducive to input processing, so we utilize copy stream method in ffmpeg\footnote{\url{https://ffmpeg.org/}} to split videos into 16-second clips.

\item \textbf{Jump Cut and Motion Calculation.}
% 我们通过跳一帧取视频，计算帧间的Learned Perceptual Image Patch Similarity (LPIPS)，然后将其中的异常点作为跳切点，将均值作为motion值。具体而言，我们利用`decord`库可以高效地跳帧读视频。读入视频后，计算lpips值得到帧间语义相似度集合$\mathcal{L}$，计算其均值$u$和方差$\sigma$。然后计算$\mathcal{L}$的Z score: $\mathcal{Z} = {l \in \mathcal{L}| \frac{l - \mu}{\sigma} }$，得到潜在异常点索引集合$\mathcal{P} = { i | z_i > z_{threshold},  i \in \text{index of} \mathcal{Z}}$。并按照策略进一步筛选异常点，得到异常点索引集合$\mathcal{P}_{final} = \{ i \in \mathcal{P} | \mathcal{L}[i] > l_\text{threshold} or ( z_i > z_\text{threshold2} and \mathcal{L}[i] > l_\text{threshold2} ) }\$。经过实验，我们的参数设定如下：$z_{threshold}=2.0, l_{threshold} = 0.35, z_\text{threshold2}=3.2, l_\text{threshold2}=0.2$.
We calculate the Learned Perceptual Image Patch Similarity (LPIPS) \cite{Zhang_Isola_Efros_Shechtman_Wang_2018} between consecutive frames. Outliers are identified as cut points, while the mean value represents motion. Specifically, we utilize the decord\footnote{\url{https://github.com/dmlc/decord}} library to efficiently read video frames with skipping. After reading the video, we calculate the LPIPS values to obtain a set of semantic similarities between frames, denoted as $l \in \mathcal{L}$, and compute its mean $\mu$ and variance $\sigma$. Then, we calculate the zero score of $\mathcal{L}$: $\mathcal{Z} = \{ z = \frac{l - \mu}{\sigma} | l \in \mathcal{L} \}$, to obtain the set of potential anomaly indices $\mathcal{P} = \{ i | z_i > z_{threshold},  z_i \in \mathcal{Z}\}$. We further filter the anomalies by $\mathcal{P}_{final} = \{ i | \mathcal{L}[i] > l_{threshold}\ or\ ( z_i > z_{threshold2}\ and\ \mathcal{L}[i] > l_{threshold2} ), i \in \mathcal{P} \} $ to obtain the final set of anomaly indices. Based on our experiments, we set the parameters as $z_{threshold}=2.0, l_{threshold} = 0.35, z_{threshold2}=3.2, l_{threshold2}=0.2$. To validate the efficacy of our method, we conduct a manual assessment of 2,000 videos. The result demonstrates that the accuracy meets our predetermined criteria.

\item \textbf{OCR Cropping.} We employ EasyOCR to detect subtitles in videos by sampling one frame per second. 
Based on our estimates for common video platforms, subtitles typically appear in the edge regions, with manual verification showing an average occurrence in 18\% of these areas. 
Therefore, we set the maximum cropping range to 20\% of both sides of video spatial size $(H,W)$, \textit{i.e.}, cropped video has $(0.6H, 0.6W)$ size and 36\% area compared to the original video in extreme cases.
We then crop subtitles appearing in the setting range, leaving any text in the central area unprocessed. 
We consider that text appearing in certain contexts, such as advertisements, speeches, or library settings is reasonable. 
In summary, we do not assume that all text in a video should be filtered out since certain words contribute significance in specific contexts, and we leave further judgments to aesthetic considerations. 
We notice that the OCR step only crops text areas without discarding videos.

\item \textbf{Aesthetic Filtration.} We use the Laion aesthetic predictor to assess the aesthetic score of a video. The aesthetic predictor effectively filters out videos that are blurry, low-resolution, overly exposed, excessively dark, or contain prominent watermarks or logos. We set a threshold of 4.75 to filter videos, as this value effectively removes extensive text and retains high aesthetic quality. We uniformly sample five frames from each video and average their scores to obtain the final aesthetic score. This filtering process eliminates approximately 40\% of videos that do not meet human aesthetic standards.

\item \textbf{Low-level Quality Filtration.} However, even when some data have high resolutions, their visual effects can still appear very blurry or exhibit a mosaic-like quality, which is attributed to two factors: \textbf{(i)} Low bitrate or DPI of the video. \textbf{(ii)} Usage of motion blur techniques in 24 FPS videos, which simulate dynamic effects by blurring the image between frames, resulting in smoother visual motion. For these videos with absolutely low quality, aesthetic filtering struggles to eliminate them since frames are resized to a resolution of 224. We aim to utilize a metric independent of the visual content that evaluates absolute video quality, focusing on issues including compression artifacts, low bitrate, and temporal jitter.
Finally, we find the technical prediction score from DOVER~\citep{wu2023exploring}, selecting videos with a technical score $\textgreater$ 0, which filters out 5\% of the videos.

\hspace*{\fill}

\item \textbf{Motion Double-Checking.} In our post-check, we find that the changes in subtitles may lead to inaccuracies in motion values because the OCR cropping step occurs after detecting motion values. Therefore, we recheck the motion values and filter out videos according to average frame similarities with $\bar{\mathcal{L}} < 0.001$ or $\bar{\mathcal{L}} > 0.3$, which account for 2\%.

\end{enumerate}

\subsection{Data Annotation}
Dense captioning provides additional semantic information for each sample, enabling the model to learn specific correspondences between text and visual features. 
Supervised by dense caption during diffusion training, the model gradually builds a conceptual understanding of various objects and scenes. 
However, the cost of manual annotation for dense captions is prohibitive, so large image-language models~\citep{wang2023cogvlm,yao2024minicpm,chen2024far,chen2023sharegpt4v,lin2024moe,liu2024improved,wang2024qwen2} and large video-language models~\citep{lin2023video,chen2024sharegpt4video,wang2024qwen2,xu2024pllava,liu2024ppllava,wang2024tarsier,jin2024chat} are typically used for annotation. 
This capability allows the model to express complex concepts in dense captions more accurately during image and video generations.

For images, the SAM dataset has available captions generated by LLaVA. Although Anytext contains some OCR-recognized characters, these are insufficient to describe the entire image. Therefore, we use InternVL2~\citep{chen2024far} and QWen2-VL-7B~\citep{wang2024qwen2} to generate captions for the images. The descriptions are as detailed and diverse as possible. The annotation prompt is: \ul{\textit{Combine this rough caption: ``\{\}'', analyze the image in a comprehensive and detailed manner. ``\{\}'' can be recognized in the image.}}

For videos, in early versions such as Open-Sora Plan v1.1, we use ShareGPT4Video-7B~\citep{chen2024sharegpt4video} to annotate a portion of the videos. Another portion is annotated with QWen2-VL-7B~\citep{wang2024qwen2}, with the input prompt: \ul{\textit{Please describe the content of this video in as much detail as possible, including the objects, scenery, animals, characters, and camera movements within the video. Please start the description with the video content directly. Please describe the content of the video and the changes that occur, in chronological order.}}

However, 7B caption models often generate prefixes like ``This image'' or ``The video''. We search all such irrelevant strings and remove them.

\begin{figure}[t]
\begin{minipage}[c]{0.5\linewidth}
    \vspace{15pt}
    \includegraphics[width=\linewidth]{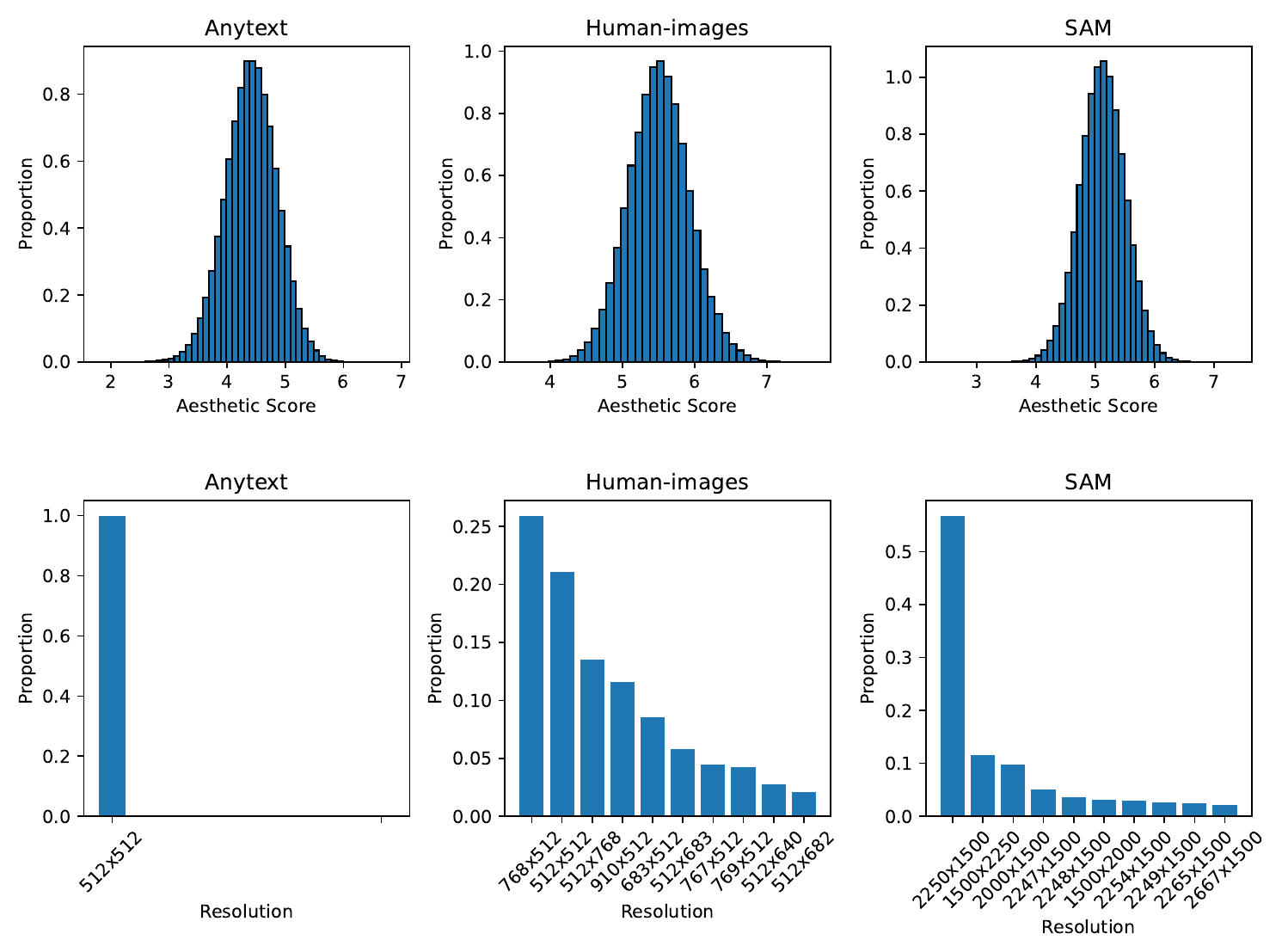}
    \caption*{(a)}
\end{minipage}
\hfill
\begin{minipage}[c]{0.5\linewidth}
  \includegraphics[width=\linewidth]{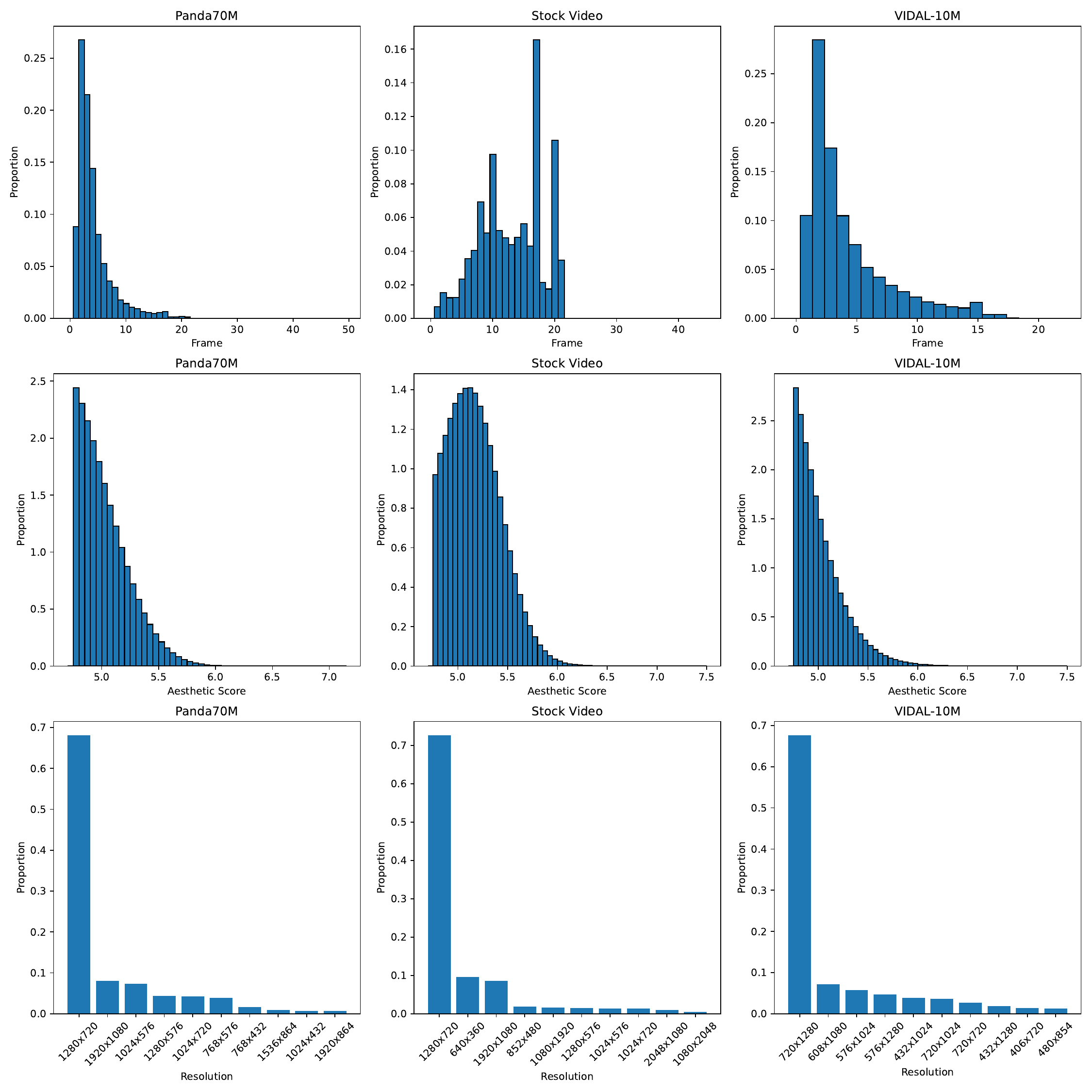}
  \caption*{(b)}
\end{minipage}
\caption{\textbf{(a) Distribution statistics of image datasets.} The first row is the aesthetic scores distribution of the data, and the second row is the resolution distribution of the data. \textbf{(b) Distribution statistics of video datasets.} The first row is the duration distribution of the data, the second row is the aesthetic score distribution of the data, and the third row is the resolution distribution of the data.}
\label{fig:data_analysis}
\vspace{-10pt}
\end{figure}

\subsection{Data Statistics}
% 各种分析数据图
% 图片只分析sam、humanimg、anytext3m，视频只分析 筛过的panda70m、vidal、pixabay、pexel、mixkit
% \item \textbf{Video Data. \textcolor{red}{[Tanghui]}} The filtered video data mainly includes Panda70M, VIDAL-10M, and several stock video websites (Pixabay, Pexels, Mixkit, etc.). We have plotted the top 10 most frequent resolutions, and the histogram of the duration distribution, aesthetic score distribution, and resolution distribution of the three kinds of data is shown in Fig.~\ref{fig:video_data}. As can be seen from the distribution map, the average duration of Panda70M and VIDAL-10M video data is short, and the aesthetic scores are relatively low. The average duration of data from stock video websites is longer, and the aesthetic quality is also improved. The resolution of the three video data is mostly 1280*720, VIDAL-10M is the vertical video dataset (height is greater than width), and the other two datasets is landscape (width is greater than height).

% \begin{wrapfigure}{l}{0.5\linewidth}
% \centering
% \includegraphics[width=\linewidth]{fig/image_data.pdf}
% \caption{\textbf{Distribution statistics of image datasets.} The first row is the aesthetic scores distribution of the data, and the second row is the resolution distribution of the data.}
% \label{fig:image_data}
% \end{wrapfigure}

\textbf{Image Data.} The filtered image data primarily includes Anytext, Human-images, and SAM. We have plotted the top-10 most frequent resolutions, along with histograms depicting the distribution of aesthetic scores, as shown in Fig.~\ref{fig:data_analysis}~(a). 
The plots indicate that the Anytext dataset has a unified resolution $512\times512$. In contrast, Human-images and SAM datasets exhibit more diverse scores and resolutions. Human-images dataset shows a range of scores and multiple resolutions, suggesting varied content, while SAM heavily favors high resolutions $2250\times1500$. Overall, Anytext is consistent, while Human-images and SAM offer greater diversity in both aesthetic scores and image resolutions.

% \begin{wrapfigure}{l}{0.5\linewidth}
% \centering
% \includegraphics[width=\linewidth]{fig/video_data.pdf}
% \caption{\textbf{Distribution statistics of video datasets.} The first row is the duration distribution of the data, the second row is the aesthetic score distribution of the data, and the third row is the resolution distribution of the data.}
% \label{fig:video_data}
% \end{wrapfigure}

\textbf{Video Data.} The filtered video data primarily includes Panda70M, VIDAL-10M, and several stock video websites (\textit{e.g.}, Pixabay, Pexels, Mixkit). We have plotted the top 10 most frequent resolutions, along with histograms depicting the distribution of video duration, aesthetic scores, and resolution across the three datasets, as shown in Fig.~\ref{fig:data_analysis}~(b). From the distribution plots, it is evident that both Panda70M and VIDAL-10M contain shorter average video durations and relatively lower aesthetic scores. In contrast, videos from stock video websites tend to have longer durations and higher aesthetic quality. Regarding resolution, the majority of videos across all three datasets are $1280\times720$, with VIDAL-10M being a vertical video dataset (height $>$ width), while the other two datasets are predominantly landscape (width $>$ height).

%% file: sec/6_result.tex
\section{Results} 

\subsection{Wavelet-Flow VAE}

Tab.~\ref{tb:vae_comparison} and Fig.~\ref{fig:reconstruction} present both quantitative and qualitative comparisons with several open-source VAEs, including Allegro~\citep{zhou2024allegro}, OD-VAE~\citep{chen2024od}, and CogVideoX~\citep{yang2024cogvideox}. The experiments utilize the Panda70M~\citep{Chen_2024_CVPR} and WebVid-10M~\citep{Bain_Nagrani_Varol_Zisserman_2021} datasets. To comprehensively evaluate reconstruction performance, we adopt the Peak Signal-to-Noise Ratio (PSNR)~\citep{Hore_Ziou_2010}, Learned Perceptual Image Patch Similarity (LPIPS)~\citep{Zhang_Isola_Efros_Shechtman_Wang_2018}, and Structural Similarity Index Measure (SSIM)~\citep{wang2004image} as the primary evaluation metrics. Furthermore, the reconstruction Fréchet Video Distance (rFVD)~\citep{Unterthiner_Steenkiste_Kurach_Marinier_Michalski_Gelly_2019} is employed to assess visual quality and temporal coherence.

As shown in Tab.~\ref{tb:vae_comparison}, WF-VAE-S achieves a throughput of 11.11 videos per second when encoding 33-frame videos at 512$\times$512 resolution. This throughput surpasses CV-VAE and OD-VAE by approximately 6$\times$ and 4$\times$, respectively. The memory cost reduces by nearly 5$\times$ and 7$\times$ compared to these baselines while achieving superior reconstruction quality. For the larger WF-VAE-L model, the encoding throughput exceeds Allegro by 7.8$\times$, with approximately 8$\times$ lower memory usage, while maintaining better evaluation metrics. These results demonstrate that the WF-VAE maintains state-of-the-art reconstruction performance while substantially reducing computational costs.

We assess the impact of lossy block-wise inference on reconstruction metrics using contemporary open-source VAE implementations \cite{yang2024cogvideox, chen2024od}, as summarized in Tab.~\ref{tb:tiling_degrade}. Specifically, we measure reconstruction performance in terms of PSNR and LPIPS on the Panda70M dataset under both block-wise and direct inference conditions. the overlap-fusion-based tiling inference of OD-VAE results in substantial performance degradation. In contrast, CogVideoX exhibits only minor degradation due to its temporal block-wise inference with caching. Notably, our proposed Causal Cache mechanism delivers reconstruction results that are numerically identical to those of direct inference, thereby confirming its lossless reconstruction capability.

%Performance metrics include Peak Signal-to-Noise Ratio (PSNR), Learned Perceptual Image Patch Similarity (LPIPS), and Fréchet Video Distance (FVD).

\begin{table*}[t]
\setlength\tabcolsep{3.3mm}
    \caption{\textbf{
    Quantitative comparison with state-of-the-art VAEs on WebVid-10M dataset.} Reconstruction metrics are evaluated on 33-frame videos at a resolution of 256$\times$256.  ``T'' and ``Mem.'' denote encoding throughput and Memory cost (GB), assessed on 33-frame videos at a resolution of 512$\times$512. The highest result is highlighted in \textbf{bold}, and the second highest result is \underline{underlined}.}
    \label{tb:vae_comparison}
  \centering
    \begin{tabular}{c|c|cc|ccc}
    \toprule
    \textbf{Channel} & \textbf{Model} & \textbf{T}$\uparrow$ & \textbf{Mem.} $\downarrow$ & \textbf{PSNR}$\uparrow$ & \textbf{LPIPS}$\downarrow$ & \textbf{rFVD}$\downarrow$ \\
    \midrule
    \multirow{5}{*}{4} & CV-VAE& 1.85 & 25.00  & 30.76  & 0.0803 & 369.23 \\
    &OD-VAE& 2.63 & 31.19  & 30.69  & 0.0553 & 255.92 \\
    &Allegro& 0.71 & 54.35  & \underline{32.18}  & 0.0524 & 209.68 \\
    &WF-VAE-S(Ours)& \textbf{11.11} & \textbf{4.70} & 31.39  & \underline{0.0517} & \underline{188.04}  \\
    &WF-VAE-L(Ours)& 5.55 & 7.00 & \textbf{32.32} & \textbf{0.0513} & \textbf{186.00} \\
    \midrule
    \multirow{2}{*}{16}&CogVideoX& 1.02 & 35.01 & 35.76 & 0.0277 & 59.83 \\
    &WF-VAE-L(Ours) & \textbf{5.55} & \textbf{7.00} & \textbf{35.79} & \textbf{0.0230} & \textbf{54.36} \\
    \bottomrule
    \end{tabular}
\end{table*}

\begin{table*}[t]
\setlength\tabcolsep{5.0mm}
\caption{\textbf{Quantitative analysis of visual quality degradation induced by block-wise inference on Panda70M.} BWI denotes Block-Wise Inference and experiments are conducted on 33 frames with 256$\times$256 resolution. Values highlighted in {\color{my_red}red} signify degradation in comparison to direct inference, whereas values highlighted in {\color{my_green}green} indicate preservation of the quality. }
\centering
        \renewcommand{\arraystretch}{1}
        \begin{tabular}{c|c|c|cc}
            \toprule
            \textbf{Channel} & \textbf{Method} & \textbf{BWI} & \textbf{PSNR}$\uparrow$ & \textbf{LPIPS}$\downarrow$ \\
            \midrule
             \multirow{4}{*}{4} & \multirow{2}{*}{OD-VAE} &  \ding{55} & 30.31 & 0.0439 \\
              & & \ding{51} & 28.51 {\color{my_red}(-1.80)} & 0.0552{\color{my_red}(+0.011)} \\
              \specialrule{0em}{1pt}{1pt}
              \cline{2-5} \specialrule{0em}{1pt}{1pt}
              & \multirow{2}{*}{WF-VAE-L (Ours)} & \ding{55} & 32.10 & 0.0411 \\
              & & \ding{51} & 32.10{\color{my_green} (-0.00)} &  0.0411{\color{my_green} (-0.000)} \\
            \midrule
             \multirow{4}{*}{16} & \multirow{2}{*}{CogVideoX} &  \ding{55} & 35.79 & 0.0198 \\
            &  & \ding{51} & 35.41{\color{my_red}(-0.38)} & 0.0218{\color{my_red}(+0.002)} \\
              \specialrule{0em}{1pt}{1pt}
              \cline{2-5} \specialrule{0em}{1pt}{1pt}
            & \multirow{2}{*}{WF-VAE-L (Ours)} & \ding{55} & 35.87 & 0.0175 \\
            & & \ding{51} & 35.87{\color{my_green} (-0.00)} & 0.0175{\color{my_green} (-0.000)} \\
            \bottomrule
        \end{tabular}
        
        \label{tb:tiling_degrade}
\end{table*}

\begin{table*}[th]
\setlength\tabcolsep{1.8pt}
\renewcommand{\arraystretch}{1.1}
\centering
\caption{\textbf{Quantitative comparison of Open-Sora Plan and other state-of-the-art methods.} ``*'' donates we use our prompt refiner to get results.}
\label{table:vbench_result}
    \begin{tabular}{l|c|cccccc|c|c}
    \toprule
    \multirow{2}{*}\textbf{Model}  & \multirow{2}{*}{\textbf{Size}}  & \textbf{Aesthetic} & \multirow{2}{*}{\textbf{Action}} & \textbf{Object} & \multirow{2}{*}{\textbf{Spatial}} & \multirow{2}{*}{\textbf{Scene}} & \textbf{Multiple} & \textbf{CH} & \textbf{GPT4o} \\
    & & \textbf{Quality} & & \textbf{Class} & & & \textbf{Objects} & \textbf{Score} & \textbf{MTScore} \\
    \midrule
    OpenSora v1.2 & 1.2B  & 56.18 & 85.8 & 83.37  & \underline{67.51}  & \underline{42.47} & \textbf{58.41}  & 51.87 & 2.50  \\
    % OpenSoraPlan v1.2  & 2.7B & 59.36  & 31.0  & 31.72  & 39.84   & 10.47  & 13.34   & 89.76   & 1.08   \\
    CogVideoX-2B  & 1.7B & 58.78   & \underline{89.0}   & 78.00  & 53.91  & 38.59 & 48.48  & 38.60 & 3.09  \\
    CogVideoX-5B & 5.6B   & 56.46  & 77.2  & 76.85  & 45.89  & 41.44  & 46.43  & 48.45   & \underline{3.36}  \\
    % Allegro \cite{zhou2024allegro}  & 2.7B  & 63.74  & 91.4 & 87.52  & 67.15& 46.72  & 59.92 & 121.2  & 3.18 \\
    Mochi-1 & 10.0B & 56.94 & \textbf{94.6} & \textbf{86.51}  & \textbf{69.24} & 36.99  & \underline{50.47} & 28.07 & \textbf{3.76}  \\
    \midrule
    OpenSoraPlan v1.3  & 2.7B & \underline{59.00}  & 81.8  & 70.97  & 44.46 & 28.56 & 35.87 & \textbf{71.00}  & 2.64  \\
    OpenSoraPlan v1.3$^*$ & 2.7B & \textbf{60.70} & 86.4  & \underline{84.72} & 49.63 & \textbf{52.92}  & 44.57 & \underline{68.39}  & 2.95  \\
    \bottomrule
    \end{tabular}
    \vspace{-10pt}
\end{table*}

\subsection{Text-to-Video} 
\label{sec:res_t2v}

We evaluate the quality of our video generation model using VBench \cite{vbench} and ChronoMagic-Bench-150 \cite{chronomagic_bench}. VBench, a commonly used metric in video generation, deconstructs “video generation quality” into several clearly defined dimensions, allowing for a fine-grained, objective assessment. However, many metrics are overly detailed and yield uniformly high scores across models, offering limited reference value. Consequently, we select \textit{Object Class}, \textit{Multiple Object}, and \textit{Human Action} dimensions to evaluate the semantic fidelity of generated objects and human actions. \textit{Aesthetic quality} is used to assess spatial generation effects, while \textit{Spatial relationship} reflected the model’s understanding of spatial relationships. For motion amplitude, we adopted ChronoMagic-Bench since motion evaluation metrics in VBench are considered inadequate.

Tab.~\ref{table:vbench_result} compares the performance of the Open-Sora Plan with other state-of-the-art models. Results indicate that the Open-Sora Plan performs exceptionally well in video generation quality, and it has significant advantages over other models in terms of aesthetic quality, smoothness, and scene restoration fidelity. In addition, our model can automatically optimize the text prompts to further improve the generation quality.

\subsection{Condition Controllers} 
\label{sec:res_i2v}

\noindent\textbf{Image-to-Video.} The video generation capability of image-to-video depends significantly on the performance of the base model and the quality of the initial frame, resulting in challenges in establishing fully objective evaluation metrics. 
To illustrate the generation ability of Open-Sora Plan, we select several showcases, as shown in Fig.~\ref{fig: showcase_i2v}, demonstrating that our model exhibits excellent image-to-video generation capabilities and realistic motion dynamics.
Furthermore, We compare the image-to-video results of several state-of-the-art methods in Fig.~\ref{fig: compre_i2v}, highlighting that Open-Sora Plan strikes an exceptional balance between the control information of the initial frame and the text. Our method maintains semantic consistency while ensuring high visual quality, demonstrating superior expressiveness compared to other models.

\begin{figure*}[t]
\centering
\includegraphics[width=\linewidth]{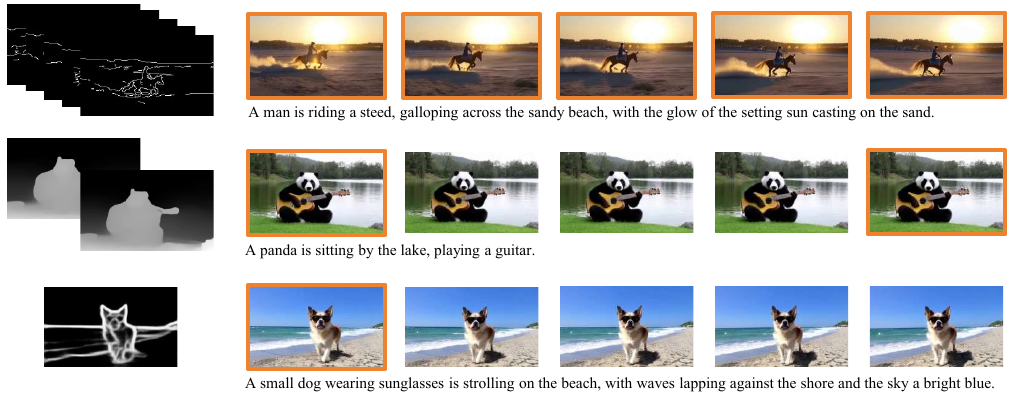}
\caption{Our structure controller can generate high-quality videos conditioned by specified structural signals corresponding to arbitrary frames.
}
\vspace{-10pt}
\label{control_result}
\end{figure*}

\noindent\textbf{Structure-to-Video.}
As shown in Fig.~\ref{control_result}, our structure condition controller enables the Open-Sora Plan text-to-image model to generate high-quality videos whose any frames (first frame, a few frames, all frames, \textit{etc.}) can be accurately controlled by given structural signals (canny, depth, sketch, \textit{etc.}).

\hspace*{\fill}

\subsection{Prompt Refiner} 
\label{sec:res_refiner}

\begin{wrapfigure}{r}{0.5\linewidth}
    \vspace{-5pt}
    \centering 
    \includegraphics[width=1.0\linewidth]{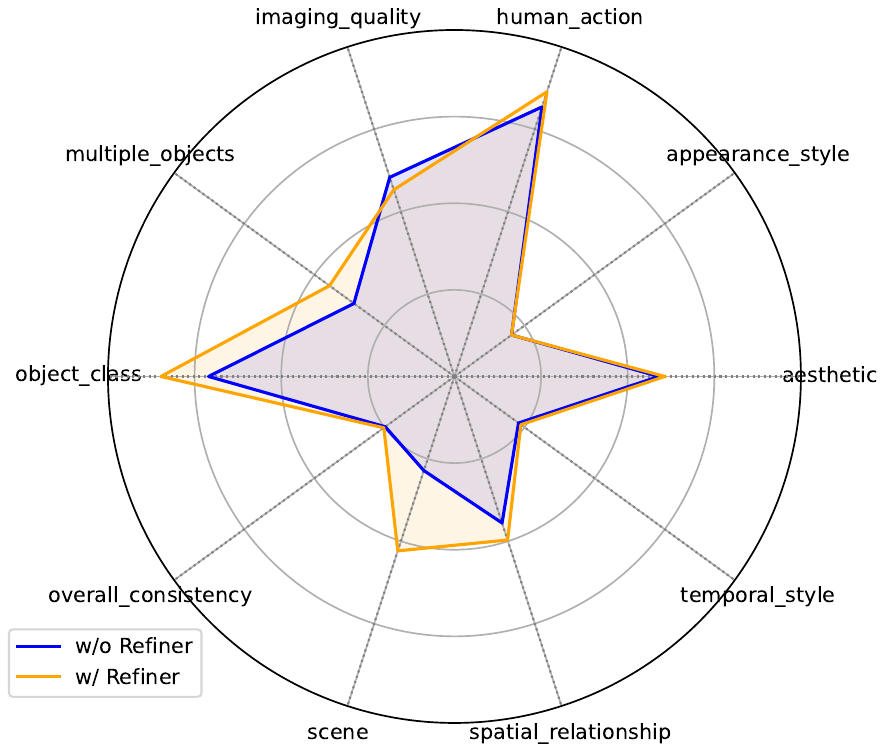}
    \caption{\textbf{Ablations results for leveraging the prompt refiner in VBench.} Evaluated videos are generated in 480p.}
    \vspace{-10pt}
    \label{fig:promptrefiner}
\end{wrapfigure}
The Open-Sora Plan leverages a substantial proportion of synthetic labels during training, resulting in superior performance in dense captioning tasks compared to shorter prompts. 
However, the evaluation prompts or user inputs are often brief, limiting the ability to accurately assess the model’s true performance. 
Following DALL-E 3~\citep{Dalle3}, we report evaluation results where our prompt refiner is employed for rewriting input prompts.

During the evaluation, we observe notable improvements in most VBench \cite{vbench} metrics when using prompt refiner, particularly in action accuracy and object description. Fig.~\ref{fig:promptrefiner} provides a radar chart that visually highlights the effectiveness of the prompt refiner. Specifically, the performance in human action generation and spatial relationship depiction improved by more than 5\%. The semantic adherence for single-object and multi-object generation increased by 15\% and 10\%, respectively. Additionally, the score for scenery generation increased by 25\%. 
Furthermore, our prompt refiner can translate multilingual into English, allowing the diffusion model to leverage training data and text encoders in English while supporting various languages for inference.

%% file: sec/7_limitation.tex
\section{Limitation and Future Work} 

\subsection{Wavelet-Flow VAE} 

Our decoder architecture is modeled after the design proposed by \cite{rombach2022high}, resulting in a greater number of parameters in the decoder compared to the encoder. While the computational cost remains manageable, we consider these additional parameters to be redundant. Consequently, in future work, we plan to streamline the model to fully exploit the advantages of our architecture.

\subsection{Transformer Denoiser} 

The current 2B model in version 1.3.0 shows performance saturation during the later stages of training. However, our model performs poor in understanding physical laws (\textit{e.g.}, a cup overflowing with milk, a car moving forward, or a person walking), thus we have three hypotheses:

\begin{itemize}

\item \textbf{Joint training of images and videos.} Models such as Open-Sora v1.2~\citep{opensora}, EasyAnimate v4~\citep{xu2024easyanimate}, and Vchitect-2.0\footnote{\url{https://github.com/Vchitect/Vchitect-2.0}} can easily generate high-visual-quality videos, possibly due to their direct inheritance of image weights (Pixart-Sigma~\citep{chen2024pixart}, HunyuanDiT~\citep{li2024hunyuan}, SD3~\citep{esser2024scaling}). They train the model with a small amount of video data to learn how to flow along the temporal dimension based on 2D images. However, we train images from scratch with only 10M-level data, which is far from sufficient. In recent work on Allegro~\citep{zhou2024allegro}, they fine-tuned a better text-to-image model based on the T2I weights from Open-Sora Plan v1.2, achieving improved text-to-video results. We have two hypotheses regarding the training strategy: \textbf{(i)} Start joint training from scratch, with images significantly outnumbering videos; \textbf{(ii)} First train a high-quality image model and then use joint training, with a higher proportion of videos at that stage. Considering the learning path and training costs, the second approach may offer more decoupling, while the first aligns better with scaling laws.

\item \textbf{The model still needs to scale.} By observing the differences between CogVideoX-2B~\citep{yang2024cogvideox} and its 5B variant, we can discover that the 5B model understands more physical laws than the 2B model. 
We speculate that instead of spending excessive effort designing for smaller models, it may be more effective to leverage scaling laws to solve these issues. In the next version, we will scale up the model to explore the boundaries of video generation.
We currently have two plans: \textbf{(i)} Continue using the Deepspeed~\citep{rasley2020deepspeed}/FSDP~\citep{zhao2023pytorch} approach, sharding the EMA and text encoder across ranks with Zero3~\citep{rasley2020deepspeed}, which is sufficient for training 10-15B models. \textbf{(ii)} Adopting MindSpeed\footnote{\url{https://gitee.com/ascend/MindSpeed}}/Megatron-LM~\citep{shoeybi2019megatron} for various parallel strategies, enabling us to scale the model up to 30B.

\item \textbf{Supervised loss in training.} Flow Matching~\citep{lipman2022flow} avoids the stability issues in Denoising Diffusion Probabilistic Models~\citep{ho2020denoising} (DDPM) when the timestep approaches 0, addressing the zero-terminal signal-to-noise ratio problem~\citep{lin2024common}. Recent works~\citep{opensora,polyak2024movie,esser2024scaling} also show that the validation loss in Flow Matching indicates whether the model is converging in the right direction, which is crucial for assessing model training progress. Whether flow-based models are more suitable than v-prediction models requires further ablation studies.
\end{itemize}

In addition to expanding the model and data scale, we will also explore other efficient algorithm implementations and improved evaluation metrics:
\begin{itemize}
    \item \textbf{Exploring more efficient architectures.} Although Skiparse Attention significantly reduces FLOPs during computation, these advantages are only noticeable with longer sequence lengths (\textit{e.g.}, resolutions above 480P). Since most pre-training is conducted at a lower resolution (\textit{e.g.}, around 320 pixels), the Skiparse Attention operation has not achieved the desired acceleration ratio in this phase. In the future, we will explore more efficient training strategies to address this issue.
    \item \textbf{Introducing more parallelization strategies.} In Movie Gen~\citep{polyak2024movie}, the role of various parallelization strategies in accelerating training for video generation models is highlighted. However, Open-Sora Plan v1.3.0 currently only employs data parallelism (DP). In the future, we plan to explore additional parallelization strategies to enhance training efficiency. Additionally, in Skiparse Attention, each token only needs to attend to at most the same $\frac{2}{k}-\frac{1}{k^2}$ tokens throughout, without requiring access to other tokens. This operation naturally suits a sequence parallelization strategy. However, the efficient implementation of this sequence parallelization in code remains a topic for further exploration.
    \item \textbf{Establishing reliable evaluation metrics.} Although works like Vbench~\citep{vbench} and Chronomagic Bench~\citep{chronomagic_bench} have proposed metrics to automate the evaluation of video model outputs, these metrics still cannot fully replace human review~\citep{polyak2024movie}. Human evaluation is labor-intensive and incurs significant costs, making it less feasible at scale. Therefore, developing more accurate and reliable automated metrics remains a key area for future research, and we will prioritize this in our work.
\end{itemize}

\subsection{Data} 
Despite ongoing improvements to our training data, the current dataset still faces several significant limitations in terms of data diversity, temporal modeling, video quality, and cross-modal information. We discuss these limitations and outline the corresponding directions for future works:
\begin{itemize}
\item \textbf{Lack of Data Diversity and Complexity.} 
The current dataset predominantly covers specific domains such as simple actions, human faces, and a narrow range of scene types. We randomly sampled 2,000 videos from Panda70M and conducted manual verification, finding that less than 1\% featured cars in motion, and there were even fewer than 10 videos of people walking. Approximately 80\% of the videos consist of half-body conversations with multiple people in front of the camera. Therefore, we speculate that the narrow data domain of Panda70M restricts the model's ability to generate many scenarios. Consequently, it lacks the ability to generate complex, dynamic scenes involving realistic human movement, object deformations, and intricate natural environments. This limitation hinders the model’s capacity to produce diverse and complex video content. Future work will focus on expanding the dataset to encompass a broader spectrum of dynamic and realistic environments, including more complex human interactions and dynamic physical effects. This expansion aims to improve the model’s generalization ability and facilitate the generation of high-quality, varied dynamic videos.

\item \textbf{Lack of Camera Movement, Video Style, and Motion Speed Annotations.}
The current dataset lacks annotations for key dynamic aspects of video content, such as camera movement, video style, and motion speed. These annotations are essential for capturing the varied visual characteristics and movement dynamics within videos. Without them, the dataset may not fully support tasks that require detailed understanding of these elements, limiting the model’s ability to handle diverse video content. In future work, we will include these annotations to enhance the dataset’s versatility and improve the model’s ability to generate more contextually rich video content.

\item \textbf{Limitations in Video Resolution and Quality.}
Although the dataset includes videos at common resolutions (\textit{e.g.}, 720P), these resolutions are insufficient for high-quality video generation tasks, such as generating detailed virtual characters or complex, high-fidelity scenes. The resolution and quality of the current dataset become limiting factors when generating fine-grained details or realistic dynamic environments. To address this limitation, future work should aim to incorporate high-resolution videos (\textit{e.g.}, 1080P, 2K), which will enable the generation of higher-quality videos with enhanced visual detail and realism.

\item \textbf{Lack of Cross-Modal Information.}
The dataset predominantly focuses on video imagery and lacks complementary modalities such as audio or other forms of multi-modal data. This absence of cross-modal information limits the flexibility and applicability of generative models, particularly in tasks that involve speech, emotions, or contextual understanding. Future research should focus on integrating multi-modal data into the dataset. This will enhance the model's ability to generate richer, more contextually nuanced content, thereby improving the overall performance and versatility of the generative system.

\end{itemize}

%% file: sec/8_conclusion.tex
\section{Conclusion}
We present Open-Sora Plan, our open-source high-quality and long-duration video generation project in this work.
In the framework aspect, we decompose the entire video generation model into a Wavelet-Flow Variational Autoencoder, a Joint Image-Video Skiparse Denoiser, and various condition controllers.
In the strategy aspect, we carefully design a min-max token strategy for efficient training, an adaptive gradient clipping strategy for preventing outflow gradients, and a prompt refiner for obtaining more appreciative results.
Furthermore, we propose a multi-dimensional data curation pipeline for automatic high-quality data exploitation.
While our Open-Sora Plan achieving a remarkable milestone, we will make more effort to promote the progress of the high-quality video generation research area and open-source community.
% \noindent\textbf{Future Works.}
% Our future efforts mainly focus on two directions: \textit{i.} Exploring the scaling law in model parameters and training data to generate more impressive results. \textit{ii.} Exploring the effective injection of various signals especially temporal conditions into models for user-controllable video generation and edits.

%% file: sec/9_author.tex
\section*{Contributors and Acknowledgements}

\subsection*{Contributors}

\noindent Bin Lin\footnotemark[1], Yunyang Ge\footnotemark[1], Xinhua Cheng\footnotemark[1], Zongjian Li, Bin Zhu, Shaodong Wang, Xianyi He, Yang Ye, Shenghai Yuan, Liuhan Chen, Tanghui Jia, Junwu Zhang, Zhenyu Tang, Yatian Pang, Bin She, Cen Yan, Zhiheng Hu, Xiaoyi Dong, Lin Chen, Zhang Pan, Xing Zhou, Shaoling Dong, Yonghong Tian, Li Yuan

\subsection*{Project Lead}
\noindent Li Yuan%\footnotemark[2]

\footnotetext[1]{Core contributors with equal contributions}
%\footnotetext[2]{Corresponding author: yuanli-ece@pku.edu.cn}

\subsection*{Acknowledgements}
\noindent We sincerely appreciate Zesen Cheng, Chengshu Zhao, Zongying Lin, Yihang Liu, Ziang Wu, Peng Jin, Hao Li for their valuable supports for our Open-Sora Plan project.

%% file: sec/X_suppl.tex
\clearpage
\setcounter{page}{1}
% \maketitlesupplementary

\section*{Appendix}
\label{sec:rationale}

\begin{figure*}[th]
    \centering    
    \includegraphics[width=\linewidth]{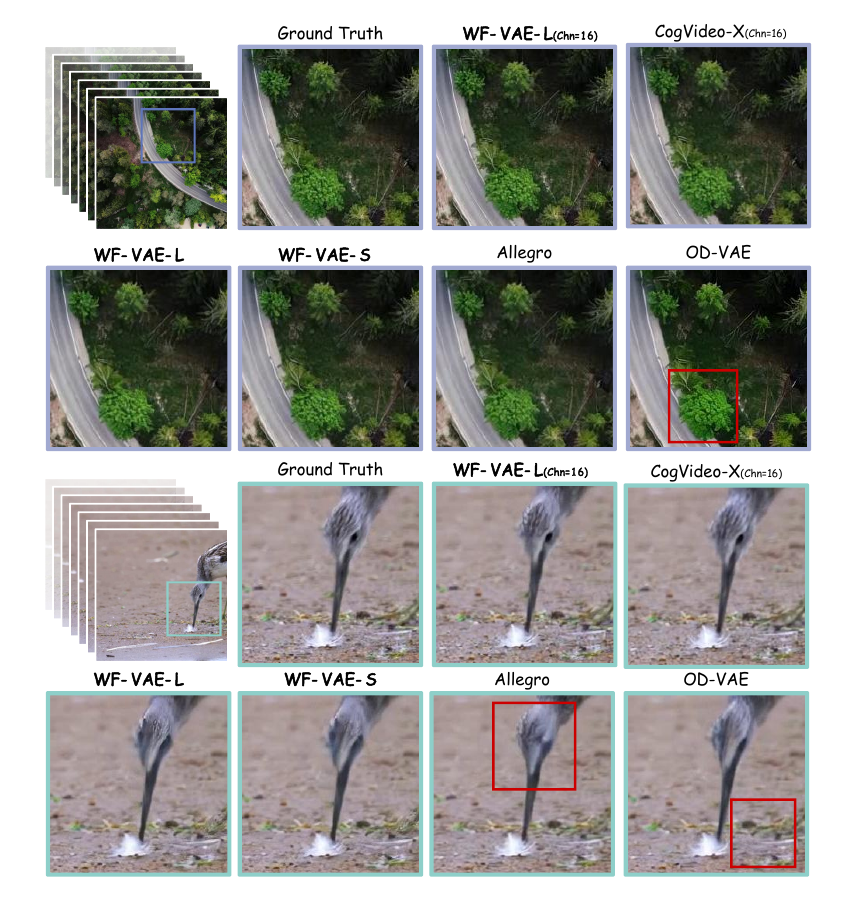}
    \caption{\textbf{Qualitative comparison of state-of-the-art VAEs.} Top: High-detail static scene reconstruction. Bottom: Dynamic scene reconstruction under motion blur.}
    \label{fig:reconstruction}
\end{figure*}

\begin{figure*}[th]
    \centering
    \includegraphics[width=\linewidth]{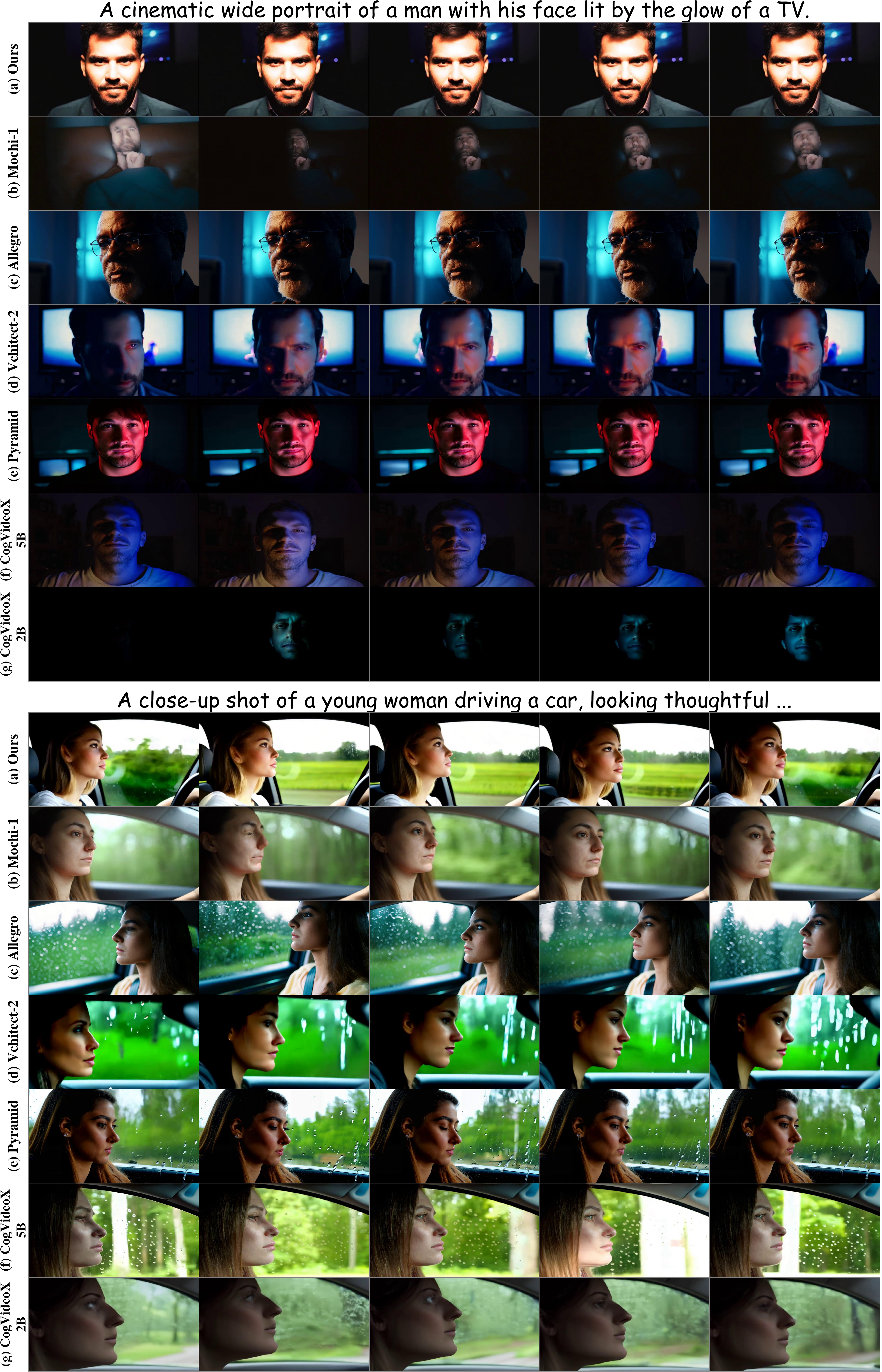}
    \caption{\textbf{Comparison among several state-of-the-art methods in Text-to-Video Task.}}
    \label{fig: compre_t2v}
\end{figure*}

\begin{figure*}[th]
    \centering
    \includegraphics[width=\linewidth]{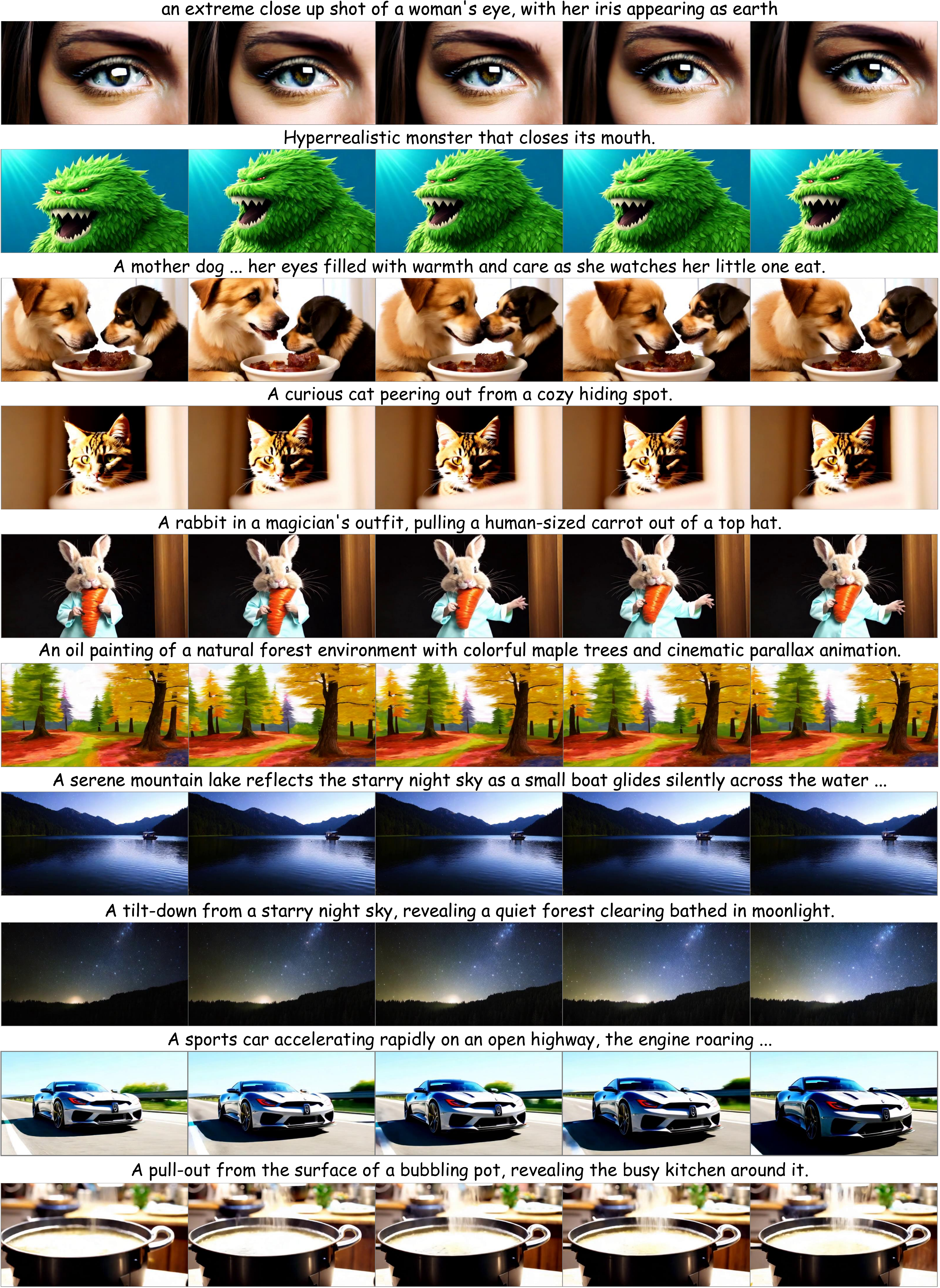}
    \caption{\textbf{Text-to-Video Showcases.}}
    \label{fig: showcase_t2v}
\end{figure*}

\begin{figure*}[th]
    \centering
    \includegraphics[width=\textwidth]{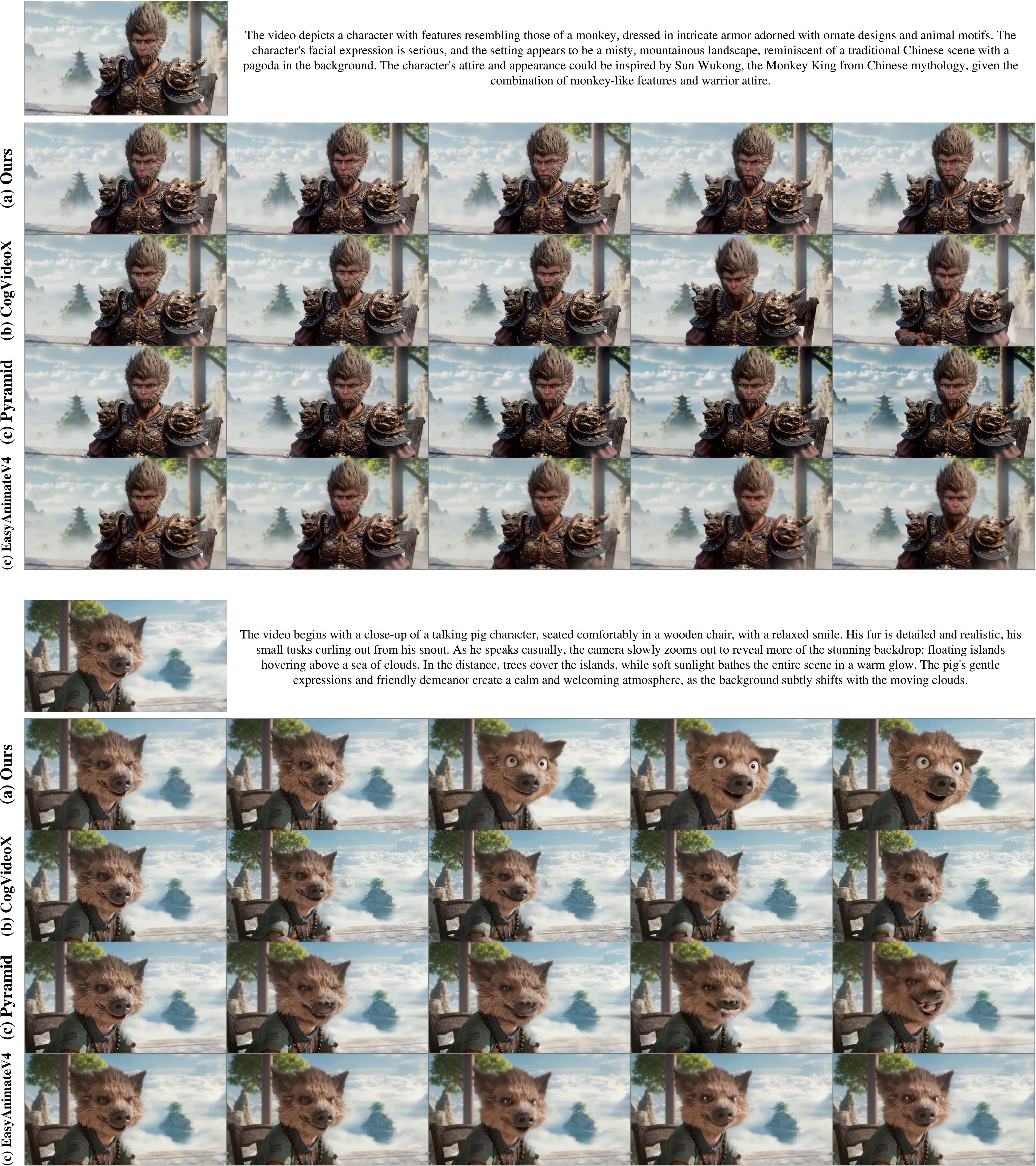}
    \caption{\textbf{Comparison among several state-of-the-art methods in Image-to-Video Task.}}
    \label{fig: compre_i2v}
\end{figure*}

\begin{figure*}[t]
    \centering
    \includegraphics[width=\linewidth]{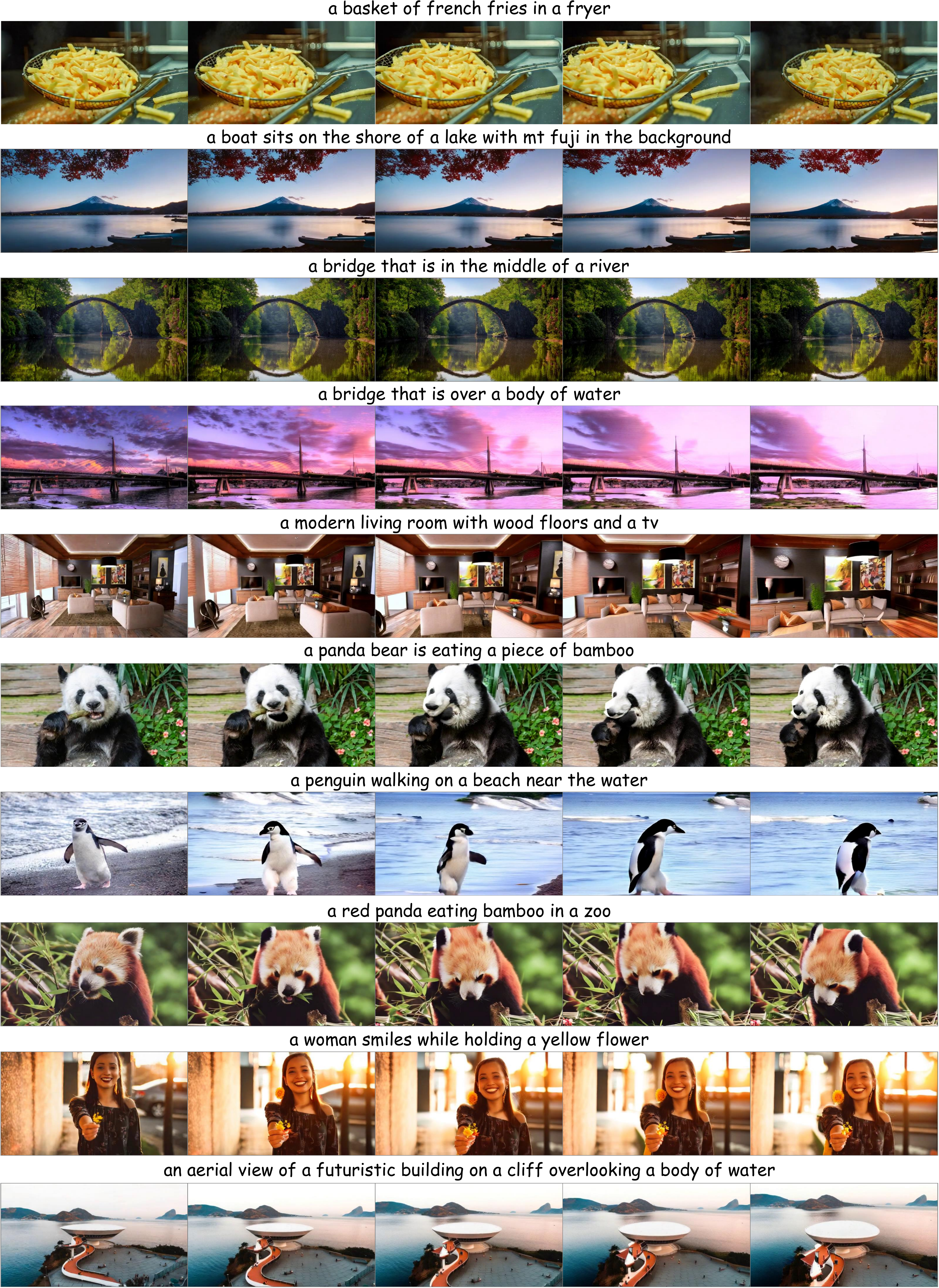}
    \caption{\textbf{Image-to-Video Showcases.}}
    \label{fig: showcase_i2v}
\end{figure*}

%% file: main.bbl
\begin{thebibliography}{89}
\providecommand{\natexlab}[1]{#1}
\providecommand{\url}[1]{\texttt{#1}}
\expandafter\ifx\csname urlstyle\endcsname\relax
  \providecommand{\doi}[1]{doi: #1}\else
  \providecommand{\doi}{doi: \begingroup \urlstyle{rm}\Url}\fi

\bibitem[Bain et~al.(2021{\natexlab{a}})Bain, Nagrani, Varol, and Zisserman]{Bain_Nagrani_Varol_Zisserman_2021}
Max Bain, Arsha Nagrani, Gul Varol, and Andrew Zisserman.
\newblock Frozen in time: A joint video and image encoder for end-to-end retrieval.
\newblock In \emph{2021 IEEE/CVF International Conference on Computer Vision (ICCV)}, 2021{\natexlab{a}}.

\bibitem[Bain et~al.(2021{\natexlab{b}})Bain, Nagrani, Varol, and Zisserman]{bain2021frozen}
Max Bain, Arsha Nagrani, G{\"u}l Varol, and Andrew Zisserman.
\newblock Frozen in time: A joint video and image encoder for end-to-end retrieval.
\newblock In \emph{Proceedings of the IEEE/CVF international conference on computer vision}, pages 1728--1738, 2021{\natexlab{b}}.

\bibitem[Betker et~al.()Betker, Goh, Jing, Brooks, Wang, Li, Ouyang, Zhuang, Lee, Guo, Manassra, Dhariwal, Chu, Jiao, and Ramesh]{Dalle3}
James Betker, Gabriel Goh, Li Jing, Tim Brooks, Jianfeng Wang, Linjie Li, Long Ouyang, Juntang Zhuang, Joyce Lee, Yufei Guo, Wesam Manassra, Prafulla Dhariwal, Casey Chu, Yunxin Jiao, and Aditya Ramesh.
\newblock Improving image generation with better captions.

\bibitem[Birkl et~al.(2023)Birkl, Wofk, and M{\"u}ller]{midas}
Reiner Birkl, Diana Wofk, and Matthias M{\"u}ller.
\newblock Midas v3. 1--a model zoo for robust monocular relative depth estimation.
\newblock \emph{arXiv preprint arXiv:2307.14460}, 2023.

\bibitem[Blattmann et~al.(2023)Blattmann, Dockhorn, Kulal, Mendelevitch, Kilian, Lorenz, Levi, English, Voleti, Letts, et~al.]{blattmann2023stable}
Andreas Blattmann, Tim Dockhorn, Sumith Kulal, Daniel Mendelevitch, Maciej Kilian, Dominik Lorenz, Yam Levi, Zion English, Vikram Voleti, Adam Letts, et~al.
\newblock Stable video diffusion: Scaling latent video diffusion models to large datasets.
\newblock \emph{arXiv preprint arXiv:2311.15127}, 2023.

\bibitem[Brooks et~al.(2024)Brooks, Peebles, Holmes, DePue, Guo, Jing, Schnurr, Taylor, Luhman, Luhman, Ng, Wang, and Ramesh]{videoworldsimulators2024}
Tim Brooks, Bill Peebles, Connor Holmes, Will DePue, Yufei Guo, Li Jing, David Schnurr, Joe Taylor, Troy Luhman, Eric Luhman, Clarence Ng, Ricky Wang, and Aditya Ramesh.
\newblock Video generation models as world simulators.
\newblock 2024.

\bibitem[Canny(1986)]{canny}
John Canny.
\newblock A computational approach to edge detection.
\newblock \emph{IEEE Transactions on pattern analysis and machine intelligence}, \penalty0 (6):\penalty0 679--698, 1986.

\bibitem[Chen et~al.(2023{\natexlab{a}})Chen, Yu, Ge, Yao, Xie, Wu, Wang, Kwok, Luo, Lu, and Li]{chen2023pixartalpha}
Junsong Chen, Jincheng Yu, Chongjian Ge, Lewei Yao, Enze Xie, Yue Wu, Zhongdao Wang, James Kwok, Ping Luo, Huchuan Lu, and Zhenguo Li.
\newblock Pixart-$\alpha$: Fast training of diffusion transformer for photorealistic text-to-image synthesis, 2023{\natexlab{a}}.

\bibitem[Chen et~al.(2024{\natexlab{a}})Chen, Ge, Xie, Wu, Yao, Ren, Wang, Luo, Lu, and Li]{chen2024pixart}
Junsong Chen, Chongjian Ge, Enze Xie, Yue Wu, Lewei Yao, Xiaozhe Ren, Zhongdao Wang, Ping Luo, Huchuan Lu, and Zhenguo Li.
\newblock Pixart-$\backslash$sigma: Weak-to-strong training of diffusion transformer for 4k text-to-image generation.
\newblock \emph{arXiv preprint arXiv:2403.04692}, 2024{\natexlab{a}}.

\bibitem[Chen et~al.(2023{\natexlab{b}})Chen, Li, Dong, Zhang, He, Wang, Zhao, and Lin]{chen2023sharegpt4v}
Lin Chen, Jinsong Li, Xiaoyi Dong, Pan Zhang, Conghui He, Jiaqi Wang, Feng Zhao, and Dahua Lin.
\newblock Sharegpt4v: Improving large multi-modal models with better captions.
\newblock \emph{arXiv preprint arXiv:2311.12793}, 2023{\natexlab{b}}.

\bibitem[Chen et~al.(2024{\natexlab{b}})Chen, Li, Lin, Zhu, Wang, Yuan, Zhou, Cheng, and Yuan]{chen2024od}
Liuhan Chen, Zongjian Li, Bin Lin, Bin Zhu, Qian Wang, Shenghai Yuan, Xing Zhou, Xinghua Cheng, and Li Yuan.
\newblock Od-vae: An omni-dimensional video compressor for improving latent video diffusion model.
\newblock \emph{arXiv preprint arXiv:2409.01199}, 2024{\natexlab{b}}.

\bibitem[Chen et~al.(2024{\natexlab{c}})Chen, Wei, Li, Dong, Zhang, Zang, Chen, Duan, Lin, Tang, et~al.]{chen2024sharegpt4video}
Lin Chen, Xilin Wei, Jinsong Li, Xiaoyi Dong, Pan Zhang, Yuhang Zang, Zehui Chen, Haodong Duan, Bin Lin, Zhenyu Tang, et~al.
\newblock Sharegpt4video: Improving video understanding and generation with better captions.
\newblock \emph{arXiv preprint arXiv:2406.04325}, 2024{\natexlab{c}}.

\bibitem[Chen et~al.(2024{\natexlab{d}})Chen, Siarohin, Menapace, Deyneka, Chao, Jeon, Fang, Lee, Ren, Yang, and Tulyakov]{Chen_2024_CVPR}
Tsai-Shien Chen, Aliaksandr Siarohin, Willi Menapace, Ekaterina Deyneka, Hsiang-wei Chao, Byung~Eun Jeon, Yuwei Fang, Hsin-Ying Lee, Jian Ren, Ming-Hsuan Yang, and Sergey Tulyakov.
\newblock Panda-70m: Captioning 70m videos with multiple cross-modality teachers.
\newblock In \emph{Proceedings of the IEEE/CVF Conference on Computer Vision and Pattern Recognition (CVPR)}, pages 13320--13331, 2024{\natexlab{d}}.

\bibitem[Chen et~al.(2024{\natexlab{e}})Chen, Siarohin, Menapace, Deyneka, Chao, Jeon, Fang, Lee, Ren, Yang, et~al.]{chen2024panda}
Tsai-Shien Chen, Aliaksandr Siarohin, Willi Menapace, Ekaterina Deyneka, Hsiang-wei Chao, Byung~Eun Jeon, Yuwei Fang, Hsin-Ying Lee, Jian Ren, Ming-Hsuan Yang, et~al.
\newblock Panda-70m: Captioning 70m videos with multiple cross-modality teachers.
\newblock In \emph{Proceedings of the IEEE/CVF Conference on Computer Vision and Pattern Recognition}, pages 13320--13331, 2024{\natexlab{e}}.

\bibitem[Chen et~al.(2024{\natexlab{f}})Chen, Wang, Tian, Ye, Gao, Cui, Tong, Hu, Luo, Ma, et~al.]{chen2024far}
Zhe Chen, Weiyun Wang, Hao Tian, Shenglong Ye, Zhangwei Gao, Erfei Cui, Wenwen Tong, Kongzhi Hu, Jiapeng Luo, Zheng Ma, et~al.
\newblock How far are we to gpt-4v? closing the gap to commercial multimodal models with open-source suites.
\newblock \emph{arXiv preprint arXiv:2404.16821}, 2024{\natexlab{f}}.

\bibitem[Dehghani et~al.(2024)Dehghani, Mustafa, Djolonga, Heek, Minderer, Caron, Steiner, Puigcerver, Geirhos, Alabdulmohsin, et~al.]{dehghani2024patch}
Mostafa Dehghani, Basil Mustafa, Josip Djolonga, Jonathan Heek, Matthias Minderer, Mathilde Caron, Andreas Steiner, Joan Puigcerver, Robert Geirhos, Ibrahim~M Alabdulmohsin, et~al.
\newblock Patch n’pack: Navit, a vision transformer for any aspect ratio and resolution.
\newblock \emph{Advances in Neural Information Processing Systems}, 36, 2024.

\bibitem[Dubey et~al.(2024)Dubey, Jauhri, Pandey, Kadian, Al-Dahle, Letman, Mathur, Schelten, Yang, Fan, et~al.]{dubey2024llama}
Abhimanyu Dubey, Abhinav Jauhri, Abhinav Pandey, Abhishek Kadian, Ahmad Al-Dahle, Aiesha Letman, Akhil Mathur, Alan Schelten, Amy Yang, Angela Fan, et~al.
\newblock The llama 3 herd of models.
\newblock \emph{arXiv preprint arXiv:2407.21783}, 2024.

\bibitem[Esser et~al.(2021)Esser, Rombach, and Ommer]{Esser_2021_CVPR}
Patrick Esser, Robin Rombach, and Bjorn Ommer.
\newblock Taming transformers for high-resolution image synthesis.
\newblock In \emph{Proceedings of the IEEE/CVF Conference on Computer Vision and Pattern Recognition (CVPR)}, pages 12873--12883, 2021.

\bibitem[Esser et~al.(2024)Esser, Kulal, Blattmann, Entezari, M{\"u}ller, Saini, Levi, Lorenz, Sauer, Boesel, et~al.]{esser2024scaling}
Patrick Esser, Sumith Kulal, Andreas Blattmann, Rahim Entezari, Jonas M{\"u}ller, Harry Saini, Yam Levi, Dominik Lorenz, Axel Sauer, Frederic Boesel, et~al.
\newblock Scaling rectified flow transformers for high-resolution image synthesis.
\newblock In \emph{Forty-first International Conference on Machine Learning}, 2024.

\bibitem[Guo et~al.(2023)Guo, Yang, Rao, Liang, Wang, Qiao, Agrawala, Lin, and Dai]{guo2023animatediff}
Yuwei Guo, Ceyuan Yang, Anyi Rao, Zhengyang Liang, Yaohui Wang, Yu Qiao, Maneesh Agrawala, Dahua Lin, and Bo Dai.
\newblock Animatediff: Animate your personalized text-to-image diffusion models without specific tuning.
\newblock \emph{arXiv preprint arXiv:2307.04725}, 2023.

\bibitem[Guo et~al.(2025)Guo, Yang, Rao, Agrawala, Lin, and Dai]{sparsectrl}
Yuwei Guo, Ceyuan Yang, Anyi Rao, Maneesh Agrawala, Dahua Lin, and Bo Dai.
\newblock Sparsectrl: Adding sparse controls to text-to-video diffusion models.
\newblock In \emph{European Conference on Computer Vision}, pages 330--348. Springer, 2025.

\bibitem[Hang et~al.(2023)Hang, Gu, Li, Bao, Chen, Hu, Geng, and Guo]{hang2023efficient}
Tiankai Hang, Shuyang Gu, Chen Li, Jianmin Bao, Dong Chen, Han Hu, Xin Geng, and Baining Guo.
\newblock Efficient diffusion training via min-snr weighting strategy.
\newblock In \emph{Proceedings of the IEEE/CVF International Conference on Computer Vision}, pages 7441--7451, 2023.

\bibitem[Ho et~al.(2020)Ho, Jain, and Abbeel]{ho2020denoising}
Jonathan Ho, Ajay Jain, and Pieter Abbeel.
\newblock Denoising diffusion probabilistic models.
\newblock \emph{Advances in neural information processing systems}, 33:\penalty0 6840--6851, 2020.

\bibitem[Hore and Ziou(2010)]{Hore_Ziou_2010}
Alain Hore and Djemel Ziou.
\newblock Image quality metrics: Psnr vs. ssim.
\newblock In \emph{2010 20th International Conference on Pattern Recognition}, 2010.

\bibitem[Huang et~al.(2024)Huang, He, Yu, Zhang, Si, Jiang, Zhang, Wu, Jin, Chanpaisit, et~al.]{vbench}
Ziqi Huang, Yinan He, Jiashuo Yu, Fan Zhang, Chenyang Si, Yuming Jiang, Yuanhan Zhang, Tianxing Wu, Qingyang Jin, Nattapol Chanpaisit, et~al.
\newblock Vbench: Comprehensive benchmark suite for video generative models.
\newblock In \emph{Proceedings of the IEEE/CVF Conference on Computer Vision and Pattern Recognition}, pages 21807--21818, 2024.

\bibitem[Jiang et~al.(2023)Jiang, Sablayrolles, Mensch, Bamford, Chaplot, Casas, Bressand, Lengyel, Lample, Saulnier, et~al.]{jiang2023mistral}
Albert~Q Jiang, Alexandre Sablayrolles, Arthur Mensch, Chris Bamford, Devendra~Singh Chaplot, Diego de~las Casas, Florian Bressand, Gianna Lengyel, Guillaume Lample, Lucile Saulnier, et~al.
\newblock Mistral 7b.
\newblock \emph{arXiv preprint arXiv:2310.06825}, 2023.

\bibitem[Jin et~al.(2024)Jin, Takanobu, Zhang, Cao, and Yuan]{jin2024chat}
Peng Jin, Ryuichi Takanobu, Wancai Zhang, Xiaochun Cao, and Li Yuan.
\newblock Chat-univi: Unified visual representation empowers large language models with image and video understanding.
\newblock In \emph{Proceedings of the IEEE/CVF Conference on Computer Vision and Pattern Recognition}, pages 13700--13710, 2024.

\bibitem[Kingma and Ba(2014)]{Kingma_Ba_2014}
DiederikP. Kingma and Jimmy Ba.
\newblock Adam: A method for stochastic optimization.
\newblock \emph{arXiv: Learning,arXiv: Learning}, 2014.

\bibitem[Kirillov et~al.(2023)Kirillov, Mintun, Ravi, Mao, Rolland, Gustafson, Xiao, Whitehead, Berg, Lo, et~al.]{kirillov2023segment}
Alexander Kirillov, Eric Mintun, Nikhila Ravi, Hanzi Mao, Chloe Rolland, Laura Gustafson, Tete Xiao, Spencer Whitehead, Alexander~C Berg, Wan-Yen Lo, et~al.
\newblock Segment anything.
\newblock In \emph{Proceedings of the IEEE/CVF International Conference on Computer Vision}, pages 4015--4026, 2023.

\bibitem[Li et~al.(2024{\natexlab{a}})Li, Yang, Kuang, Wu, Wang, Xiao, and Chen]{controlnet_plus_plus}
Ming Li, Taojiannan Yang, Huafeng Kuang, Jie Wu, Zhaoning Wang, Xuefeng Xiao, and Chen Chen.
\newblock Controlnet++: Improving conditional controls with efficient consistency feedback.
\newblock In \emph{European Conference on Computer Vision (ECCV)}, 2024{\natexlab{a}}.

\bibitem[Li et~al.(2024{\natexlab{b}})Li, Lin, Ye, Chen, Cheng, Yuan, and Yuan]{li2024wfvaeenhancingvideovae}
Zongjian Li, Bin Lin, Yang Ye, Liuhan Chen, Xinhua Cheng, Shenghai Yuan, and Li Yuan.
\newblock Wf-vae: Enhancing video vae by wavelet-driven energy flow for latent video diffusion model, 2024{\natexlab{b}}.

\bibitem[Li et~al.(2024{\natexlab{c}})Li, Zhang, Lin, Xiong, Long, Deng, Zhang, Liu, Huang, Xiao, et~al.]{li2024hunyuan}
Zhimin Li, Jianwei Zhang, Qin Lin, Jiangfeng Xiong, Yanxin Long, Xinchi Deng, Yingfang Zhang, Xingchao Liu, Minbin Huang, Zedong Xiao, et~al.
\newblock Hunyuan-dit: A powerful multi-resolution diffusion transformer with fine-grained chinese understanding.
\newblock \emph{arXiv preprint arXiv:2405.08748}, 2024{\natexlab{c}}.

\bibitem[Liao et~al.(2021)Liao, Tu, Xia, Liu, Zhou, Yuan, and Hu]{liao2021ascend}
Heng Liao, Jiajin Tu, Jing Xia, Hu Liu, Xiping Zhou, Honghui Yuan, and Yuxing Hu.
\newblock Ascend: a scalable and unified architecture for ubiquitous deep neural network computing: Industry track paper.
\newblock In \emph{2021 IEEE International Symposium on High-Performance Computer Architecture (HPCA)}, pages 789--801. IEEE, 2021.

\bibitem[Lin et~al.(2023)Lin, Ye, Zhu, Cui, Ning, Jin, and Yuan]{lin2023video}
Bin Lin, Yang Ye, Bin Zhu, Jiaxi Cui, Munan Ning, Peng Jin, and Li Yuan.
\newblock Video-llava: Learning united visual representation by alignment before projection.
\newblock \emph{arXiv preprint arXiv:2311.10122}, 2023.

\bibitem[Lin et~al.(2024{\natexlab{a}})Lin, Tang, Ye, Cui, Zhu, Jin, Huang, Zhang, Pang, Ning, et~al.]{lin2024moe}
Bin Lin, Zhenyu Tang, Yang Ye, Jiaxi Cui, Bin Zhu, Peng Jin, Jinfa Huang, Junwu Zhang, Yatian Pang, Munan Ning, et~al.
\newblock Moe-llava: Mixture of experts for large vision-language models.
\newblock \emph{arXiv preprint arXiv:2401.15947}, 2024{\natexlab{a}}.

\bibitem[Lin et~al.(2024{\natexlab{b}})Lin, Liu, Li, and Yang]{lin2024common}
Shanchuan Lin, Bingchen Liu, Jiashi Li, and Xiao Yang.
\newblock Common diffusion noise schedules and sample steps are flawed.
\newblock In \emph{Proceedings of the IEEE/CVF winter conference on applications of computer vision}, pages 5404--5411, 2024{\natexlab{b}}.

\bibitem[Lin et~al.(2014)Lin, Maire, Belongie, Hays, Perona, Ramanan, Doll{\'a}r, and Zitnick]{lin2014microsoft}
Tsung-Yi Lin, Michael Maire, Serge Belongie, James Hays, Pietro Perona, Deva Ramanan, Piotr Doll{\'a}r, and C~Lawrence Zitnick.
\newblock Microsoft coco: Common objects in context.
\newblock In \emph{Computer Vision--ECCV 2014: 13th European Conference, Zurich, Switzerland, September 6-12, 2014, Proceedings, Part V 13}, pages 740--755. Springer, 2014.

\bibitem[Lipman et~al.(2022)Lipman, Chen, Ben-Hamu, Nickel, and Le]{lipman2022flow}
Yaron Lipman, Ricky~TQ Chen, Heli Ben-Hamu, Maximilian Nickel, and Matt Le.
\newblock Flow matching for generative modeling.
\newblock \emph{arXiv preprint arXiv:2210.02747}, 2022.

\bibitem[Liu et~al.(2024{\natexlab{a}})Liu, Akhgari, Visheratin, Kamko, Xu, Shrirao, Souza, Doshi, and Li]{liu2024playground}
Bingchen Liu, Ehsan Akhgari, Alexander Visheratin, Aleks Kamko, Linmiao Xu, Shivam Shrirao, Joao Souza, Suhail Doshi, and Daiqing Li.
\newblock Playground v3: Improving text-to-image alignment with deep-fusion large language models.
\newblock \emph{arXiv preprint arXiv:2409.10695}, 2024{\natexlab{a}}.

\bibitem[Liu et~al.(2024{\natexlab{b}})Liu, Li, Li, and Lee]{liu2024improved}
Haotian Liu, Chunyuan Li, Yuheng Li, and Yong~Jae Lee.
\newblock Improved baselines with visual instruction tuning.
\newblock In \emph{Proceedings of the IEEE/CVF Conference on Computer Vision and Pattern Recognition}, pages 26296--26306, 2024{\natexlab{b}}.

\bibitem[Liu et~al.(2024{\natexlab{c}})Liu, Li, Wu, and Lee]{liu2024visual}
Haotian Liu, Chunyuan Li, Qingyang Wu, and Yong~Jae Lee.
\newblock Visual instruction tuning.
\newblock \emph{Advances in neural information processing systems}, 36, 2024{\natexlab{c}}.

\bibitem[Liu et~al.(2024{\natexlab{d}})Liu, Tang, Liu, Ge, Shan, Li, and Yang]{liu2024ppllava}
Ruyang Liu, Haoran Tang, Haibo Liu, Yixiao Ge, Ying Shan, Chen Li, and Jiankun Yang.
\newblock Ppllava: Varied video sequence understanding with prompt guidance.
\newblock \emph{arXiv preprint arXiv:2411.02327}, 2024{\natexlab{d}}.

\bibitem[Loshchilov and Hutter(2019)]{loshchilov2019decoupledweightdecayregularization}
Ilya Loshchilov and Frank Hutter.
\newblock Decoupled weight decay regularization, 2019.

\bibitem[Lu et~al.(2024)Lu, Wang, Huang, Wu, Liu, Ouyang, and Bai]{Lu2024FiT}
Zeyu Lu, Zidong Wang, Di Huang, Chengyue Wu, Xihui Liu, Wanli Ouyang, and Lei Bai.
\newblock Fit: Flexible vision transformer for diffusion model.
\newblock \emph{arXiv preprint arXiv:2402.12376}, 2024.

\bibitem[Mou et~al.(2024)Mou, Wang, Xie, Wu, Zhang, Qi, and Shan]{t2iadapter}
Chong Mou, Xintao Wang, Liangbin Xie, Yanze Wu, Jian Zhang, Zhongang Qi, and Ying Shan.
\newblock T2i-adapter: Learning adapters to dig out more controllable ability for text-to-image diffusion models.
\newblock In \emph{Proceedings of the AAAI Conference on Artificial Intelligence}, pages 4296--4304, 2024.

\bibitem[OpenAI(2023)]{openai2023gpt4}
OpenAI.
\newblock Gpt-4 technical report, 2023.

\bibitem[Peebles and Xie(2023)]{peebles2023scalable}
William Peebles and Saining Xie.
\newblock Scalable diffusion models with transformers.
\newblock In \emph{Proceedings of the IEEE/CVF International Conference on Computer Vision}, pages 4195--4205, 2023.

\bibitem[Peng et~al.(2024)Peng, Wang, Zhang, Li, Yang, and Jia]{controlnext}
Bohao Peng, Jian Wang, Yuechen Zhang, Wenbo Li, Ming-Chang Yang, and Jiaya Jia.
\newblock Controlnext: Powerful and efficient control for image and video generation.
\newblock \emph{arXiv preprint arXiv:2408.06070}, 2024.

\bibitem[Polyak et~al.(2024)Polyak, Zohar, Brown, Tjandra, Sinha, Lee, Vyas, Shi, Ma, Chuang, et~al.]{polyak2024movie}
Adam Polyak, Amit Zohar, Andrew Brown, Andros Tjandra, Animesh Sinha, Ann Lee, Apoorv Vyas, Bowen Shi, Chih-Yao Ma, Ching-Yao Chuang, et~al.
\newblock Movie gen: A cast of media foundation models.
\newblock \emph{arXiv preprint arXiv:2410.13720}, 2024.

\bibitem[Radford et~al.(2021)Radford, Kim, Hallacy, Ramesh, Goh, Agarwal, Sastry, Askell, Mishkin, Clark, et~al.]{clip}
Alec Radford, Jong~Wook Kim, Chris Hallacy, Aditya Ramesh, Gabriel Goh, Sandhini Agarwal, Girish Sastry, Amanda Askell, Pamela Mishkin, Jack Clark, et~al.
\newblock Learning transferable visual models from natural language supervision.
\newblock In \emph{International conference on machine learning}, pages 8748--8763. PMLR, 2021.

\bibitem[Rasley et~al.(2020)Rasley, Rajbhandari, Ruwase, and He]{rasley2020deepspeed}
Jeff Rasley, Samyam Rajbhandari, Olatunji Ruwase, and Yuxiong He.
\newblock Deepspeed: System optimizations enable training deep learning models with over 100 billion parameters.
\newblock In \emph{Proceedings of the 26th ACM SIGKDD International Conference on Knowledge Discovery \& Data Mining}, pages 3505--3506, 2020.

\bibitem[Rombach et~al.(2022{\natexlab{a}})Rombach, Blattmann, Lorenz, Esser, and Ommer]{rombach2022high}
Robin Rombach, Andreas Blattmann, Dominik Lorenz, Patrick Esser, and Bj{\"o}rn Ommer.
\newblock High-resolution image synthesis with latent diffusion models.
\newblock In \emph{Proceedings of the IEEE/CVF conference on computer vision and pattern recognition}, pages 10684--10695, 2022{\natexlab{a}}.

\bibitem[Rombach et~al.(2022{\natexlab{b}})Rombach, Blattmann, Lorenz, Esser, and Ommer]{stable_diffusion}
Robin Rombach, Andreas Blattmann, Dominik Lorenz, Patrick Esser, and Bj\"orn Ommer.
\newblock High-resolution image synthesis with latent diffusion models.
\newblock In \emph{Proceedings of the IEEE/CVF Conference on Computer Vision and Pattern Recognition (CVPR)}, pages 10684--10695, 2022{\natexlab{b}}.

\bibitem[Schuhmann et~al.(2022)Schuhmann, Beaumont, Vencu, Gordon, Wightman, Cherti, Coombes, Katta, Mullis, Wortsman, et~al.]{schuhmann2022laion}
Christoph Schuhmann, Romain Beaumont, Richard Vencu, Cade Gordon, Ross Wightman, Mehdi Cherti, Theo Coombes, Aarush Katta, Clayton Mullis, Mitchell Wortsman, et~al.
\newblock Laion-5b: An open large-scale dataset for training next generation image-text models.
\newblock \emph{Advances in Neural Information Processing Systems}, 35:\penalty0 25278--25294, 2022.

\bibitem[Shoeybi et~al.(2019)Shoeybi, Patwary, Puri, LeGresley, Casper, and Catanzaro]{shoeybi2019megatron}
Mohammad Shoeybi, Mostofa Patwary, Raul Puri, Patrick LeGresley, Jared Casper, and Bryan Catanzaro.
\newblock Megatron-lm: Training multi-billion parameter language models using model parallelism.
\newblock \emph{arXiv preprint arXiv:1909.08053}, 2019.

\bibitem[Song et~al.(2020)Song, Meng, and Ermon]{song2020denoising}
Jiaming Song, Chenlin Meng, and Stefano Ermon.
\newblock Denoising diffusion implicit models.
\newblock \emph{arXiv preprint arXiv:2010.02502}, 2020.

\bibitem[Su et~al.(2024)Su, Ahmed, Lu, Pan, Bo, and Liu]{su2024roformer}
Jianlin Su, Murtadha Ahmed, Yu Lu, Shengfeng Pan, Wen Bo, and Yunfeng Liu.
\newblock Roformer: Enhanced transformer with rotary position embedding.
\newblock \emph{Neurocomputing}, 568:\penalty0 127063, 2024.

\bibitem[Su et~al.(2021)Su, Liu, Yu, Hu, Liao, Tian, Pietik{\"a}inen, and Liu]{pidinet}
Zhuo Su, Wenzhe Liu, Zitong Yu, Dewen Hu, Qing Liao, Qi Tian, Matti Pietik{\"a}inen, and Li Liu.
\newblock Pixel difference networks for efficient edge detection.
\newblock In \emph{Proceedings of the IEEE/CVF international conference on computer vision}, pages 5117--5127, 2021.

\bibitem[Sun et~al.(2024)Sun, Pan, Ge, Li, Duan, Wu, Zhang, Zhou, Qin, Wang, et~al.]{sun2024journeydb}
Keqiang Sun, Junting Pan, Yuying Ge, Hao Li, Haodong Duan, Xiaoshi Wu, Renrui Zhang, Aojun Zhou, Zipeng Qin, Yi Wang, et~al.
\newblock Journeydb: A benchmark for generative image understanding.
\newblock \emph{Advances in Neural Information Processing Systems}, 36, 2024.

\bibitem[Tuo et~al.(2023)Tuo, Xiang, He, Geng, and Xie]{tuo2023anytext}
Yuxiang Tuo, Wangmeng Xiang, Jun-Yan He, Yifeng Geng, and Xuansong Xie.
\newblock Anytext: Multilingual visual text generation and editing.
\newblock \emph{arXiv preprint arXiv:2311.03054}, 2023.

\bibitem[Unterthiner et~al.(2019)Unterthiner, Steenkiste, Kurach, Marinier, Michalski, and Gelly]{Unterthiner_Steenkiste_Kurach_Marinier_Michalski_Gelly_2019}
Thomas Unterthiner, Sjoerdvan Steenkiste, Karol Kurach, Raphaël Marinier, Marcin Michalski, and Sylvain Gelly.
\newblock Fvd: A new metric for video generation.
\newblock \emph{International Conference on Learning Representations,International Conference on Learning Representations}, 2019.

\bibitem[Vaswani(2017)]{vaswani2017attention}
A Vaswani.
\newblock Attention is all you need.
\newblock \emph{Advances in Neural Information Processing Systems}, 2017.

\bibitem[Wang et~al.(2024{\natexlab{a}})Wang, Yuan, and Zhang]{wang2024tarsier}
Jiawei Wang, Liping Yuan, and Yuchen Zhang.
\newblock Tarsier: Recipes for training and evaluating large video description models.
\newblock \emph{arXiv preprint arXiv:2407.00634}, 2024{\natexlab{a}}.

\bibitem[Wang et~al.(2022{\natexlab{a}})Wang, Yang, Men, Lin, Bai, Li, Ma, Zhou, Zhou, and Yang]{wang2022ofa}
Peng Wang, An Yang, Rui Men, Junyang Lin, Shuai Bai, Zhikang Li, Jianxin Ma, Chang Zhou, Jingren Zhou, and Hongxia Yang.
\newblock Ofa: Unifying architectures, tasks, and modalities through a simple sequence-to-sequence learning framework.
\newblock In \emph{International conference on machine learning}, pages 23318--23340. PMLR, 2022{\natexlab{a}}.

\bibitem[Wang et~al.(2024{\natexlab{b}})Wang, Bai, Tan, Wang, Fan, Bai, Chen, Liu, Wang, Ge, et~al.]{wang2024qwen2}
Peng Wang, Shuai Bai, Sinan Tan, Shijie Wang, Zhihao Fan, Jinze Bai, Keqin Chen, Xuejing Liu, Jialin Wang, Wenbin Ge, et~al.
\newblock Qwen2-vl: Enhancing vision-language model's perception of the world at any resolution.
\newblock \emph{arXiv preprint arXiv:2409.12191}, 2024{\natexlab{b}}.

\bibitem[Wang et~al.(2023)Wang, Lv, Yu, Hong, Qi, Wang, Ji, Yang, Zhao, Song, et~al.]{wang2023cogvlm}
Weihan Wang, Qingsong Lv, Wenmeng Yu, Wenyi Hong, Ji Qi, Yan Wang, Junhui Ji, Zhuoyi Yang, Lei Zhao, Xixuan Song, et~al.
\newblock Cogvlm: Visual expert for pretrained language models.
\newblock \emph{arXiv preprint arXiv:2311.03079}, 2023.

\bibitem[Wang et~al.(2004)Wang, Bovik, Sheikh, and Simoncelli]{wang2004image}
Zhou Wang, Alan~C Bovik, Hamid~R Sheikh, and Eero~P Simoncelli.
\newblock Image quality assessment: from error visibility to structural similarity.
\newblock \emph{IEEE transactions on image processing}, 13\penalty0 (4):\penalty0 600--612, 2004.

\bibitem[Wang et~al.(2024{\natexlab{c}})Wang, Lu, Huang, Zhou, Ouyang, et~al.]{wang2024fitv2}
ZiDong Wang, Zeyu Lu, Di Huang, Cai Zhou, Wanli Ouyang, et~al.
\newblock Fitv2: Scalable and improved flexible vision transformer for diffusion model.
\newblock \emph{arXiv preprint arXiv:2410.13925}, 2024{\natexlab{c}}.

\bibitem[Wang et~al.(2022{\natexlab{b}})Wang, Montoya, Munechika, Yang, Hoover, and Chau]{wang2022diffusiondb}
Zijie~J Wang, Evan Montoya, David Munechika, Haoyang Yang, Benjamin Hoover, and Duen~Horng Chau.
\newblock Diffusiondb: A large-scale prompt gallery dataset for text-to-image generative models.
\newblock \emph{arXiv preprint arXiv:2210.14896}, 2022{\natexlab{b}}.

\bibitem[Wu et~al.(2023)Wu, Zhang, Liao, Chen, Hou, Wang, Sun, Yan, and Lin]{wu2023exploring}
Haoning Wu, Erli Zhang, Liang Liao, Chaofeng Chen, Jingwen Hou, Annan Wang, Wenxiu Sun, Qiong Yan, and Weisi Lin.
\newblock Exploring video quality assessment on user generated contents from aesthetic and technical perspectives.
\newblock In \emph{Proceedings of the IEEE/CVF International Conference on Computer Vision}, pages 20144--20154, 2023.

\bibitem[Xing et~al.(2025)Xing, Xia, Zhang, Chen, Yu, Liu, Liu, Wang, Shan, and Wong]{dynamicrafter}
Jinbo Xing, Menghan Xia, Yong Zhang, Haoxin Chen, Wangbo Yu, Hanyuan Liu, Gongye Liu, Xintao Wang, Ying Shan, and Tien-Tsin Wong.
\newblock Dynamicrafter: Animating open-domain images with video diffusion priors.
\newblock In \emph{European Conference on Computer Vision}, pages 399--417. Springer, 2025.

\bibitem[Xu et~al.(2024{\natexlab{a}})Xu, Zou, Huang, Chen, Liu, Cheng, Shi, and Huang]{easyanimate}
Jiaqi Xu, Xinyi Zou, Kunzhe Huang, Yunkuo Chen, Bo Liu, MengLi Cheng, Xing Shi, and Jun Huang.
\newblock Easyanimate: A high-performance long video generation method based on transformer architecture.
\newblock \emph{arXiv preprint arXiv:2405.18991}, 2024{\natexlab{a}}.

\bibitem[Xu et~al.(2024{\natexlab{b}})Xu, Zou, Huang, Chen, Liu, Cheng, Shi, and Huang]{xu2024easyanimate}
Jiaqi Xu, Xinyi Zou, Kunzhe Huang, Yunkuo Chen, Bo Liu, MengLi Cheng, Xing Shi, and Jun Huang.
\newblock Easyanimate: A high-performance long video generation method based on transformer architecture.
\newblock \emph{arXiv preprint arXiv:2405.18991}, 2024{\natexlab{b}}.

\bibitem[Xu et~al.(2024{\natexlab{c}})Xu, Zhao, Zhou, Lin, Ng, and Feng]{xu2024pllava}
Lin Xu, Yilin Zhao, Daquan Zhou, Zhijie Lin, See~Kiong Ng, and Jiashi Feng.
\newblock Pllava: Parameter-free llava extension from images to videos for video dense captioning.
\newblock \emph{arXiv preprint arXiv:2404.16994}, 2024{\natexlab{c}}.

\bibitem[Xue et~al.(2022)Xue, Hang, Zeng, Sun, Liu, Yang, Fu, and Guo]{xue2022hdvila}
Hongwei Xue, Tiankai Hang, Yanhong Zeng, Yuchong Sun, Bei Liu, Huan Yang, Jianlong Fu, and Baining Guo.
\newblock Advancing high-resolution video-language representation with large-scale video transcriptions.
\newblock In \emph{International Conference on Computer Vision and Pattern Recognition (CVPR)}, 2022.

\bibitem[Xue(2020)]{xue2020mt5}
L Xue.
\newblock mt5: A massively multilingual pre-trained text-to-text transformer.
\newblock \emph{arXiv preprint arXiv:2010.11934}, 2020.

\bibitem[Yang et~al.(2024{\natexlab{a}})Yang, Yang, Hui, Zheng, Yu, Zhou, Li, Li, Liu, Huang, et~al.]{yang2024qwen2}
An Yang, Baosong Yang, Binyuan Hui, Bo Zheng, Bowen Yu, Chang Zhou, Chengpeng Li, Chengyuan Li, Dayiheng Liu, Fei Huang, et~al.
\newblock Qwen2 technical report.
\newblock \emph{arXiv preprint arXiv:2407.10671}, 2024{\natexlab{a}}.

\bibitem[Yang et~al.(2024{\natexlab{b}})Yang, Teng, Zheng, Ding, Huang, Xu, Yang, Hong, Zhang, Feng, et~al.]{yang2024cogvideox}
Zhuoyi Yang, Jiayan Teng, Wendi Zheng, Ming Ding, Shiyu Huang, Jiazheng Xu, Yuanming Yang, Wenyi Hong, Xiaohan Zhang, Guanyu Feng, et~al.
\newblock Cogvideox: Text-to-video diffusion models with an expert transformer.
\newblock \emph{arXiv preprint arXiv:2408.06072}, 2024{\natexlab{b}}.

\bibitem[Yao et~al.(2024)Yao, Yu, Zhang, Wang, Cui, Zhu, Cai, Li, Zhao, He, et~al.]{yao2024minicpm}
Yuan Yao, Tianyu Yu, Ao Zhang, Chongyi Wang, Junbo Cui, Hongji Zhu, Tianchi Cai, Haoyu Li, Weilin Zhao, Zhihui He, et~al.
\newblock Minicpm-v: A gpt-4v level mllm on your phone.
\newblock \emph{arXiv preprint arXiv:2408.01800}, 2024.

\bibitem[Ye et~al.(2023)Ye, Xu, Xu, Ye, Yan, Zhou, Wang, Hu, Shi, Shi, et~al.]{ye2023mplug}
Qinghao Ye, Haiyang Xu, Guohai Xu, Jiabo Ye, Ming Yan, Yiyang Zhou, Junyang Wang, Anwen Hu, Pengcheng Shi, Yaya Shi, et~al.
\newblock mplug-owl: Modularization empowers large language models with multimodality.
\newblock \emph{arXiv preprint arXiv:2304.14178}, 2023.

\bibitem[Young et~al.(2024)Young, Chen, Li, Huang, Zhang, Zhang, Li, Zhu, Chen, Chang, et~al.]{young2024yi}
Alex Young, Bei Chen, Chao Li, Chengen Huang, Ge Zhang, Guanwei Zhang, Heng Li, Jiangcheng Zhu, Jianqun Chen, Jing Chang, et~al.
\newblock Yi: Open foundation models by 01. ai.
\newblock \emph{arXiv preprint arXiv:2403.04652}, 2024.

\bibitem[Yu et~al.(2024)Yu, Lezama, Gundavarapu, Versari, Sohn, Minnen, Cheng, Birodkar, Gupta, Gu, Hauptmann, Gong, Yang, Essa, Ross, and Jiang]{yu2024languagemodelbeatsdiffusion}
Lijun Yu, José Lezama, Nitesh~B. Gundavarapu, Luca Versari, Kihyuk Sohn, David Minnen, Yong Cheng, Vighnesh Birodkar, Agrim Gupta, Xiuye Gu, Alexander~G. Hauptmann, Boqing Gong, Ming-Hsuan Yang, Irfan Essa, David~A. Ross, and Lu Jiang.
\newblock Language model beats diffusion -- tokenizer is key to visual generation, 2024.

\bibitem[Yuan et~al.(2024)Yuan, Huang, Xu, Liu, Zhang, Shi, Zhu, Cheng, Luo, and Yuan]{chronomagic_bench}
Shenghai Yuan, Jinfa Huang, Yongqi Xu, Yaoyang Liu, Shaofeng Zhang, Yujun Shi, Ruijie Zhu, Xinhua Cheng, Jiebo Luo, and Li Yuan.
\newblock Chronomagic-bench: A benchmark for metamorphic evaluation of text-to-time-lapse video generation.
\newblock \emph{arXiv preprint arXiv:2406.18522}, 2024.

\bibitem[Zhang et~al.(2023)Zhang, Rao, and Agrawala]{controlnet}
Lvmin Zhang, Anyi Rao, and Maneesh Agrawala.
\newblock Adding conditional control to text-to-image diffusion models.
\newblock In \emph{Proceedings of the IEEE/CVF International Conference on Computer Vision}, pages 3836--3847, 2023.

\bibitem[Zhang et~al.(2018)Zhang, Isola, Efros, Shechtman, and Wang]{Zhang_Isola_Efros_Shechtman_Wang_2018}
Richard Zhang, Phillip Isola, Alexei~A. Efros, Eli Shechtman, and Oliver Wang.
\newblock The unreasonable effectiveness of deep features as a perceptual metric.
\newblock In \emph{2018 IEEE/CVF Conference on Computer Vision and Pattern Recognition}, 2018.

\bibitem[Zhao et~al.(2023)Zhao, Gu, Varma, Luo, Huang, Xu, Wright, Shojanazeri, Ott, Shleifer, et~al.]{zhao2023pytorch}
Yanli Zhao, Andrew Gu, Rohan Varma, Liang Luo, Chien-Chin Huang, Min Xu, Less Wright, Hamid Shojanazeri, Myle Ott, Sam Shleifer, et~al.
\newblock Pytorch fsdp: experiences on scaling fully sharded data parallel.
\newblock \emph{arXiv preprint arXiv:2304.11277}, 2023.

\bibitem[Zheng et~al.(2024)Zheng, Peng, Yang, Shen, Li, Liu, Zhou, Li, and You]{opensora}
Zangwei Zheng, Xiangyu Peng, Tianji Yang, Chenhui Shen, Shenggui Li, Hongxin Liu, Yukun Zhou, Tianyi Li, and Yang You.
\newblock Open-sora: Democratizing efficient video production for all, 2024.

\bibitem[Zhou et~al.(2024)Zhou, Wang, Cai, and Yang]{zhou2024allegro}
Yuan Zhou, Qiuyue Wang, Yuxuan Cai, and Huan Yang.
\newblock Allegro: Open the black box of commercial-level video generation model.
\newblock \emph{arXiv preprint arXiv:2410.15458}, 2024.

\bibitem[Zhu et~al.(2023)Zhu, Lin, Ning, Yan, Cui, Wang, Pang, Jiang, Zhang, Li, et~al.]{zhu2023languagebind}
Bin Zhu, Bin Lin, Munan Ning, Yang Yan, Jiaxi Cui, HongFa Wang, Yatian Pang, Wenhao Jiang, Junwu Zhang, Zongwei Li, et~al.
\newblock Languagebind: Extending video-language pretraining to n-modality by language-based semantic alignment.
\newblock \emph{arXiv preprint arXiv:2310.01852}, 2023.

\end{thebibliography}
